\pdfoutput=1

\documentclass[11pt]{article}

\usepackage[]{acl}

\usepackage{times}
\usepackage{latexsym}

\usepackage[T1]{fontenc}

\usepackage[utf8]{inputenc}

\usepackage{microtype}

\usepackage{inconsolata}

\newcommand\sect[1]{\S\ref{#1}}

\usepackage{amsmath}
\usepackage{graphicx}
\usepackage{booktabs}
\usepackage{colortbl}
\usepackage{multirow}
\usepackage{makecell}
\usepackage{tablefootnote}
\usepackage{paralist}
\usepackage{mathabx}
\usepackage{bbm} 
\usepackage{comment}
\usepackage{subcaption}
\usepackage{mathtools}
\usepackage{xspace}
\usepackage{xcolor}
\usepackage{float}
\usepackage{pifont}%
\newcommand{\cmark}{\ding{51}}%
\newcommand{\xmark}{\ding{55}}%

\DeclareMathOperator{\detect}{detect}
\DeclareMathOperator{\pert}{pert}

\DeclareMathOperator{\as}{AS}
\DeclareMathOperator{\asr}{ASR}
\DeclareMathOperator{\dfr}{DFR}
\DeclareMathOperator{\lar}{LAR}

\DeclareMathOperator{\dsr}{DSR}
\DeclareMathOperator{\attack}{A}

\definecolor{mycolor}{rgb}{0.9, 0.9, 0.9} %

\title{Whispers of Doubt Amidst Echoes of Triumph in NLP Robustness}
\author{
\begin{tabular}{c}
Ashim Gupta \quad\quad\quad Rishanth Rajendhran \quad\quad\quad Nathan Stringham\\ 
Vivek Srikumar \quad\quad\quad Ana Marasovi\'c
\end{tabular}\\
Kahlert School of Computing \\
University of Utah \\
\texttt{ashim@cs.utah.edu}}

\begin{document}
\maketitle
\begin{abstract}
\emph{Do larger and more performant models resolve NLP's longstanding robustness issues?} We investigate this question using over 20 models of different sizes spanning different architectural choices and pretraining objectives. %
We conduct evaluations using (a) %
out-of-domain and challenge test sets, (b) behavioral testing with CheckLists, (c) contrast sets, and (d) adversarial inputs. %
Our analysis reveals that not all out-of-domain tests provide insight into robustness. 
Evaluating with CheckLists and contrast sets shows significant gaps in model performance; merely scaling models does not make them adequately robust. %
Finally, we point out that current approaches for adversarial evaluations of models are themselves problematic: they can be easily thwarted, and in their current forms, do not represent a sufficiently deep probe of model robustness. %
We conclude that not only is the question of robustness in NLP as yet unresolved, but even some of the approaches to measure robustness need to be reassessed. 
\end{abstract}
\begin{figure*}[t]
\begin{minipage}[b]{0.68\textwidth}
\includegraphics[width=\textwidth]{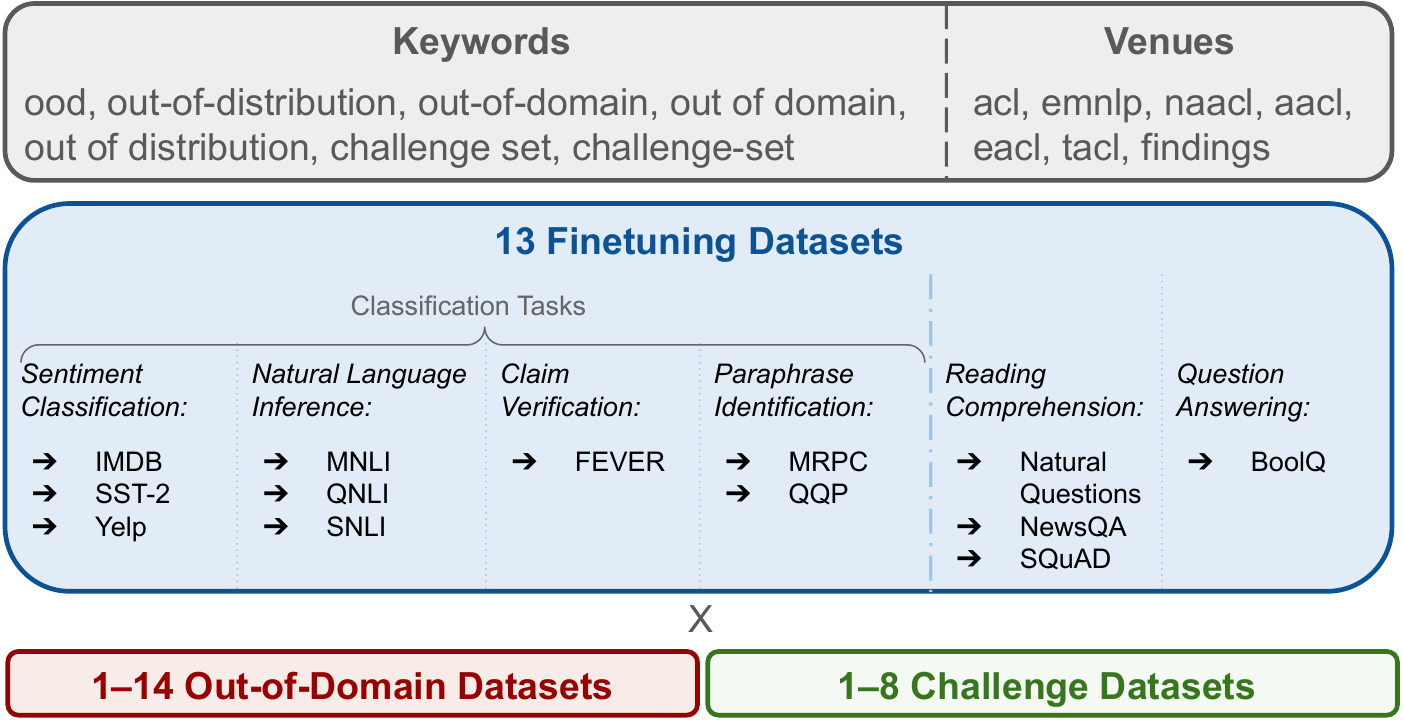}
\caption{Finetuning and evaluation datasets determined by analyzing train-test splits in *ACL/EMNLP publications from 2020--2022 (\sect{sec:setup}). Individual train-test splits are reported in Table \ref{tab:train_eval2}. They represent the most common data setups for studying two popular aspects of NLP robustness.}
\label{fig:meta_analysis}
\end{minipage}
\hfill
\begin{minipage}[b]{0.29\textwidth}
\resizebox{\textwidth}{!}{%
\begin{tabular}{llrl}
\toprule
& \textbf{Model}  & \textbf{Size}  & \textbf{PT} \\
\midrule
\multirow{4.5}{*}{\rotatebox{90}{\small \emph{Encoder-Only}}}  & RoBERTa-Base     & 124M  & \multirow{2}{*}{MLM}                   \\
& RoBERTa-Large    & 355M    &                    \\
\arrayrulecolor{black!20}\cmidrule{2-4}
& DeBERTa-v3-Base  & 184M    & \multirow{2}{*}{MLM}                    \\
& DeBERTa-v3-Large & 435M   &                    \\
\arrayrulecolor{black}\midrule
\multirow{10}{*}{\rotatebox{90}{\small \emph{Decoder-Only}}} & OPT-125M         & 125M    & \multirow{6}{*}{LM}                     \\
& OPT-350M         & 331M    &                     \\
& OPT-1.3B         & 1.3B  &                     \\
& OPT-2.7B         & 2.7B  &                     \\
& OPT-6.7B         & 6.7B &                     \\
& OPT-13B         & 12.8B &                     \\
\arrayrulecolor{black!20}\cmidrule{2-4}
& GPT-2    & 124M   & \multirow{4}{*}{LM}                     \\
& GPT-2-Medium     & 354M    &                     \\
& GPT-2-Large      & 774M   &                     \\
& GPT-2-XL         &   1.6B       &                     \\
\arrayrulecolor{black}\midrule
\multirow{5}{*}{\rotatebox{90}{\small\emph{Encoder-Decoder}}} & T5-Small         & 60M     & \multirow{5}{*}{\rotatebox{90}{\makecell[l]{text-to-text\\MLM + MTL}}}       \\
& T5-Base          & 222M    &        \\
& T5-Large         & 737M    &      \\
& T5-XL (3B)         & 2.8B  &       \\
& T5-XXL (11B)       & 11.3B &    \\
\bottomrule
\end{tabular}%
}
\captionof{table}{Finetuning models. }
\label{tab:models}
\end{minipage}
\end{figure*}

\begin{figure*}[t]
  \centering
    \includegraphics[width=\textwidth]{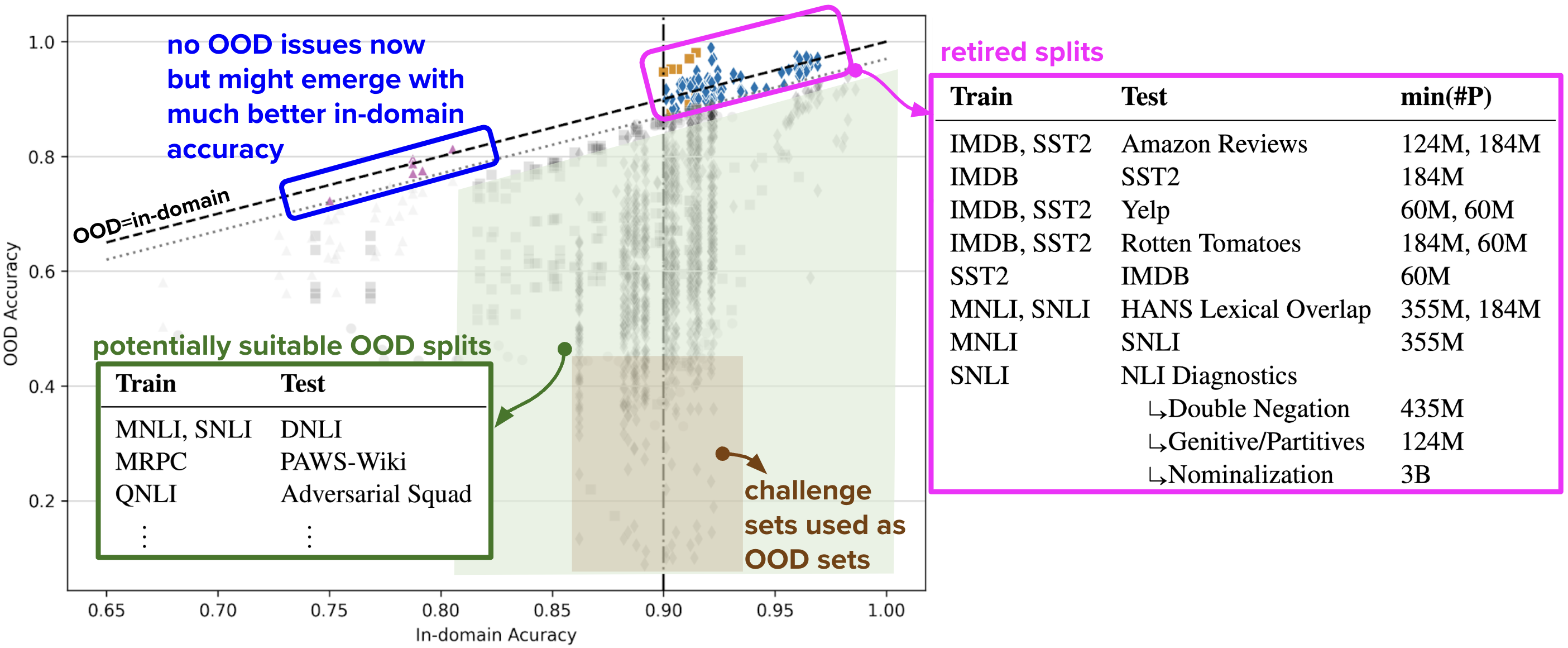}
    \caption{In-domain vs.\ OOD accuracy of 19 models finetuned for 4 types of \emph{classification} tasks across 9 training datasets and 1--14 OOD datasets per training set. %
    The dashed line is where OOD accuracy equals in-domain and the dotted where it is at most 3\% lower. min(\#P) is the number of parameters of the smallest model that achieves the latter. The gray points below the dotted line are linked with data splits that do not appear in the upper right region.
    }
    \label{fig:ood_main}
\end{figure*}

\section{Introduction}

The versatility and consumer growth of commercial LLMs like ChatGPT give the impression that robustness evaluations such as out-of-domain (OOD) and stress testing are no longer relevant. %
We argue that they remain important. %
Many applications do not need the broad range of skills offered by general-purpose models, from writing clinical notes to drawing \LaTeX~unicorns~\cite{DBLP:journals/corr/abs-2303-12712}. %
Specializing moderate-scale models with finetuning still works better when a model must perform a specific task.\footnote{For example, see this blogpost (retrieved Nov 15, 2023): \href{https://www.anyscale.com/blog/fine-tuning-llama-2-a-comprehensive-case-study-for-tailoring-models-to-unique-applications}{Fine-Tuning Llama-2: A Comprehensive Case Study for Tailoring Models to Unique Applications}.} 
Some NLP research embraces this view, but takes it to an extreme. %
\citet{bowman-2022-dangers} highlights the continued use of the 2018 BERT-base model as a baseline, even though larger and better-pretrained models can be finetuned on a single GPU today.
The rush to innovate upon general-purpose models and the disregard of stronger baselines make it unclear where the field stands in terms of established robustness evaluations. %
We seek to address this gap. 

We first point to popular experimental setups that, while of interest, may be outdated for robustness studies. %
Specifically, we record the train-test splits of 177 ACL publications that include keywords in Figure  \ref{fig:meta_analysis}. %
After filtering, 101 splits remain that we use to finetune and evaluate 19 models that differ in (i) transformer type (encoder-/decoder-only, encoder-decoder), (ii) model size (60M to 13B), and (iii) pretraining objectives (LM/MLM only, additional multitask pretraining). %
Table~\ref{tab:models} lists the models we finetune, and for focused evaluations about the in-context setting and scaling, we looked at Mistral, T\"uLU, and LLaMA-2.

We show that for 14 train/OOD-test pairs, a finetuned model that is over 90\% accuracy on the standard test set does not break on the OOD test set. %
Further research with these data splits is less likely to advance our understanding of OOD robustness. %
We also show that a few challenge sets continue to break NLI and reading comprehension models, but sentiment classifiers are robust to stress tests. %

Through behavioral testing with the CheckList methodology~\cite{ribeiro-etal-2020-beyond}, we show that highly accurate models still struggle with the most basic task phenomena. Larger models help, but do not fully resolve the issue; there is scant evidence that increasing size further would be beneficial.  %
Next, we employ contrast set evaluations~\cite{gardner-etal-2020-evaluating}, which measure model accuracy in a set of subtly different 
and differently labeled examples.
These evaluations continue to expose major model weaknesses, raising the question of why they are not more widely used. %
Finally, we stress a crucial finding\,---\,some robustness evaluations can themselves be fragile. %
We demonstrate this by revealing that the success of adversarial attacks is exaggerated. %
We define a more reliable metric for assessing adversarial attacks, but the overarching lesson is broader than the new metric: be cautious about assuming that prevailing evaluation methods are well-designed.

As the LLM landscape evolves, new challenges arise such as preventing the generation of text that assists unlawful or harmful activities. %
While these challenges are important, we show that many longstanding robustness issues in NLP remain relevant and unresolved. %
If we cared about these challenges before, why should we stop now?~\footnote{We release the code at: \url{https://github.com/utahnlp/scaling_robustness/}}

\section{Rethinking Common OOD Splits}
\label{sec:setup}

This section investigates the robustness of models in two scenarios:
\begin{inparaenum}[a)]
\item  when the train and test data sources differ  (\emph{OOD evaluation}), and %
\item  when inputs are designed to challenge models \cite[\emph{challenge set} or \emph{stress} test;][]{naik-etal-2018-stress}. %
\end{inparaenum}
We focus on experimental setups from recent ACL  publications to determine if they still present challenges.

\paragraph{Common OOD/Challenge Data Splits.} We manually collate train-test splits mentioned in 177 ACL publications from the years 2020--2022, containing one of the keywords listed in Figure~\ref{fig:meta_analysis}.\footnote{We used \href{https://www.semanticscholar.org/}{Semantic Scholar} for this effort.} %
From these, we select 13 training datasets (spanning 6 task types) that appear in at least 4 papers reporting an OOD or challenge set evaluation (Figure~\ref{fig:meta_analysis}, middle). %
Upon filtering to splits containing those training datasets, 101 splits remain; they are listed in Table~\ref{tab:tab:train_eval1} in the Appendix. %
Some splits have been used both for OOD and challenge set testing. 

\paragraph{Models.} We separately finetune the 19 models from Table~\ref{tab:models} on  the training datasets\footnote{See Appendix \ref{app:sec:experimental} for more details.} %
and report average performance across 3 random seeds. %
In addition, we assess the robustness of few-shot in-context learning in these settings using the base Mistral-7B model~\cite{jiang2023mistral}.\footnote{
Per the model release information, Mistral-7B's instruction data
does not include standard NLP training sets via collections such as (Super-)NaturalInstructions \cite{mishra-etal-2022-cross, wang-etal-2022-super} or T0 \cite{DBLP:conf/iclr/SanhWRBSACSRDBX22}. %
We expect that testing it on selected train-test splits aligns with the principle of OOD evaluation.} %

Finetuning the original, non-quantized version of models up to 13B parameters with a dataset with hundreds of thousands of examples such as MNLI or AGNews requires 
 model parallelism even with the largest GPUs like NVIDIA A100 80 GB. This is approaching the resource limits for many organizations. Future work could use our findings to further investigate the impact of quantization on robustness in fine-tuning scenarios.

\paragraph{Q1: Do commonly used OOD splits remain a valid choice for investigating OOD robustness?}

No, 14 splits involving an OOD test set may not adequately evaluate the accuracy of \emph{finetuned} models under distributional shifts. %
The upper right of Figure \ref{fig:ood_main}  lists train-test splits where at least one finetuned  model obtains in-domain accuracy over 90\% (right of the vertical line) that does not drop more than 3\% OOD (above the dotted line).  %
We allow a 3\% drop to account for variations due to randomness. %
These results are detailed in Tables \ref{tab:appendix_class_splits_part_1}--\ref{tab:appendix_class_splits_part_2} (Appendix). %
Close in-domain and OOD accuracy suggests that the model is robust  \cite{NEURIPS2020_d8330f85}. %
If accuracy also exceeds 90\%, supposed OOD issues are unlikely and concern over them is overstated\,---\,especially when achieved by a small model, as seen in most upper-right data splits. %

In contrast to these 14 successes, many OOD test sets remain robustness failures,
corresponding to results below the dotted line. %
They are the most fitting choices for research on  OOD robustness. %
They include almost all splits used for claim verification and paraphrase identification. %
Moreover, no reading comprehension model\,---\,\emph{finetuned or few-shot}\,---\,achieves an F1 score over 90\% while also maintaining similar OOD and in-domain scores (Tables \ref{tab:appendix_qa_splits}--\ref{tab:appendix_qa_splits_mistral}, Appendix). %
Sentiment classification and NLI offer limited opportunities for breakthroughs in OOD robustness compared to other tasks.  %

Splits with very low OOD accuracy ($<$44\%) have challenge sets as test sets. %
\citet{arora-etal-2021-types} show that challenge sets represent an OOD type not fully captured by changes in background features (e.g., genre) or semantic features (e.g., unseen classes). %
We echo their recommendation to explicitly state the targeted OOD type; challenge sets may not reflect typical domain shifts, like genre change, and broader OOD concerns may not apply. 

The suitability of splits that are above the dotted line but to the left of the vertical 90\% line is uncertain.
They do not show OOD concerns now, but might as in-domain accuracy improves. %
The few-shot setting with Mistral-7B exemplifies this, with similarly low in-domain and OOD accuracies (Tables \ref{tab:appendix_qa_splits}--\ref{tab:appendix_qa_splits_mistral}, Appendix). %

\paragraph{Q2: Are challenge sets still ``stressful''?}

To answer this question, we consider models that are at least 85\% accurate in-domain: these highly accurate models are candidates for deployment, and require stress-testing. Table~\ref{tab:challenge_set_results} shows the best accuracy on each challenge set (and their subparts) among these models, and also the difference between that model's in-domain and challenge set accuracy. %
In contrast to sentiment classifiers, we observe that NLI, paraphrase identification, and reading comprehension models, finetuned or few-shot, are not robust to many of their challenge sets. The discussion below focuses on NLI. 

For models finetuned on MNLI, QNLI, or SNLI, the following datasets are still challenging: ANLI, HANS, SNLI CAD, SNLI-hard, and some of the tests in the NLI Diagnostic and Stress Test collections.  %
None of these models reach 85\% accuracy on challenge sets, except SNLI-hard, which shows notable disparity between its in-domain and challenge set accuracies. %
Table \ref{tab:nli_diagnostics_stress_test} (Appendix) provides a breakdown of NLI Diagnostics and Stress Test results. %
We also observe that when a challenge set ceases to be ``stressful'', various model types across different sizes are robust. %
Figure \ref{fig:challenge_main} (Appendix) illustrates this with ``Breaking NLI''. %

Popular NLI datasets have several issues~\cite{bowman-dahl-2021-will}, to the point that they are commonly used to study data shortcuts~\cite[e.g.,][]{ross-etal-2022-self,wu-etal-2022-generating}. %
Why should we expect  models trained on them be robust? 
To study the robustness of models trained on higher quality data, we train T5-11B on the WANLI dataset~\cite{liu-etal-2022-wanli}, with an in-domain accuracy of 78.3\%. %
Table \ref{tab:challenge_set_t5_11b_wanli} (Appendix) reports its challenge set accuracies. %
Ignoring the above criterion about the model's readiness for stress testing, we analyze its effectiveness under stress testing in terms of the difference between its in-domain and challenge set accuracies. %
We see that it is more robust on several challenge sets than models trained on previous NLI datasets. %
Specifically, it does not break on any HANS partition, SNLI CAD, or SNLI-hard. %
Training on higher-quality data appears to be a promising approach towards robust NLI and future work should focus on stress testing such NLI models.   

In the few-shot setting, Mistral-7B achieves in-domain NLI accuracies $<$70\% and reading comprehension F1 $<$85\%; neither meet our criterion for stress testing. 
Table~\ref{tab:icl_challenge_set_results} (Appendix) compares its in-domain and challenge set accuracies.
Most NLI challenge sets result in failures. %

\begin{figure*}[t]
  \centering
  \begin{subfigure}{\textwidth}
    \includegraphics[width=\textwidth]{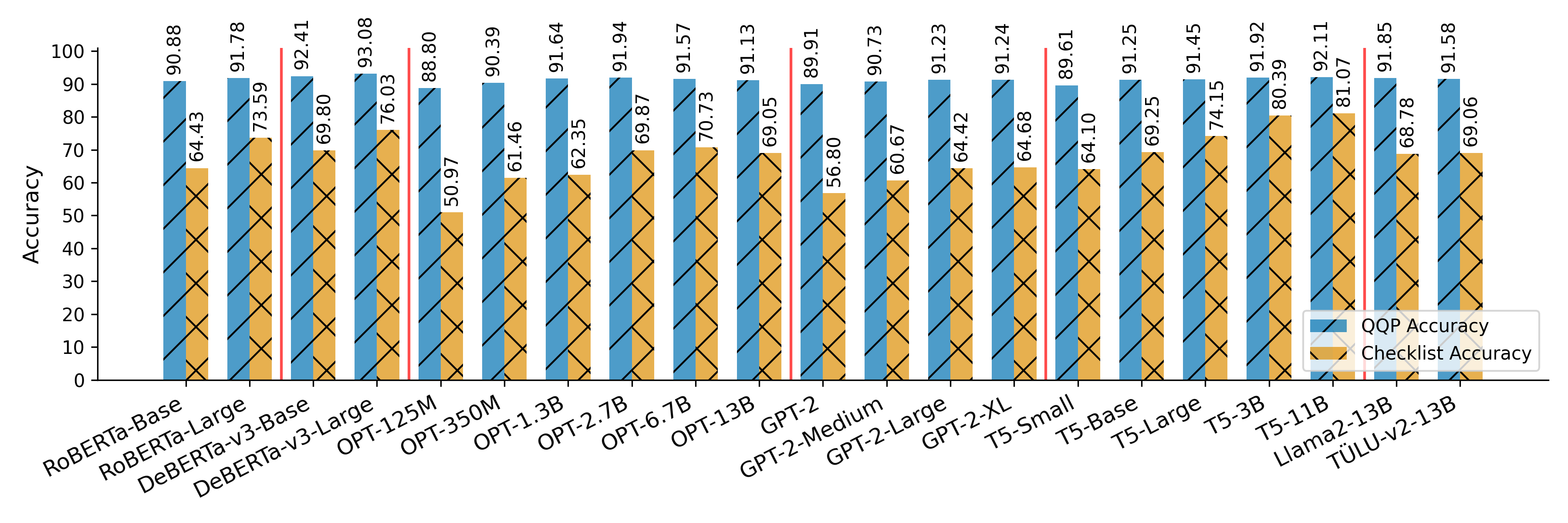}
    \caption{QQP (duplicate questions identification)}
    \label{fig:qqp_vs_checklist_accuracy}
    \end{subfigure}
    \begin{subfigure}{\textwidth}
    \centering
    \includegraphics[width=\textwidth]{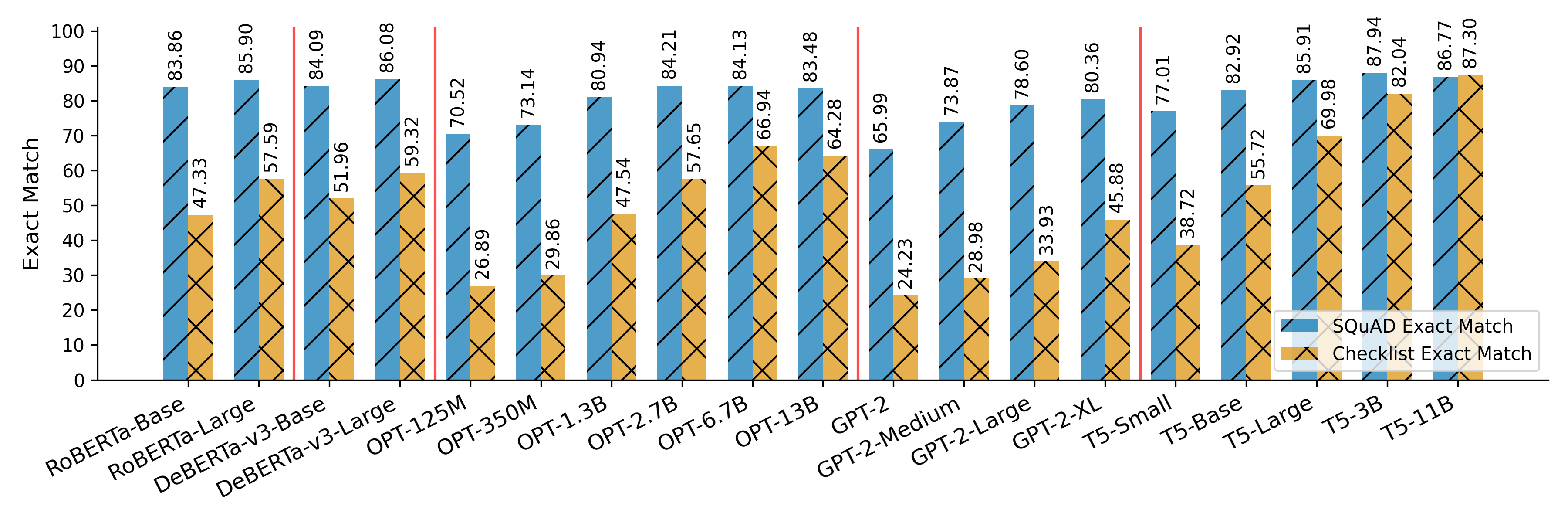}
    \caption{SQuAD} %
    \label{fig:squad_vs_checklist_accuracy}
    \end{subfigure}
\caption{The task performance on the standard test set vs.\ CheckList performance.}
\label{fig:checklist_main_figure}
\end{figure*}

\section{Highly Accurate Models Still Stumble On The Basics}
\label{sec:checklist}

Even task-specialized models that are accurate on standard datasets may fall short on  basic task-related skills. %
We examine if this holds for models based on different architectures and pretraining objectives, and scaled up to 100$\times$ larger sizes.

\begin{table}[t]
\centering
\resizebox{\columnwidth}{!}{%
\begin{tabular}{llrr}
\toprule
& \textbf{Test} (\textbf{Category}) & \textbf{Acc/F1} & \textbf{Diff} \\
\midrule
\multirow{4.5}{*}{\rotatebox{90}{\small \emph{Sentiment Class.}}}  & C-IMDB & 94.5 & 2.1 \\
\arrayrulecolor{black!20}\cmidrule{2-4}
& IMDB Contrast (all)  & 99.6 & -4.3\\
& IMDB Contrast  (contrast) & 95.7 & 0.3\\
& IMDB Contrast  (original) & 99.6 & -4.3 \\
\arrayrulecolor{black}\midrule
\multirow{14}{*}{\rotatebox{90}{\small \emph{Natural Language Inference}}}  & \cellcolor{mycolor} ANLI (r1) & \cellcolor{mycolor} 66.0 & \cellcolor{mycolor} 26.1\\
& \cellcolor{mycolor} ANLI (r2) & \cellcolor{mycolor} 53.5 & \cellcolor{mycolor} 38.6\\
& \cellcolor{mycolor} ANLI (r3) & \cellcolor{mycolor} 49.6 & \cellcolor{mycolor} 42.5\\
\arrayrulecolor{black!20}\cmidrule{2-4}
& Breaking NLI  & 97.9 & -6.7 \\
\arrayrulecolor{black!20}\cmidrule{2-4}
& HANS (all)& 98.9 & -6.8\\
& \cellcolor{mycolor} HANS (constituent) & \cellcolor{mycolor} 71.2 & \cellcolor{mycolor} 21.1\\
& HANS (lexical overlap) & 98.9 & -6.8 \\
& \cellcolor{mycolor} HANS (subsequence) & \cellcolor{mycolor} 70.3 & \cellcolor{mycolor} 21.9\\
\arrayrulecolor{black!20}\cmidrule{2-4}
& MNLI-hard (val matched) & 88.4 & 3.0\\
& MNLI-hard (val mismatched) & 88.2 & 3.3\\
\arrayrulecolor{black!20}\cmidrule{2-4}
& \cellcolor{mycolor} NLI Diagnostics (min--max) & \cellcolor{mycolor} 30.0--95.0 & \cellcolor{mycolor} 61.5 / -3.5\\
\arrayrulecolor{black!20}\cmidrule{2-4}
&  \cellcolor{mycolor} Stress Test (min--max) & \cellcolor{mycolor} 76.8--90.6 & \cellcolor{mycolor} 14.4 / 0.9 \\
\arrayrulecolor{black!20}\cmidrule{2-4}
& \cellcolor{mycolor} SNLI CAD  & \cellcolor{mycolor} 82.9 & \cellcolor{mycolor} 9.2\\
\arrayrulecolor{black!20}\cmidrule{2-4}
&  \cellcolor{mycolor} SNLI-hard & \cellcolor{mycolor} 85.5 & \cellcolor{mycolor} \underline{6.6}\\
\arrayrulecolor{black}\midrule
& \cellcolor{mycolor} PAWS-QQP  & \cellcolor{mycolor} 50.9 & \cellcolor{mycolor} 42.2\\
\arrayrulecolor{black}\midrule
\multirow{11}{*}{\rotatebox{90}{\small \emph{Reading Comprehension}}} & AddOneSent &	85.0 & \underline{8.8} \\
&  \cellcolor{mycolor}AddSent &	\cellcolor{mycolor} 84.1	& \cellcolor{mycolor} \underline{9.7}\\ 
&  \cellcolor{mycolor} Adversarial Paraphrased &	\cellcolor{mycolor} 85.9	& \cellcolor{mycolor} \underline{8.1}\\ 
& \cellcolor{mycolor} BoolQ CAD &	\cellcolor{mycolor} 78.1	& \cellcolor{mycolor} 11.1 \\
& \cellcolor{mycolor} BoolQ Contrast Set &	\cellcolor{mycolor} 79.3	& \cellcolor{mycolor} 9.9 \\
& \cellcolor{mycolor} MultiRC &	\cellcolor{mycolor} 71.8	& \cellcolor{mycolor} 14.3 \\
& \cellcolor{mycolor} NaturalQuestions &	\cellcolor{mycolor} 66.5	& \cellcolor{mycolor} 26.4 \\
& \cellcolor{mycolor} NewsQA &	\cellcolor{mycolor} 66.6	&\cellcolor{mycolor}  24.9\\ 
& Non-Adversarial Paraphrased &	92.6	& 1.4 \\
& \cellcolor{mycolor} Quoref &	\cellcolor{mycolor} 64.1	& \cellcolor{mycolor} 27.7 \\
& SQuAD-hard &	92.5	& 1.3 \\ 
\arrayrulecolor{black}\bottomrule
\end{tabular}
}
\caption{The max.\ challenge set performance for models with 85+\% in-domain accuracy (classification) or F1 (reading comprehension). \textbf{Diff} shows the gap between in-domain and challenge set results (higher means poorer generalization). Shaded rows mark datasets that remain challenging for associated models.}
\label{tab:challenge_set_results}
\end{table}

\paragraph{Background.} The CheckList~\cite{ribeiro-etal-2020-beyond} methodology helps test whether models have capabilities that are expected for a given task. %
\citeauthor{ribeiro-etal-2020-beyond} suggest considering the following capabilities: ``Vocabulary+POS (important words or word types for the task), Taxonomy (synonyms, antonyms, etc), Robustness (to typos, irrelevant changes, etc), NER (appropriately understanding named entities), Fairness, Temporal (understanding order of events), Negation, Coreference, Semantic Role Labeling (understanding roles such as agent, object, etc), and Logic (ability to handle
symmetry, consistency, and conjunctions).''
To categorize potential capability failures, they introduce three test types: 
\begin{inparaenum}[(i)]
\item minimum functionality tests (MFTs), 
\item invariance tests (INVs), and 
\item directional expectation tests (DIRs). %
\end{inparaenum}
MFTs check that a model works on simple examples, akin to unit testing in software engineering. %
INVs confirm that minor label-preserving input changes do not change model predictions. %
If such modifications do alter labels, DIRs validate that the model predictions also change. %
Tables \ref{tab:checklist_tests_1_to_32}--\ref{tab:checklist_tests_33_to_53} show examples of 53 tests for the task of identifying duplicate questions. %
\citeauthor{ribeiro-etal-2020-beyond}'s models achieve accuracy above 90, but CheckList reveals that these seemingly accurate models often lack key capabilities.

\paragraph{Q1: Are we at a stage where accurate models meet the expectations for their capabilities?} No. 
Comparing task accuracies with CheckList accuracies of 19 models finetuned for QQP and SQuAD in Figure~\ref{fig:checklist_main_figure} %
reveals notable discrepancies. QQP accuracies are consistently high across models ($>$89\%), but CheckList accuracies vary and are substantially lower, even as low as 51\% (OPT-125M). %
It is reasonable to expect that these seemingly performant models will excel at relatively simple CheckList tests. Yet QQP models achieve  $>$95\% accuracy for fewer than 45\% of the tests; see Table \ref{tab:checklist_summary}, row 1 (Appendix). %
The same table also shows that for many tests (15--35\% of all tests), the accuracy is lower than 60\%. %
Moreover, no QQP model achieves $>$60\% accuracy on tests \texttt{\{6,12,23,26,40,42,48\}} that span 5 capabilities.   

\paragraph{Q2: Are larger models more capable?} Generally yes, but with limits and irregularities. Checklist improvements across model sizes for each model group (separated by vertical lines in Figure~\ref{fig:checklist_main_figure}) level off suboptimally. %

On specific QQP tests, accuracy does not always monotonically improve with model size (Figures~\ref{fig:checklist_figures_roberta}--\ref{fig:checklist_figures_t5}, Appendix). %
A 10\% accuracy drop occurs for several tests when scaling from one model size to another (Table \ref{tab:checklist_summary}, row 6). %
Also, a substantial fraction of tests where a model version achieves 95+\% accuracy is when the model accuracy is flat across sizes, not where scaling helps  (rows 2, 3). %

This does not negate the advantages of scaling altogether. For example, T5-3B/11B have  better checklist accuracies than their smaller variants. %
Table \ref{tab:checklist_summary} also reports the fraction of tests where the accuracy of the smallest and largest model versions differ by 10\% or more, without notable drops during scaling. %
To check if T5-11B's performance is surpassed by newer models, we also finetune T\"uLU-2-13B \cite{ivison2023camels} and LLaMA-2-13B \cite{touvron2023llama} on QQP; we do not observe improvements. 
In summary, scaling helps but is not a holistic solution, and how to finetune specialized models that have the necessary skills to do a given task robustly remains open.

\begin{figure*}[!ht]
    \centering
    \includegraphics[width=\textwidth, trim={0 0.3cm 0 0.3cm},clip]{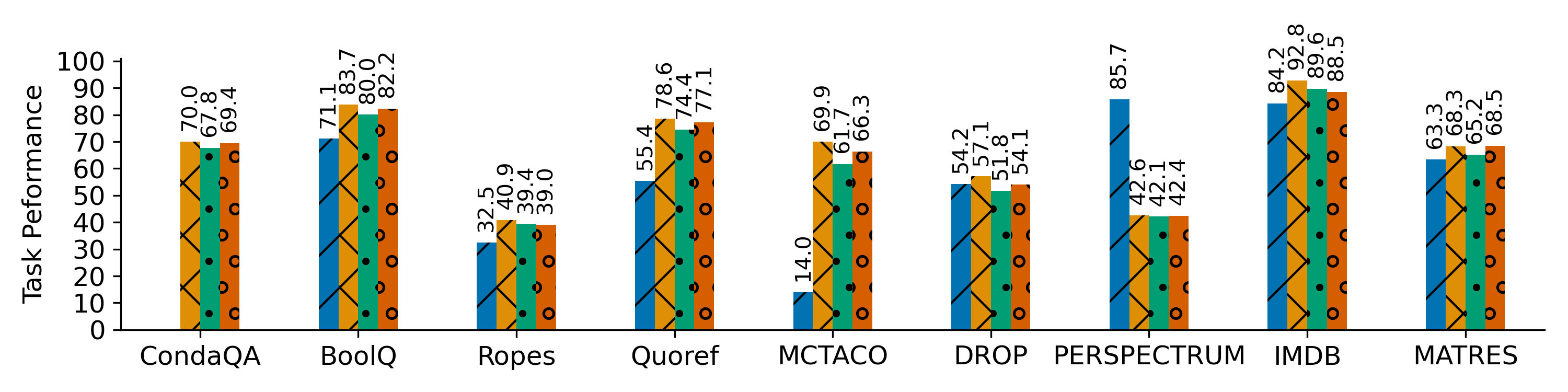}
    \includegraphics[width=\textwidth, trim={0 0.3cm 0 0.3cm},clip]{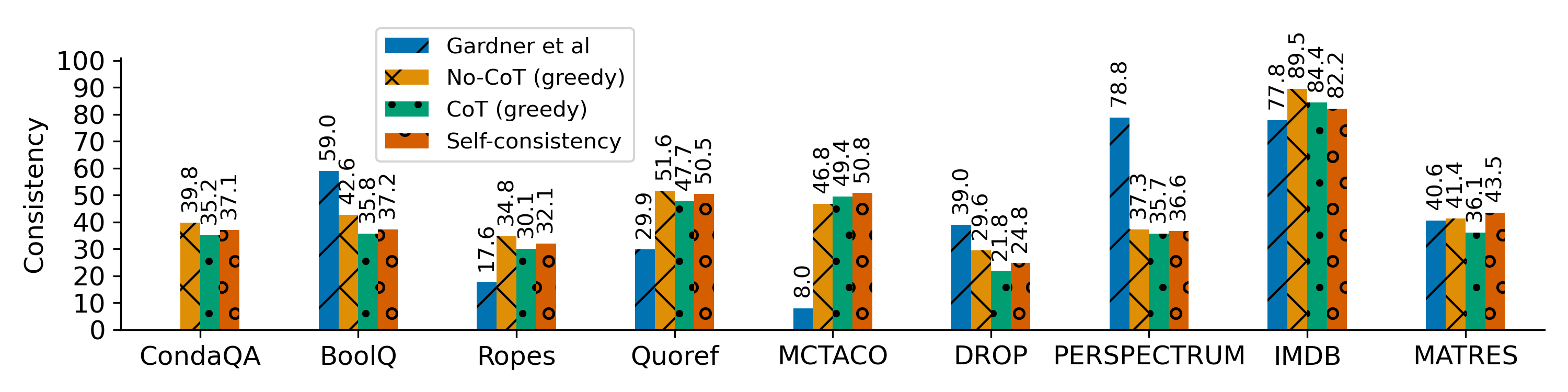}
    \caption{Flan-T5-11B performance with standard measures (accuracy, F1, token-F1) vs.\ contrast set consistency. The model's instruction finetuning data includes training data for all tasks except CondaQA. Prompts include an instruction, 8 examples, and optionally explanations for chain-of-thought prompting and self-consistency decoding.} 
\label{fig:contrast_sets_results}
\end{figure*}

\section{Better Evaluation Paradigms Exist}
\label{sec:contrast_sets}

Evaluating with sets of mutually dependent examples can uncover model fragility. %
Yet, models'  continue to be assessed by their performance on benchmarks such as MMLU~\cite{DBLP:conf/iclr/HendrycksBBZMSS21} that consist solely of i.i.d.\ instances. %
Are evaluations that go beyond the i.i.d.\ assumption %
unnecessary, or simply overlooked?

\paragraph{Background.} Motivated by the ongoing challenge of creating artifact-free NLP benchmarks, \citet{gardner-etal-2020-evaluating} propose to evaluate with sets of examples that are minimally different from each other. %
They report \emph{contrast set consistency}, a measure of how often a model correctly addresses every example in a set. %
Models that succeed on standard benchmarks by exploiting data shortcuts can do this rarely. %
\citet{gardner-etal-2020-evaluating} create contrast sets for 10 datasets, one of which is multimodal. %
\citet{ravichander-etal-2022-condaqa} introduce CondaQA, another contrastive dataset. %

\paragraph{Experimental Setup.} Instead of training 10 (unimodal datasets) $\times$ 19 (models) $\times$ 3 (seeds) = 570 models, we reexamine  contrast set evaluation with Flan-T5-11B  \cite{chung2022scaling} whose instruction finetuning includes training data of all datasets in question except CondaQA. %
We prompt Flan-T5 with an instruction and 8 demonstrations (Tables ~\ref{tab:cot-contrast1}--\ref{tab:nocot-contrast4}, Appendix). %
Since Flan models are instruction finetuned with chain-of-thought (CoT) prompting \cite{DBLP:conf/nips/Wei0SBIXCLZ22} and self-consistency \cite{DBLP:conf/iclr/0002WSLCNCZ23}, we also try these reasoning-boosting addons to get the highest consistency possible.\footnote{Explanations are written by one of the authors.}

\paragraph{Q1: Is there still a gap between original test sets and their contrastive counterparts?} Yes, Figure~\ref{fig:contrast_sets_results} shows that Flan-T5-11B's performance on only the original instances in contrast sets is much higher than its consistency for each contrast set.\footnote{We do not report UD Parsing results. We do not get a reasonable performance with prompting.} 
Surprisingly, the benefits of including explanations are negligible\,---\,relative to the standard greedy decoding, chain-of-thoughts do not improve consistency in a single case, and self-consistency improves it only for MCTACO and MATRES. %

\citet{gardner-etal-2020-evaluating} demonstrate the feasibility of creating a high-quality contrast set with 1K examples from an existing dataset in just a week's work by an expert. %
Given the confirmed benefits of contrast datasets in robust reasoning evaluation, the development of contrastive versions of popular benchmarks such as BBH \cite{suzgun-etal-2023-challenging} and MMLU seems beneficial. %
Mindful of the challenges in evaluating across multiple benchmarks, each with dozens of sub-tasks, we deem it is more strategic to direct a portion of resources to evaluations that extend beyond benchmarks i.i.d.\ with test sets than having more of the same. 

\paragraph{Q2: Has consistency improved despite gaps?} Only in some cases. The consistency values for Ropes, Quoref, MCTACO, and IMDB show large gains over~\citet{gardner-etal-2020-evaluating}. Yet, they remain low across the board (except for IMDB). %
Moreover, although Flan-T5-11B's standard task performance exceeds the results reported by \citet{gardner-etal-2020-evaluating} across almost all tasks, these improvements do not uniformly translate into increased consistency (e.g., BoolQ and DROP). %
We also experimented with more recent LLMs like Llama-2~\citep{touvron2023llama} and T\"{u}lu-2~\citep{ivison2023camels}, but found them, on average, to be less consistent than Flan-T5 (see Appendix~\ref{subsec:tulu_llama} for results). 
In summary, building models that are consistently robust in local neighborhoods remains a challenge. 

\begin{figure*}[t]
  \centering
  \begin{subfigure}{\textwidth}
    \centering
    \includegraphics[width=\textwidth, trim={0 0.5cm 0 0.3cm},clip]{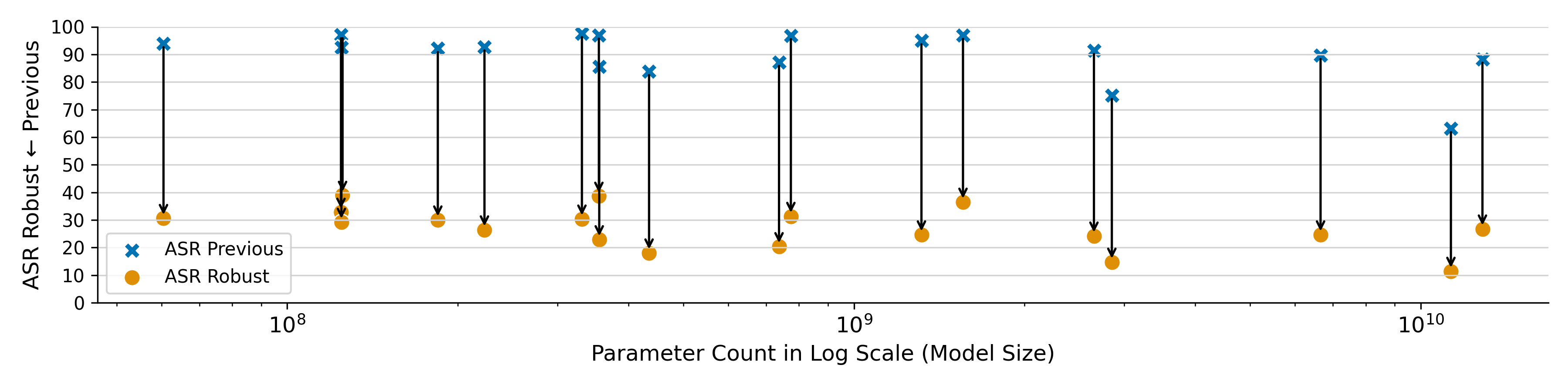}
    \caption{MNLI}
    \label{fig:asr_drop_deepwordbug_mnli}
  \end{subfigure}
  \begin{subfigure}{\textwidth}
    \centering
    \includegraphics[width=\textwidth, trim={0 0.5cm 0 0.3cm},clip]{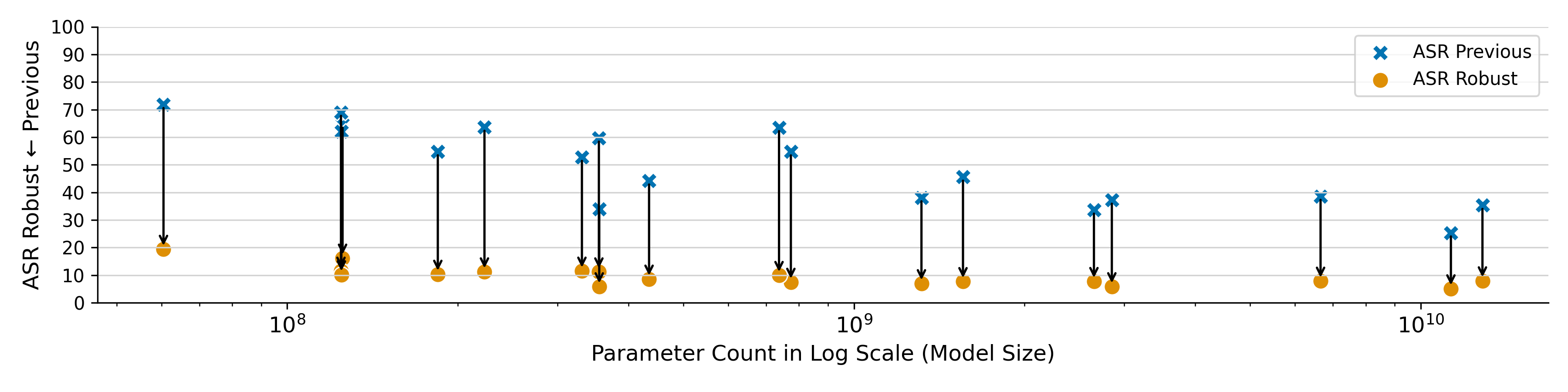}
    \caption{AGNews}
    \label{fig:asr_drop_deepwordbug_agnews}
  \end{subfigure}
  \caption{The change in the attack success rate (ASR) as measured in prior work (\ref{eq:asr_prev}) vs.\ our robust modification (\ref{eq:asr_robust}). \texttt{TextFooler} is used to train the defense and \texttt{DeepWordBug} to fool 19 finetuned models in Table \ref{tab:models}.}
  \label{fig:asr_drop}
\end{figure*}

\section{Evaluating the Evaluators: The True Success of Adversarial Attacks}
\label{sec:adversarial}

Previous sections highlight the robustness and fragility of today's models. Here we bring to light a meta-issue: even some methods for evaluating robustness need improvements to ensure our conclusions are valid.

\paragraph{Background.} Adversarial attacks (prevalent in 2020--2022 NLP papers) make imperceptible changes to task inputs that fool models into making mistakes. %
In computer vision, %
``imperceptibility'' is realized by adding a small noise to the image without altering its appearance~\cite{DBLP:journals/corr/GoodfellowSS14}. %
Applying this idea to text is harder:  nearest neighbors of noisy token embeddings could be tokens that change the original meaning. %
The challenge of adversarially attacking text, studying their potential harm, and designing  countermeasures has been a major research focus.

\paragraph{Attacks \& Defense.} We analyze the following attack methods for NLP models: \texttt{TextFooler} \cite{DBLP:conf/aaai/JinJZS20}, \texttt{BAE} \cite{garg-ramakrishnan-2020-bae}, \texttt{TextBugger} \cite{DBLP:conf/ndss/LiJDLW19}, \texttt{PWWS} \cite{ren-etal-2019-generating}, and \texttt{DeepWordBug} \cite{DBLP:conf/sp/GaoLSQ18}; see Appendix \ref{sec:appendix_attacks} for details. %
All attacks edit until the prediction is altered and find edits in a black-box setting, i.e., model internals are not accessible. %

Following \citet{raina-gales-2022-residue}, we train a binary classifier that distinguishes real from adversarial examples. %
We finetune BERT-base \cite{devlin-etal-2019-bert} with a task's training examples and their adversarial versions generated by \texttt{TextFooler} %
and use this defense 
against all attacks.\footnote{
Since attackers can perturb tokens and/or characters, training 
against a mix of both
could mount a stronger defense.}

\paragraph{Rigorously Determining Success of an Attack.} Attacks are evaluated with \emph{attack success rate}\,---\,a fraction of instances for which an edit that alters the model's correct prediction is found: 
\begin{equation}
    \asr_{\text{prev}} = \frac{1}{|\mathcal{C}|}\displaystyle\sum\limits_{x \in \mathcal{C}} \mathbbm{1}_{\{y_p(\pert(x))\neq y_g(x)\}}
    \label{eq:asr_prev}
\end{equation}
where $\mathcal{C}=\{x \in \mathcal{X}: y_p(x)=y_g(x)\}$, $y_g(\cdot)$ are gold labels, $y_p(\cdot)$ predicted, and $\pert$ is a method that alters  examples. %
However, this metric ignores the well-formedness of attacks and the effectiveness of any defenses against them. %
We enhance the measurement to account for these.

Foremost, $\pert$ should be imperceptible with respect to the gold label, i.e., $y_g(\pert(x))=y_g(x), \forall x \in \mathcal{X}$. %
Moreover, we expect that an AI system has a defense that first detects whether an input is an attack: $\detect(\cdot) \in \{\texttt{real}, \texttt{attack}\}$. %
We redefine attack success rate as:
\begin{align}
\asr_{\text{our}} &= \frac{1}{|\mathcal{C}|}\displaystyle\sum\limits_{x \in \mathcal{C}} \as(x) \label{eq:asr_robust}\\
    \as(x) &= 
    \begin{cases} 1, &   \big(y_p(\pert(x))\neq y_g(x)\big) \land \\
    &\big(\detect(\pert(x))=\texttt{real}\big) \land\\
    & \big(y_g(\pert(x))= y_g(x)\big) \\
    0, &  \text{otherwise}
    \end{cases} \notag
\end{align}

The challenge in properly calculating $\asr$ is ensuring that the assumption that  $y_g(\pert(x))=y_g(x)$ truly holds. %
In our initial analysis of perturbed examples, we found that this is often not fulfilled. %
Since assessing $y_g(\pert(x))$ manually requires recruiting and training annotators, %
we suggest using a highly accurate model for the task. %

$\asr$ trivially becomes zero if  $\detect$ labels everything as an attack.
Thus, $\asr$ must be complemented with the \emph{defense failure rate}, the rate at which the defense marks real examples as attacks: 
\begin{equation}
    \dfr =  \frac{1}{|\mathcal{X}|}\sum\limits_{x \in \mathcal{X}} \mathbbm{1}_{\{\detect(x)=\texttt{attack}\}}
\end{equation}
If the attack success rate or defense failure rate is high, we deem the attacker successful. 
\begin{figure*}[t]
  \centering
  \begin{subfigure}{0.49\textwidth}
    \centering
    \includegraphics[width=\textwidth, trim={0 0 0 1.0cm},clip]{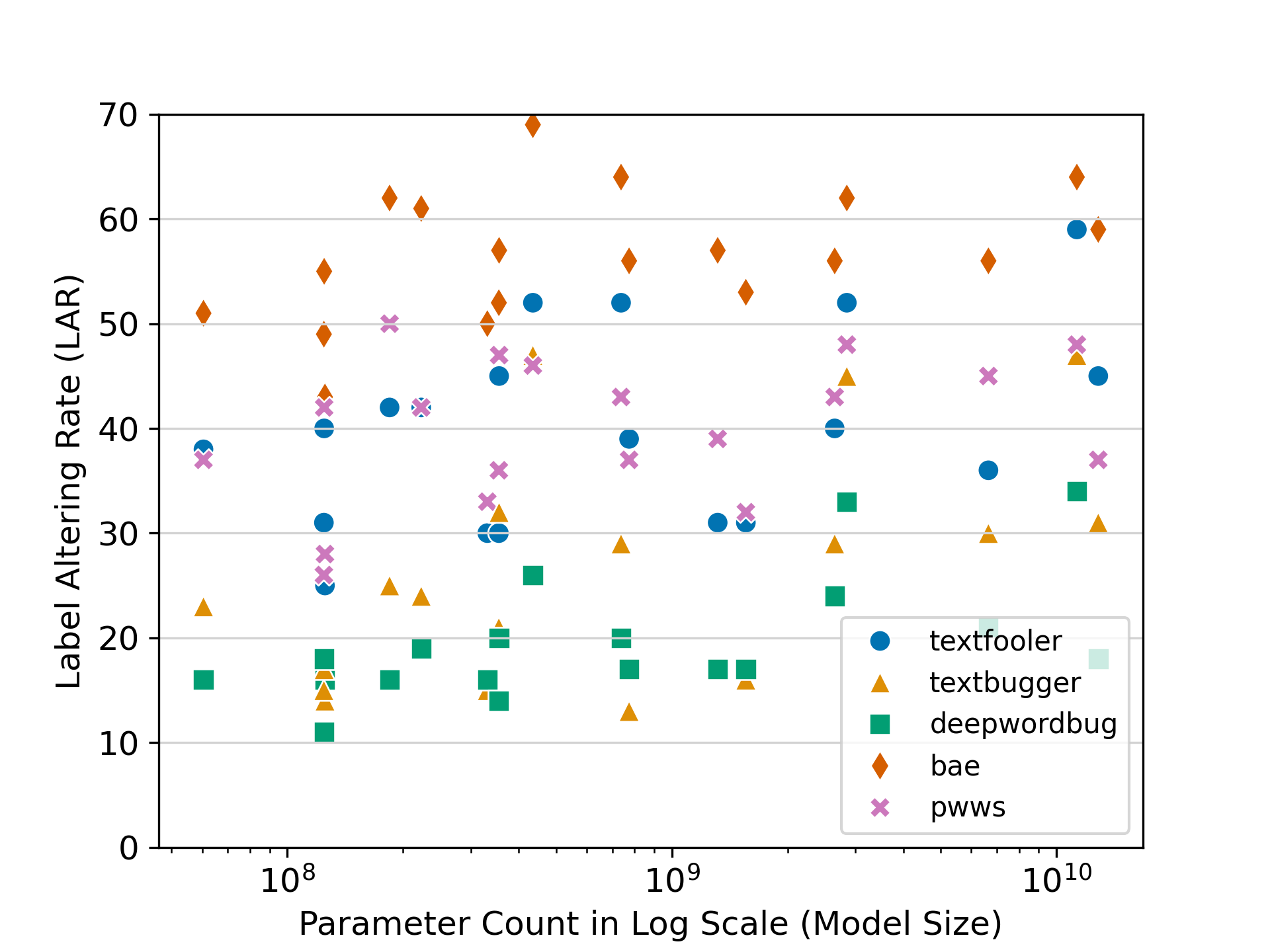}
    \caption{$\lar$}
    \label{fig:lar_lmni}
  \end{subfigure}
  \begin{subfigure}{0.49\textwidth}
    \centering
    \includegraphics[width=\textwidth, trim={0 0 0 1.0cm},clip]{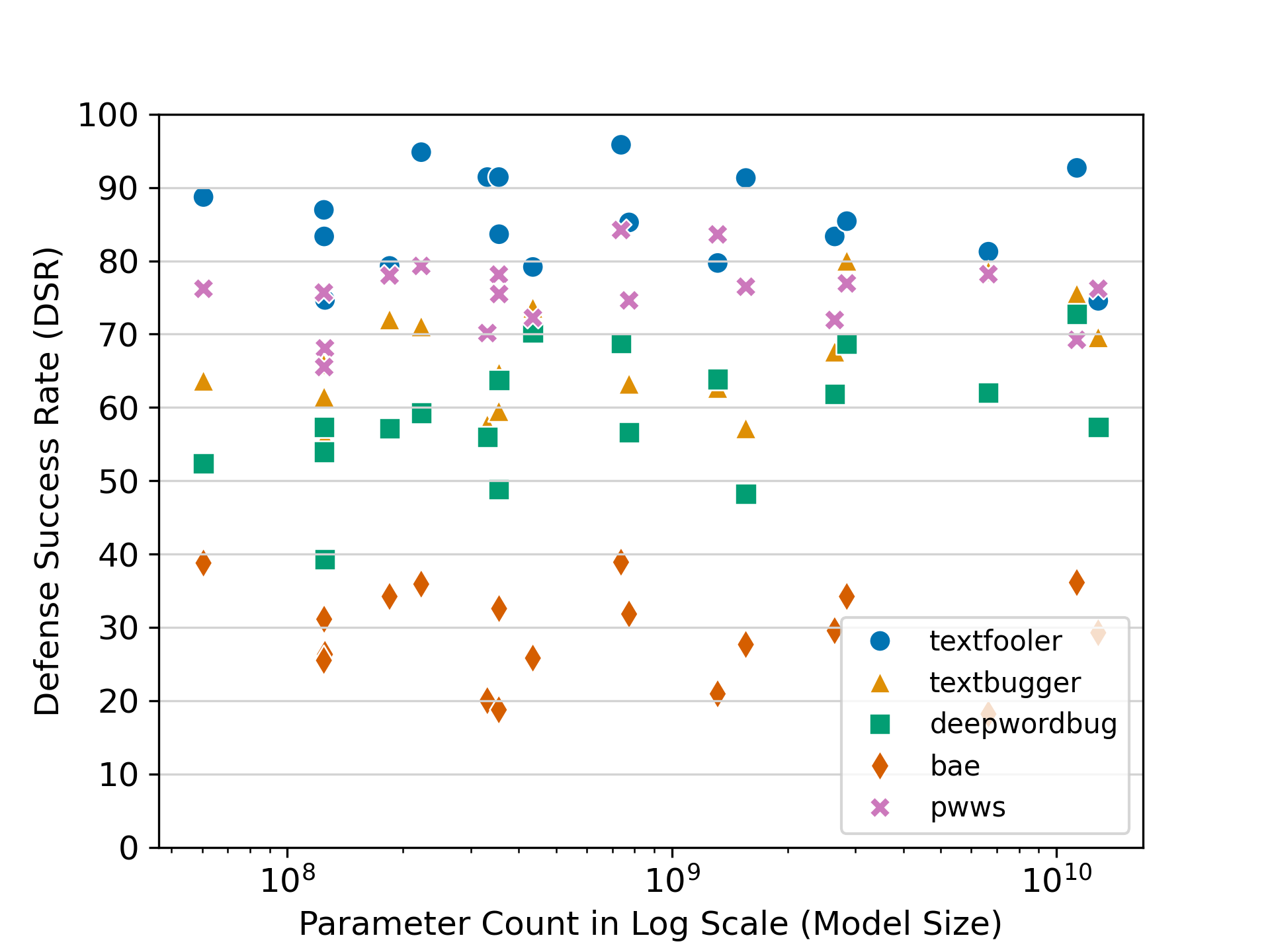}
    \caption{$\dsr$}
    \label{fig:dsr_mnli}
  \end{subfigure}
  \caption{Label altering rate ($\lar$; \ref{eq:lar}) and defense success rate ($\dsr$; \ref{eq:dsr}) obtained in the MNLI setting. Higher values of these measurements contribute to worse effectiveness of attacks, i.e., lower attack success rate.}
  \label{fig:dsr_lar_mnli}
\end{figure*}

\paragraph{Q1: How successful are the attacks?} Less than expected. We test the five attacks to deceive the models in Table~\ref{tab:models} finetuned for AGNews and MNLI. %
The $\dfr$ is only 6.61 on MNLI and 0.83 on AGNews, i.e., the system is still functional.\footnote{We average DFR on the matched and mismatched MNLI %
splits. We use the full val/test MNLI/AGNews data for $\dfr$.} Tables~\ref{tab:agnews-bae}--\ref{tab:mnli-textfooler} (Appendix) show generated adversarial examples. %

We measure $\asr$ on 100 examples sampled from MNLI-matched validation and AGNews test sets. We use GPT-4~\cite{openai2023gpt4} to determine $y_g(\pert(x))$ for these, and estimate that it has an error of 10--20\%; see Appendix \ref{sec:appendix:gpt_analysis} for details.
Figure~\ref{fig:asr_drop} shows the $\asr_{\text{prev}}\to\asr_{\text{our}}$ decline for  \texttt{DeepWordBug}.
The drops for the other four attacks are given in Figures \ref{fig:asr_drop_appendix_mnli}--\ref{fig:asr_drop_appendix_agnews} in the Appendix. %
\texttt{DeepWordBug} modifies characters, not tokens like \texttt{TextFooler}, and  exhibits the smallest decline in effectiveness among attacks for MNLI; but the drop is still substantial. %
It compromises 51.5\% MNLI examples fewer than indicated by the prior $\asr$, and 20.3\% fewer AGNews examples. %
As expected, the drop is more pronounced for \texttt{TextFooler} which was used for training the defense; see \ref{fig:asr_drop_textfooler_mnli} and \ref{fig:asr_drop_textfooler_agnews}. %
For instance, its actual success rate when attacking GPT2-XL (MNLI) falls to just 6\%, a stark contrast to the 94.9\% suggested by the prior $\asr$. %
\paragraph{Q2: Why do attacks rarely succeed?}
To better isolate the factors that lead to lower $\asr$, we report two additional measurements. %
First, \emph{label altering rate}\,---\,a rate at which the true label of the perturbed examples, $y_g(\pert(x))$, is changed: 
\begin{equation}
    \lar = \frac{1}{|\mathcal{C}|}\sum\limits_{x \in \mathcal{C}} \mathbbm{1}_{\{y_g(\pert(x)) \neq y_g(x)\}} \label{eq:lar}
\end{equation} 
Second, \emph{defense success rate}\,---\,a rate at which the defense detects well-formed attacks: %
\begin{align}
    &\dsr = \frac{1}{|\mathcal{C}|}\quad\sum\limits_{\mathclap{\substack{x \in \mathcal{C}\\ \attack(\pert(x))=1}}}\quad \mathbbm{1}_{\{\detect(\pert(x))=\texttt{attack}\}}\label{eq:dsr}\\
    &\attack(\pert(x)) =
    \begin{cases}
    1, & (y_g(\pert(x)) = y_g(x)) \land\\
       & (y_p(\pert(x)) \neq y_g(x))\\
    0, & \text{otherwise}
    \end{cases}\notag
\end{align} 
Attackers cannot fool models on original examples they cannot handle correctly, so $\dsr$ is checked only for examples in $\mathcal{C}$. Note that $\dsr$ and $\dfr$ do not sum to 100.\footnote{Another measurement to consider is the rate at which the true label of the perturbed examples is unchanged, but the attacker does not fool the model into making a wrong prediction. However, given that $\asr_{\text{prev}}$ is high, we know that this rate is low and thus does not explain the drop in $\asr$.}

Figure \ref{fig:dsr_lar_mnli} and Figure \ref{fig:dsr_lar_agnews} (Appendix) show $\lar$ and $\dsr$ for MNLI and AGNews respectively. %
The $\lar$ and $\dsr$ together underscore the need for multiple criteria to be met for an attack to be truly successful. %
The average $\lar$ across 19 models for 3/5 attacks is 40\%, even higher for \texttt{BAE}. %
These findings show that assuming labels remain unchanged under perturbations is not justified.
Although \texttt{DeepWordBug} exhibits a lower rate of label change, its robust ASR measurement is also low.
This is because a defense trained with \texttt{TextFooler}'s perturbations also detects 40--84\% of attacks that were not part of its training; except \texttt{BAE}'s, which manages to bypass the defense more effectively.
AGNews is less affected by label changes: $\lar$ for AGNews ranges from 11\% to 32\%. %
However, the defense is effective: across models, it catches over 75\% of perturbations 
from all attacks except  \texttt{BAE}'s. %

\section{Related Work}

Several efforts address the question of NLP robustness through %
surveys~\citep{wang-etal-2022-measure,Hupkes2023}, new benchmarks~\citep{ye-etal-2021-crossfit,yang-etal-2023-glue,yuan2023revisiting}, toolkits~\citep{goel-etal-2021-robustness}, etc.
Unlike studies like GLUE-X~\citep{yang-etal-2023-glue}, and BOSS~\citep{yuan2023revisiting} which propose new benchmarks for %
OOD assessment, our focus is not on new benchmarks. Instead, %
we identify OOD data splits that are no longer challenging. 
Our results, detailed in~\sect{sec:setup}, reveal that NLI models trained on MNLI demonstrate robust generalization to examples from MNLI-mismatched, and SNLI, but struggle with those from DNLI. This suggests that future evaluations should %
use this %
challenging train-test split. Notably, GLUE-X includes the two datasets already addressed for OOD evaluation but excludes DNLI.

HELM~\citep{liang2022holistic} advocates evaluating NLP models across dimensions like fairness, bias, robustness, and efficiency, emphasizing breadth and leaving room for more detailed exploration. %
For instance, its robustness assessments only involve small, semantics-preserving automatic transformations. In contrast, our evaluation is more comprehensive, encompassing domain generalization (\sect{sec:setup}), behavioral testing through Checklists (\sect{sec:checklist}), consistency evaluations (via contrast sets in \sect{sec:contrast_sets}), and adversarial robustness (\sect{sec:adversarial}). Additionally, while~\citet{awadalla-etal-2022-exploring} explore distributional robustness for QA models trained on SQuAD~\citep{rajpurkar-etal-2016-squad}, our analysis extends further, encompassing two diverse QA tasks: Reading Comprehension on SQuAD, NewsQA~\citep{trischler-etal-2017-newsqa}, and NaturalQuestions~\citep{kwiatkowski-etal-2019-natural}, as well as %
BoolQ~\citep{clark-etal-2019-boolq}.

We are not the first to discuss challenges with adversarial attack evaluations in NLP. %
For instance, \citet{morris-etal-2020-reevaluating} suggest additional constraints for filtering out adversarial examples. %
Our \sect{sec:adversarial} proposal establishes a new protocol for systematically evaluating adversarial attacks. 

\section{Conclusions}
We thoroughly investigate robustness of finetuned models up to 13B parameters from four perspectives: generalization under distributional shifts, checklist-based behavioral analysis, contrast set consistency, and robustness to adversarial attacks. Our results collectively show that the current state of NLP robustness is multifaceted:  

\begin{compactitem}
\item We identify experimental setups, such as certain train-OOD test splits and challenge set evaluations of finetuned sentiment classifiers (\sect{sec:setup}), that seem largely resolved. We caution against robustness research that does not advance beyond these areas. 
\item Our findings illustrate progress post-BERT. In \sect{sec:checklist}, we show that larger models generally exhibit more basic necessary task skills, and in \sect{sec:contrast_sets}, that contrast sets consistency has improved since their inception in 2020. We caution against robustness research that relies solely on BERT-era models as baselines.
\item Despite the progress, larger models are still not flawless. How to finetune specialized models that have the necessary skills to do a given task, that are robust in local neighborhoods, when stress tested, and in the presence of certain distribution shifts, remains open. 
\item With adversarial attacks (\sect{sec:adversarial}), we demonstrate that established and popular methods for evaluating robustness need improvements to ensure our conclusions are valid.  
\end{compactitem}

\section{Limitations}

Although we present an extensive analysis, we still focus on the robustness of models that can be finetuned with moderate computing resources for free. %
Therefore, we do not cover extremely large models, like those with hundreds of billions of parameters as PaLM \cite{DBLP:journals/jmlr/ChowdheryNDBMRBCSGSSTMRBTSPRDHPBAI23}, or proprietary models finetuned with paid services like \href{https://platform.openai.com/docs/guides/fine-tuning}{OpenAI's}. %
Additionally, despite extending our analysis beyond typical classification tasks, we note that the robustness for many other less-explored tasks remains largely unexamined. %

 Prior work on adversarial attacks reports additional metrics: (i) the average percentage of original tokens/characters that are edited, and (ii) the average number of queries per sample. %
 Smaller values of these two metrics mean that the edit of the original instance is small (i.e., the edit is ``imperceptible'') and that the attack is efficient. %
 Another condition that requires that $x$ and $\pert(x)$ are paraphrases and/or that the edit size is minimal can be included in Eq (\ref{eq:asr_robust}). %
 The former is easier to verify with GPT-4 because there is no universal threshold for the edit size. %
 With this requirement, we expect $\asr$ to reduce even more. 

Finally, although we address the goal of illuminating where the field stands in terms of established evaluations, we recognize there is an opportunity to expand our analysis into a benchmark or to collaborate with existing ones.

\section*{Acknowledgements}

We thank the anonymous reviewers and the metareviewer for their helpful feedback. We also thank Kyle Lo, Luca Soldaini, and members of the UtahNLP group for valuable insights, Luca Soldaini for retrieving ACL publications for us, and users of the CHPC cluster who were patient with us. The support and resources from the Center for High Performance Computing at the University of Utah are gratefully acknowledged.  Ashim Gupta is supported by the Bloomberg Data Science Ph.D. Fellowship. This material is based in part upon work supported by the National Science Foundation under Grants \#2007398 and \#2217154.

\bibliography{anthology,custom}

\appendix

\section{Additional Experimental Details}
\label{app:sec:experimental}
We finetune models across five model families with all the available sizes among them, giving us a total of 19 models per task. For models that are too big to train on a single GPU, we utilize DeepSpeed to perform multi-GPU training~\citep{rajbhandari2020zero}. For all the models except those that require DeepSpeed, we train three models with different random seeds (seeds 1, 2, and 3). In total, we have 51 models for each task, not counting models we use for in-context learning analyses.

\subsection{Training Details}

\paragraph{Learning Rate Search.} During preliminary experiments, we observed significant variance in task performance across different random seeds when using the default learning rates. We perform a learning rate search across six learning rates to select the best learning rate: \{1e-4, 5e-4, 1e-5, 5e-5, 1e-6, 5e-6\}. To reduce the time required for learning rate search, we train each model to 1000 training steps and select the learning rate with the lowest training loss. We found this strategy works well and stabilizes training across all models we used. 

\paragraph{Other Hyperparameters.} All classification models are trained with a batch size of 32 and the QA models are trained with a batch size of 8 and are determined based on the availability of the GPU resources. Wherever necessary, we employ gradient accumulation with multiple GPUs to emulate these batch sizes. 
All QA models are trained with the sequence length of 512 and a document stride of 128. The maximum sequence lengths for classification tasks are dependent on each task as they vary in terms of the size of input text. All NLI and paraphrase identification models are trained with a sequence length of 256, while sentiment classification claim verification models, and topic classification (on AGNews) models are all trained with a sequence length of 512. During our preliminary experiments, we find that training for three epochs was sufficient and therefore we train all models for three epochs. 

\paragraph{Toolkits.} We train all our models using the Transformers library from Huggingface~\citep{wolf-etal-2020-transformers} with the PyTorch backend~\citep{paszke2019pytorch}. For evaluating adversarial attacks, we use the TextAttack~\citep{morris-etal-2020-textattack} library. At the time of evaluation, the library did not support attacking text-to-text models like the T5 model for the classification tasks and is therefore implemented by ourselves. %
For few-shot in-context evaluations, we compared the popular \href{https://github.com/EleutherAI/lm-evaluation-harness/}{lm-evaluation-harness}~\citep{eval-harness} with \href{https://github.com/mosaicml/llm-foundry/}{llm-foundry} and found \texttt{lm-evaluation-harness} to generally work better in reproducing the reported results. 

\paragraph{QA Models: Span Classification vs.\ Generative.} For question answering, we use the auto-regressive language models (i.e., OPT/GPT) in their generative form instead of span classification~\citep{awadalla-etal-2022-exploring}. 

\subsection{Pre-processing Evaluation Sets}
For some evaluation sets, we did not find any publicly available data splits and therefore construct our own. For out-of-domain evaluation with the QNLI dataset (the task of determining if a sentence answers a question or not),~\citet{swayamdipta-etal-2020-dataset} use the adversarial SQuAD data from ~\citet{jia-liang-2017-adversarial}. 
Since we did not find a publicly available version of this, we pre-process the data released by~\citet{jia-liang-2017-adversarial}  where we extract the last sentence from each passage along with the question to construct the evaluation instances.
The adversarial instances are marked with the label \texttt{not-entailment}.  

Similarly, the Amazon Reviews dataset has been used in a number of out-of-domain evaluation settings for sentiment classification models.
Different papers use different domains for training and evaluation. Therefore, we sample 10000 examples randomly from six genres (appliances, beauty, fashion, gift cards, magazines, and software).

Additionally, we found that the Twitter paraphrase corpus was not available online and thus we contact authors to get that data.\footnote{\url{https://languagenet.github.io/}} The dataset provides paraphrase ratings on a scale of 1 to 6. We discard the examples with ratings 3 (recommended by authors) and classify those from 1-2 as \texttt{not-paraphrase} and from 4-6 as \texttt{paraphrase}.

The original QuAC dataset~\citep{choi-etal-2018-quac} contains question-answers in a multi-turn dialogue format and is therefore not directly applicable for reading comprehension. We use the converter script provided by~\citet{sen-saffari-2020-models} to convert to the SQuAD format.\footnote{\url{https://github.com/amazon-science/qa-dataset-converter}}

Finally, the MultiRC dataset~\citep{khashabi-etal-2018-looking} which is used for out-of-domain evaluation with BoolQ, we extract only the yes/no questions from the train set as we find the validation set does not have enough of those yes/no questions.

\subsection{Llama-2-7B vs Mistral-7B}
For the in-context learning experiments, we consider two high-performing open-models, Llama-2-7B \citep{touvron2023llama} and Mistral-7B \citep{jiang2023mistral}.
To decide which to use we first compare the performance of the models on a subset of our evaluation sets including MNLI-mismatched, SST, QQP, and SQuAD.
We find that Mistral-7B outperforms Llama-2 across all tasks in both zero-shot and 8-shot settings. 
Additionally, Mistral-7B shows a consistent improvement going from zero-shot to 8-shot settings while Llama-2 shows a drop in accuracy on MNLI-mismatched and a drop in f1-score for QQP. These results lead us to choose Mistral for our experiments.

\section{Robustness to Adversarial Attacks}
\label{sec:appendix:gpt_analysis}
\begin{figure*}[]
  \centering
  \begin{subfigure}{0.49\textwidth}
    \centering
    \includegraphics[width=\textwidth]{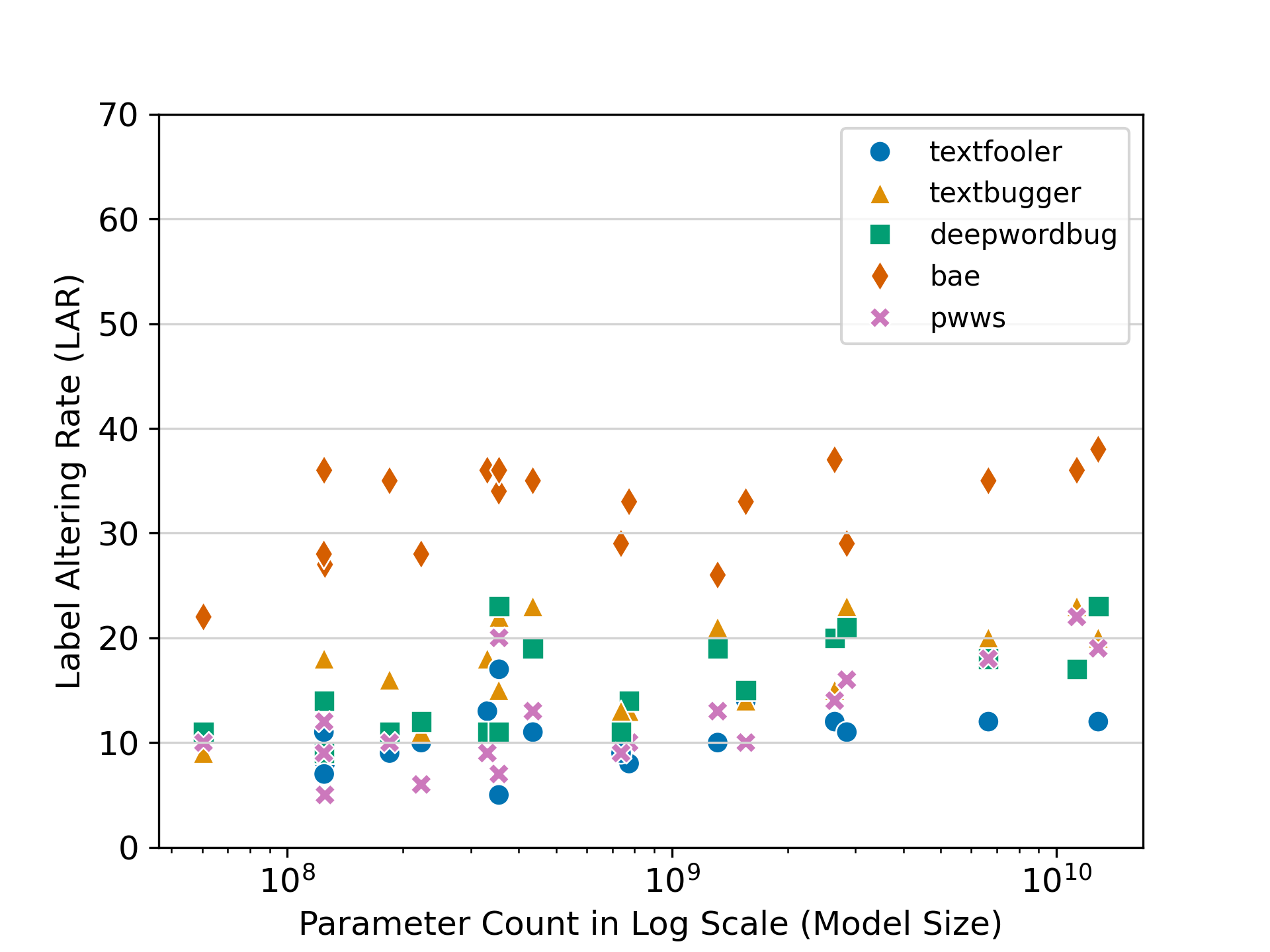}
    \caption{$\lar$}
    \label{fig:lar_lmni}
  \end{subfigure}
  \begin{subfigure}{0.49\textwidth}
    \centering
    \includegraphics[width=\textwidth]{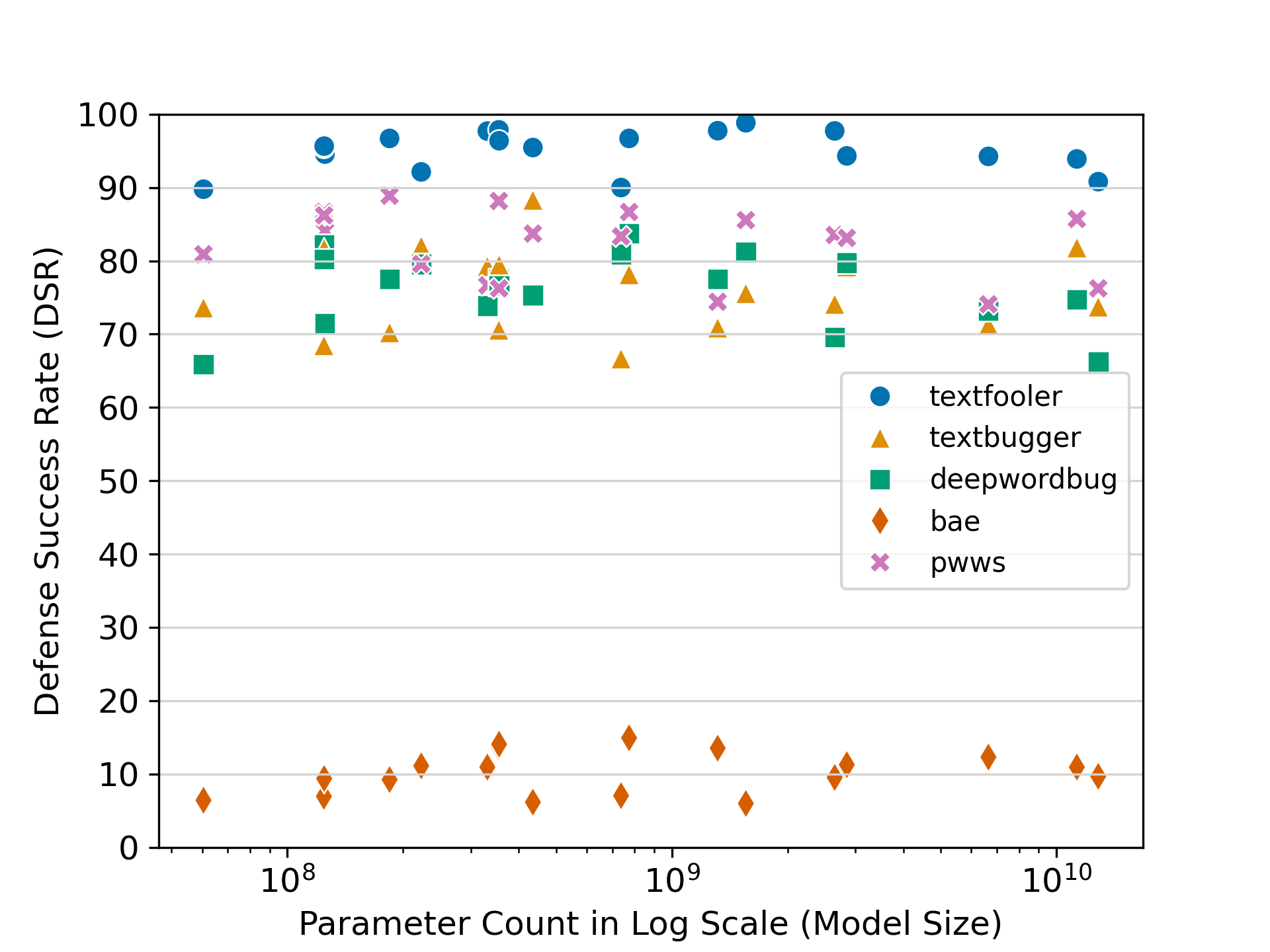}
    \caption{$\dsr$}
    \label{fig:dsr_agnews}
  \end{subfigure}
  \caption{Label altering rate ($\lar$; \ref{eq:lar}) and defense success rate ($\dsr$; \ref{eq:dsr}) obtained in the \textbf{AGNews} setting. Higher values of these measurements contribute to worse effectiveness of attacks, i.e., lower attack success rate.}
  \label{fig:dsr_lar_agnews}
\end{figure*}

\begin{figure*}[t]
  \centering
  \begin{subfigure}{\textwidth}
    \centering
    \includegraphics[width=\textwidth]{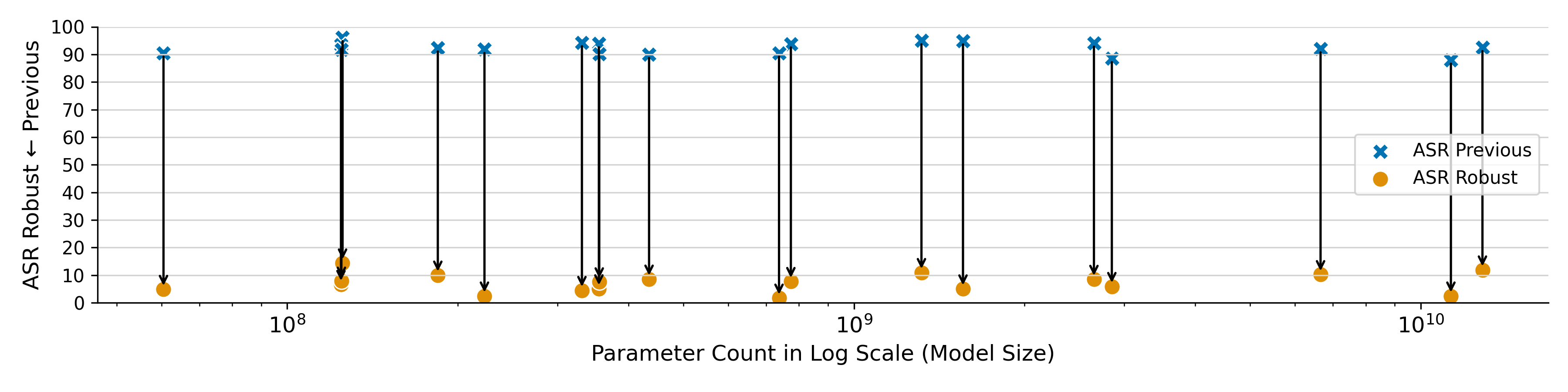}
    \caption{\texttt{TextFooler} used to fool the model.}
    \label{fig:asr_drop_textfooler_mnli}
  \end{subfigure}
  \begin{subfigure}{\textwidth}
    \centering
    \includegraphics[width=\textwidth]{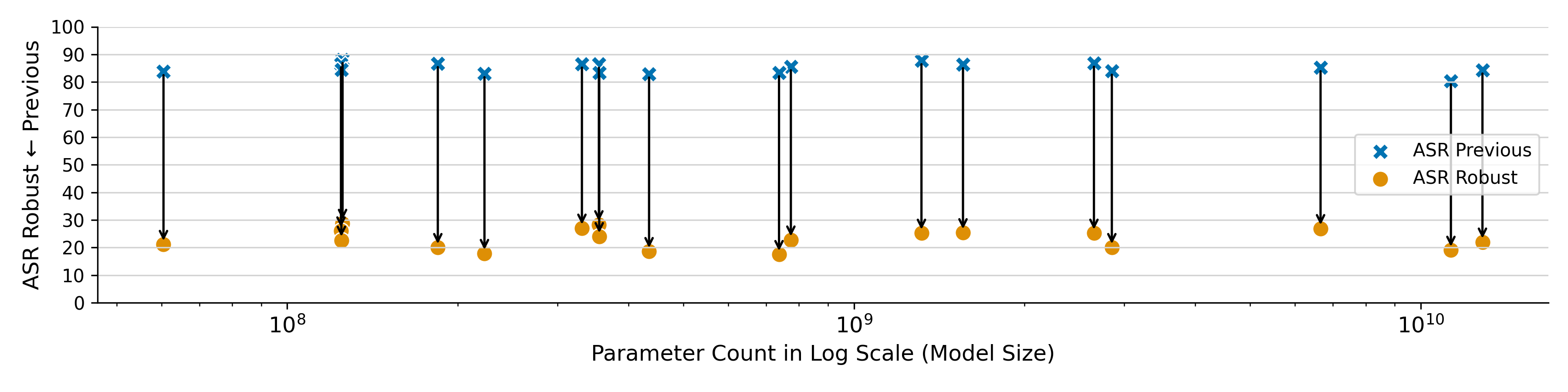}
    \caption{\texttt{BAE} used to fool the model.}
    \label{fig:asr_drop_bae_mnli}
  \end{subfigure}
  \begin{subfigure}{\textwidth}
    \centering
    \includegraphics[width=\textwidth]{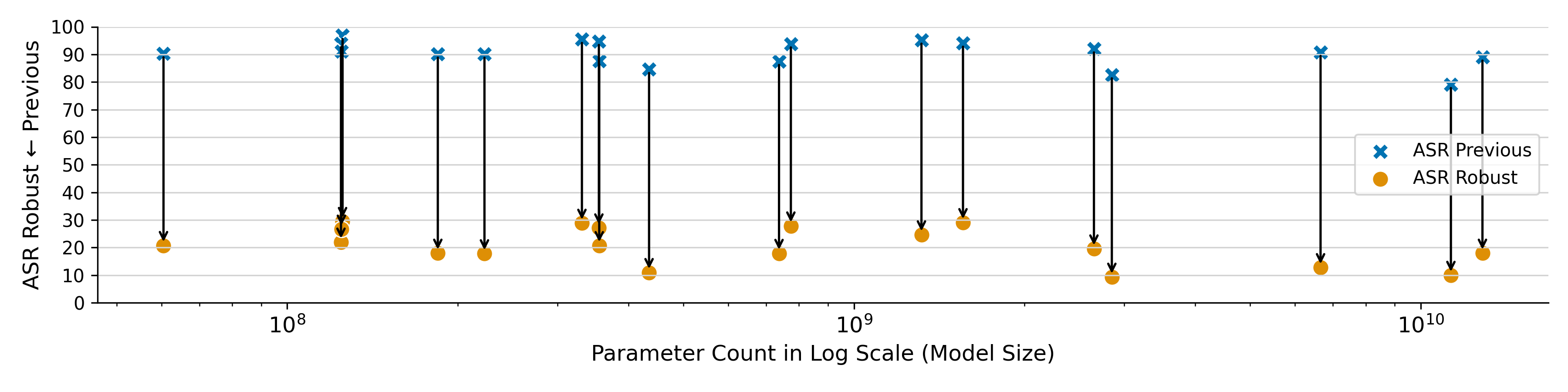}
    \caption{\texttt{TextBugger} used to fool the model.}
    \label{fig:asr_drop_textbugger_mnli}
  \end{subfigure}
  \begin{subfigure}{\textwidth}
    \centering
    \includegraphics[width=\textwidth]{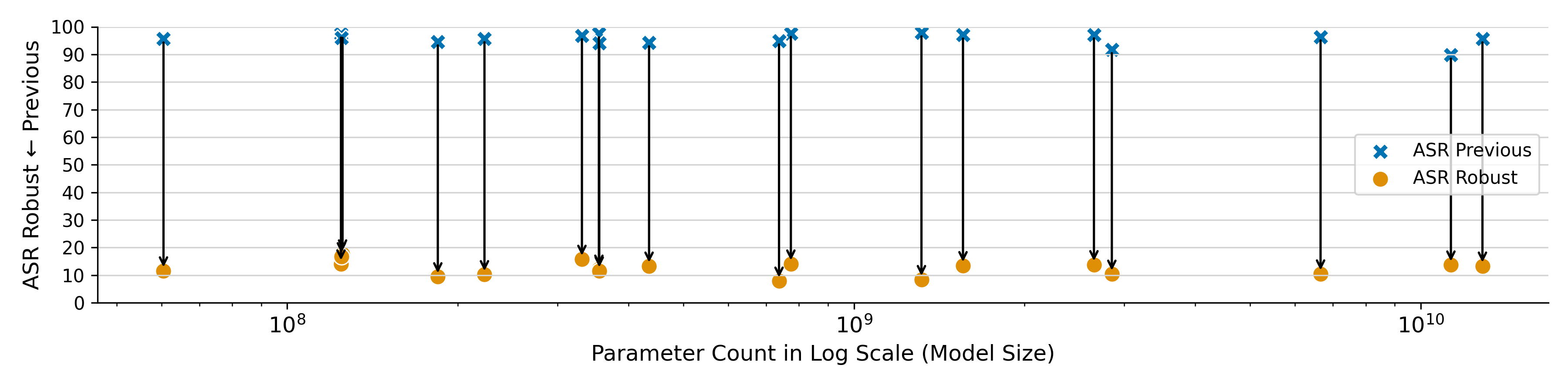}
    \caption{\texttt{PWWS} used to fool the model.}
    \label{fig:asr_drop_pwws_mnli}
  \end{subfigure}
  \caption{The change in the attack success rate (ASR) as measured in prior work (\ref{eq:asr_prev}) vs.\ our robust modification (\ref{eq:asr_robust}) in the \textbf{MNLI} experimental setup. The attack type is unknown because \texttt{TextFooler} is used to train the defense.}
  \label{fig:asr_drop_appendix_mnli}
\end{figure*}

\begin{figure*}[t]
  \centering
  \begin{subfigure}{\textwidth}
    \centering
    \includegraphics[width=\textwidth]{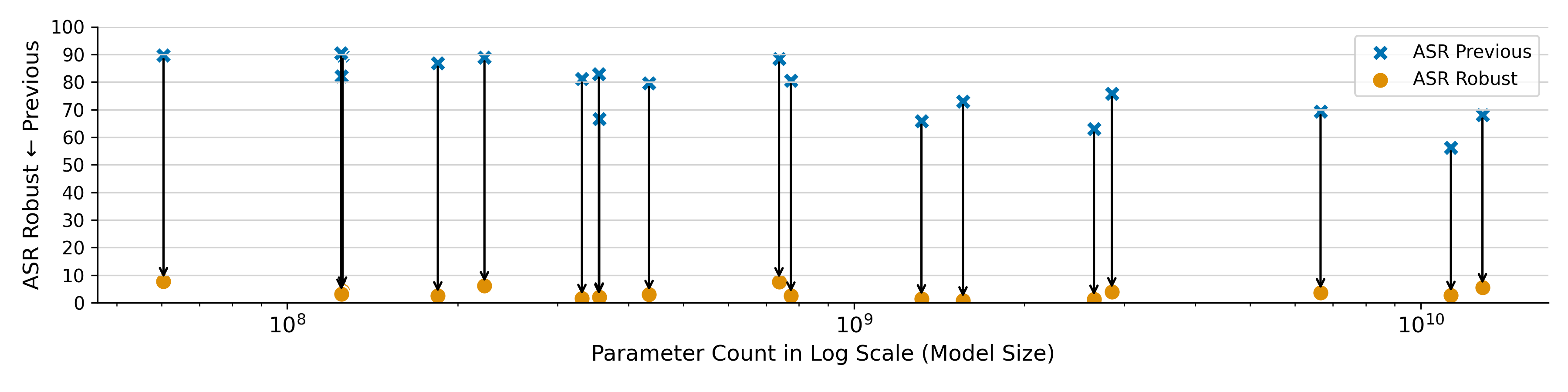}
    \caption{\texttt{TextFooler} used to fool the model.}
    \label{fig:asr_drop_textfooler_agnews}
  \end{subfigure}
  \begin{subfigure}{\textwidth}
    \centering
    \includegraphics[width=\textwidth]{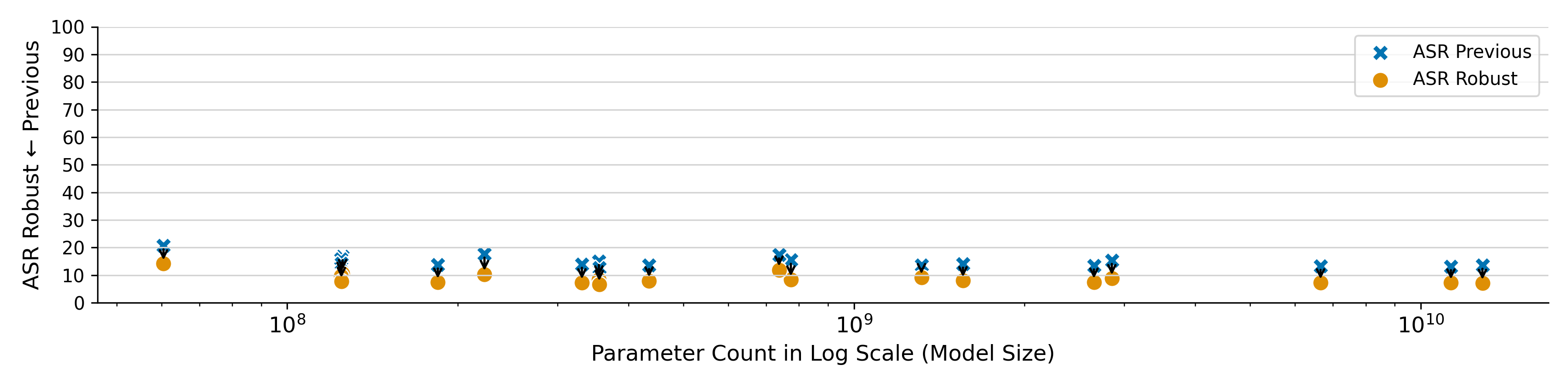}
    \caption{\texttt{BAE} used to fool the model.}
    \label{fig:asr_drop_bae_agnews}
  \end{subfigure}
  \begin{subfigure}{\textwidth}
    \centering
    \includegraphics[width=\textwidth]{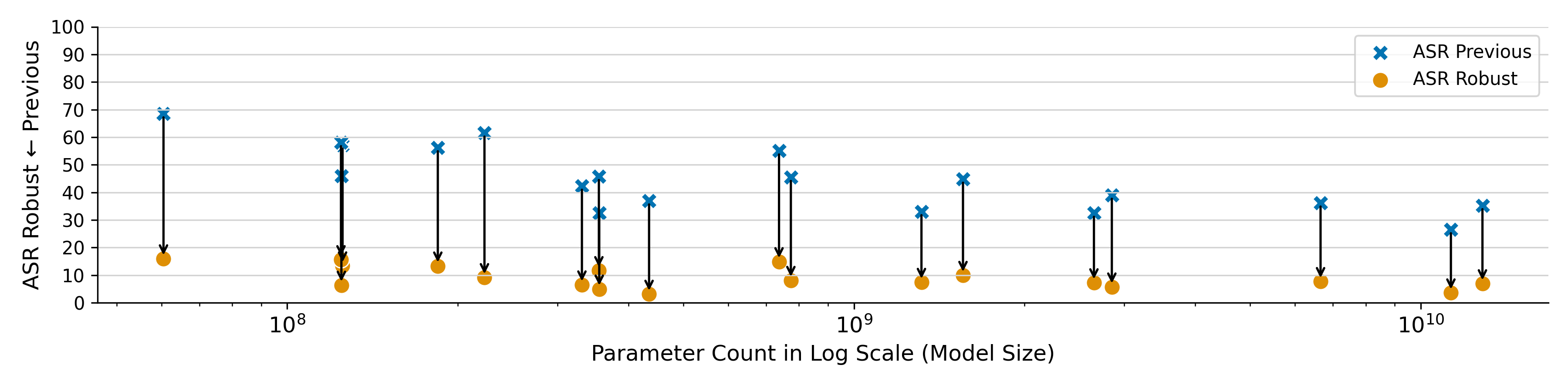}
    \caption{\texttt{TextBugger} used to fool the model.}
    \label{fig:asr_drop_textbugger_agnews}
  \end{subfigure}
  \begin{subfigure}{\textwidth}
    \centering
    \includegraphics[width=\textwidth]{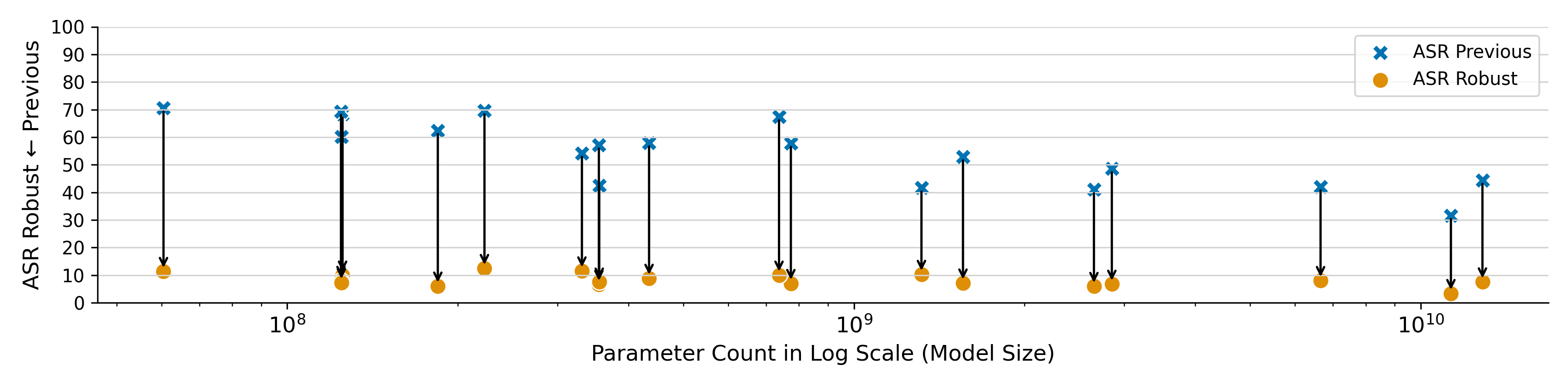}
    \caption{\texttt{PWWS} used to fool the model.}
    \label{fig:asr_drop_pwws_agnews}
  \end{subfigure}
  \caption{The change in the attack success rate (ASR) as measured in prior work (\ref{eq:asr_prev}) vs.\ our robust modification (\ref{eq:asr_robust}) in the \textbf{AGNews} experimental setup. The attack type is unknown because \texttt{TextFooler} is used to train the defense.}
  \label{fig:asr_drop_appendix_agnews}
\end{figure*}

We provide additional results that supplement the
main body of the paper. 
\begin{compactitem}
    \item Figure \ref{fig:asr_drop_appendix_mnli}--\ref{fig:asr_drop_appendix_agnews}  show the ASR drop for attacks not included in Figure \ref{fig:asr_drop}.
    \item Figures \ref{fig:dsr_lar_mnli} and \ref{fig:dsr_lar_agnews} show the $\lar$ and $\dsr$ values in the MNLI and AGNews experimental setup.
    \item Table \ref{tab:adversarial-gpt4-analysis} shows the label mismatch between a human (one of the authors in this case) and those assigned by GPT-4. Please refer to \ref{sec:gpt_lar_analysis} for discussion and anlaysis.
    \item Tables \ref{tab:agnews-bae}--\ref{tab:mnli-textfooler} provide examples produced by the five attack methods. 
\end{compactitem}

\subsection{Descriptions of Attacks}
\label{sec:appendix_attacks}
We analyze the following commonly used attack methods for NLP models: 
\begin{compactitem}
\item \texttt{TextFooler} \cite{DBLP:conf/aaai/JinJZS20}: Measures a token's importance with the change in the prediction score after removing it. Important tokens are replaced with possible synonyms found in an embedding space that have the same POS tag. Other attacks we study identify important tokens similar to \texttt{TextFooler}.
\item \texttt{BAE} \cite{garg-ramakrishnan-2020-bae}: Replaces or extends important tokens with MLM \cite{devlin-etal-2019-bert}.
\item \texttt{TextBugger} \cite{DBLP:conf/ndss/LiJDLW19}: Combines character edits with embedding-based and heuristic token edits. 
\item \texttt{PWWS} \cite{ren-etal-2019-generating}:  Replaces important tokens with WordNet synonyms with special care for named entities. Additionally, it constructs a priority order for candidate edits.
\item \texttt{DeepWordBug} \cite{DBLP:conf/sp/GaoLSQ18}: Edits important tokens with 4 character-level heuristics. 
\end{compactitem}

\subsection{Analysis of Using GPT-4 to Determine Label Mismatch}
\label{sec:gpt_lar_analysis}

As mentioned in the main body of the text, we use GPT-4~\citep{openai2023gpt4} to determine the label of a perturbed example (i.e. $y_g(\pert(x))$). 
To assess the accuracy of annotations from GPT-4, we manually evaluate 100 perturbed instances using \texttt{DeepWordBug} and \texttt{TextFooler} for both MNLI and AGNews datasets. 

While annotating, we find that many of the examples generated by the attack algorithms are difficult to assign the labels to. Specifically, these cases arise when substitutions introduced by the attack algorithm either render the example incomprehensible or create difficulty in distinguishing between two potential labels (see for instance the first example in~\ref{tab:mnli-textfooler}). Therefore, in addition to task labels, we also identify such bad/invalid examples and report them as the percentage of total annotated examples. For measuring the agreement between human assigned labels, and GPT-4 assigned labels, we report it as a percentage of examples that are assigned one of the task labels by the human annotator.

The results are reported in table~\ref{tab:adversarial-gpt4-analysis}. 
We observe that, across all cases, GPT-4 has a higher agreement with human label (> 80 \%). 
Additionally, \texttt{DeepWordBug} has a much lower rate of invalid examples than \texttt{TextFooler}. This makes sense because word-level substitutions made by \texttt{TextFooler} have a higher chance of destroying the meaning as compared to the character-level substitutions made by \texttt{DeepWordBug}.
\begin{table}[t]
\centering
\resizebox{\columnwidth}{!}{
\begin{tabular}{@{}llcc}
\toprule
Dataset                 & Attack      & Agreement (\%) &  Invalid (\%) \\ \midrule
\multirow{2}{*}{MNLI}   & \texttt{TextFooler}  & 80.0            & 35.0                  \\
                        & \texttt{DeepWordBug} & 84.3            & 17.0                  \\
\midrule
\multirow{2}{*}{AGNews} & \texttt{TextFooler}  & 98.9            & 12.0                  \\
                        & \texttt{DeepWordBug} & 95.8            & 4.0                   \\ \bottomrule 
\end{tabular}
}
\caption{Agreement between GPT-4 labels and a human labeler, and \% of examples classified by human as invalid. }
\label{tab:adversarial-gpt4-analysis}
\end{table}

\section{Contrast Set Evaluation}
\begin{figure*}[!ht]
    \centering
    \includegraphics[width=\textwidth]{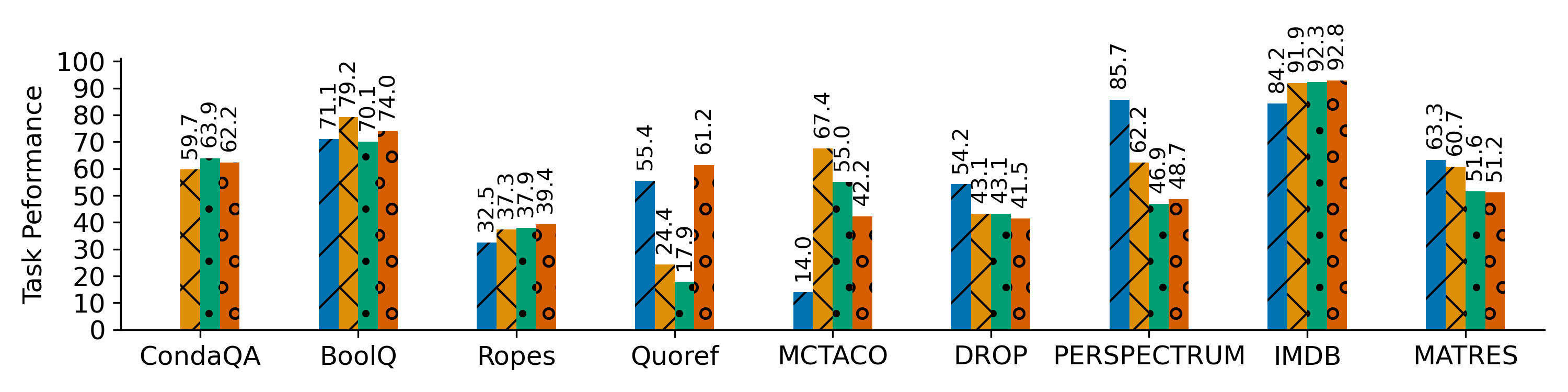}
    \includegraphics[width=\textwidth]{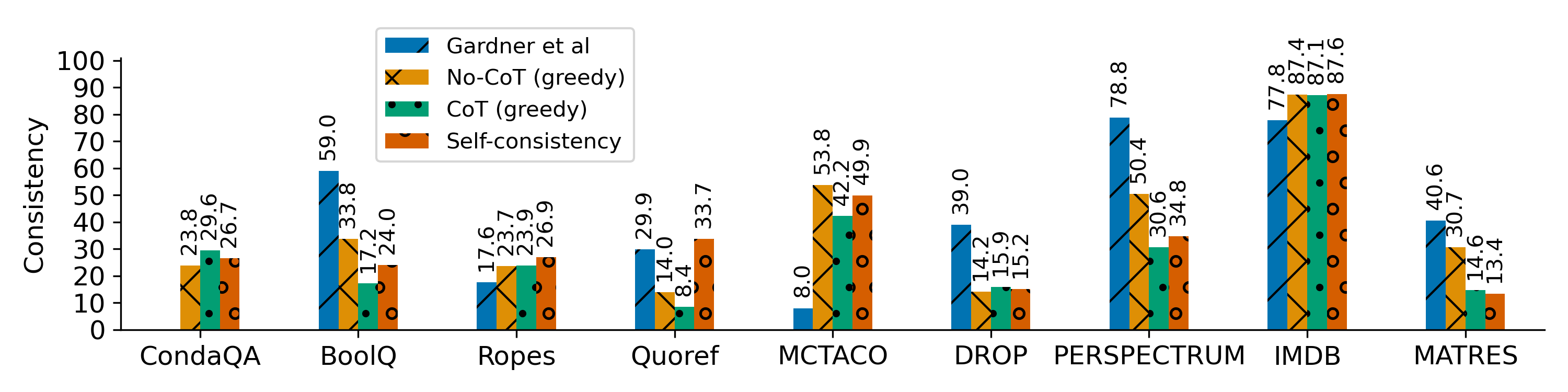}
    \caption{T\"{u}lu-2-13B performance with standard measures (accuracy, F1, token-F1) vs.\ contrast set consistency. Prompts include an instruction, 8 examples, and optionally explanations for chain-of-thought prompting and self-consistency decoding.} 
\label{fig:contrast_sets_results_tulu2}
\end{figure*}

\begin{figure*}[!ht]
    \centering
    \includegraphics[width=\textwidth]{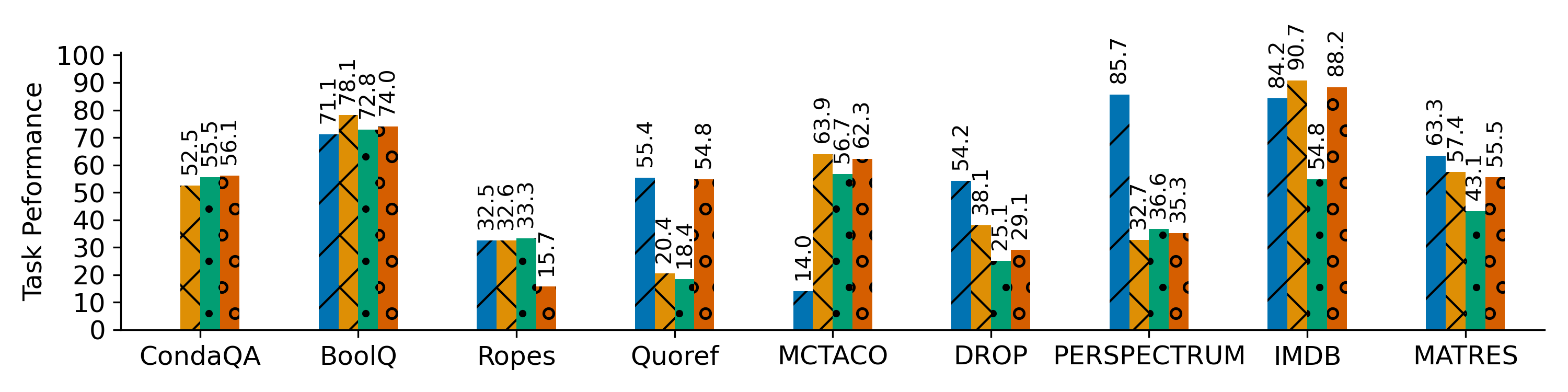}
    \includegraphics[width=\textwidth]{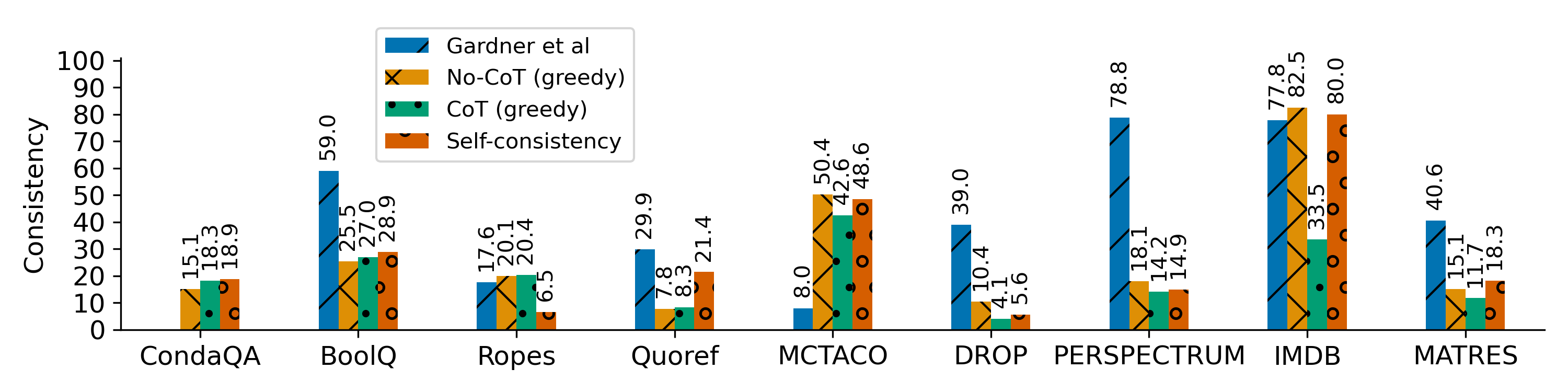}
    \caption{Llama-2-13B-chat performance with standard measures (accuracy, F1, token-F1) vs.\ contrast set consistency. Prompts include an instruction, 8 examples, and optionally explanations for chain-of-thought prompting and self-consistency decoding.} 
\label{fig:contrast_sets_results_llama2_chat}
\end{figure*}

We provide additional contrast set evaluation results:
\begin{compactitem}
\item Figure \ref{fig:contrast_sets_results_tulu2} shows the task performance and consistency of T{\"u}LU-2-13B model on the contrast sets.
\item Figure \ref{fig:contrast_sets_results_llama2_chat} shows the task performance and consistency of Llama2-13B-chat model on the contrast sets.
\end{compactitem}

\subsection{T\"{u}lu-2-13B vs Llama-2-13B}
\label{subsec:tulu_llama}
In most tasks, T\"{u}lu-2 outperforms Llama2 both in task performance and consistency. This is expected since T{\"u}lu-2 is a finetuned version of Llama2.
\subsection{T\"{u}lu-2-13B vs FLAN-T5-11B}
FLAN-T5 does better than T\"{u}lu-2 on all tasks except perspectrum on which T\"{u}lu-2 achieves both a higher task performance and consistency. T\"{u}lu-2 is less consistent than FLAN-T5 even on tasks such as Boolq and ropes where the gap between the task performance is small. Just as was the case with FLAN-T5, chain-of-thoughts and self-consistency decoding improve neither task performance nor consistency for T\"{u}lu-2. Despite better pretraining, there is still a huge gap between task performance and consistency across tasks. This suggests  that issues of robustness persist.
\clearpage 

\begin{table*}
\resizebox{\textwidth}{!}{%
\begin{tabular}{llllcp{8cm}}
\toprule
\textbf{Train} & \textbf{Eval} & \textbf{Metrics} & \textbf{Evaluation Types} & \textbf{Included?} & \textbf{Remarks} \\
\midrule
\multirow{10}{*}{\rotatebox[origin=c]{90}{MNLI}} & \href{https://github.com/tommccoy1/hans/}{HANS} \citep{mccoy-etal-2019-right} & Accuracy & challenge, domain &  \cmark  &  \\
\arrayrulecolor{black!20}\cmidrule{2-6}
 & \href{https://github.com/rabeehk/robust-nli/}{MNLI-hard} \citep{gururangan-etal-2018-annotation} & Accuracy & challenge, domain &  \cmark  &  \\
 \arrayrulecolor{black!20}\cmidrule{2-6}
 & \href{https://github.com/facebookresearch/anli}{ANLI} \citep{nie-etal-2020-adversarial} & Accuracy & challenge &  \cmark  &  \\
 \arrayrulecolor{black!20}\cmidrule{2-6}
 & \href{https://github.com/BIU-NLP/Breaking\_NLI}{Breaking NLI} \citep{glockner-etal-2018-breaking} & Accuracy & challenge &  \cmark  &  \\
 \arrayrulecolor{black!20}\cmidrule{2-6}
 & \href{https://gluebenchmark.com/diagnostics}{NLI Diagnostics} \citep{wang-etal-2018-glue} & Matthew's Corr, Accuracy & challenge &  \cmark  &  \\
 \arrayrulecolor{black!20}\cmidrule{2-6}
 & \href{https://github.com/AbhilashaRavichander/NLI\_StressTest}{Stress Test} \citep{naik-etal-2018-stress} & Accuracy & challenge &  \cmark  &  \\
 \arrayrulecolor{black!20}\cmidrule{2-6}
 & \href{https://huggingface.co/datasets/snli}{SNLI} \citep{bowman-etal-2015-large} & Accuracy & domain &  \cmark  &  \\
 \arrayrulecolor{black!20}\cmidrule{2-6}
 & \href{https://wellecks.com/dialogue\_nli/}{DNLI} \citep{welleck-etal-2019-dialogue} & Accuracy & domain &  \cmark  &  \\
 \arrayrulecolor{black!20}\cmidrule{2-6}
 & \href{https://huggingface.co/datasets/glue}{MNLI-Matched} \citep{williams-etal-2018-broad} & Accuracy & in-domain &  \cmark  &  \\
 \arrayrulecolor{black!20}\cmidrule{2-6}
 & \href{https://huggingface.co/datasets/glue}{MNLI-Mismatched} \citep{williams-etal-2018-broad} & Accuracy & in-domain &  \cmark  &  \\
 \midrule
\multirow{10}{*}{\rotatebox[origin=c]{90}{SNLI}} & \href{https://huggingface.co/datasets/glue}{MNLI} \citep{williams-etal-2018-broad} & Accuracy & domain &  \cmark  &  \\
\arrayrulecolor{black!20}\cmidrule{2-6}
 & \href{https://github.com/tommccoy1/hans/}{HANS} \citep{mccoy-etal-2019-right} & Accuracy & challenge, domain &  \cmark  &  \\
 \arrayrulecolor{black!20}\cmidrule{2-6}
 & \href{https://github.com/facebookresearch/anli}{ANLI} \citep{nie-etal-2020-adversarial}& Accuracy & challenge &  \cmark  &  \\
 \arrayrulecolor{black!20}\cmidrule{2-6}
 & \href{https://github.com/rabeehk/robust-nli/}{SNLI-hard} \citep{gururangan-etal-2018-annotation} & Accuracy & challenge &  \cmark  &  \\
 \arrayrulecolor{black!20}\cmidrule{2-6}
 & \href{https://gluebenchmark.com/diagnostics}{NLI Diagnostics} \citep{wang-etal-2018-glue} & Matthew's Corr, Accuracy & challenge, domain &  \cmark  &  \\
 \arrayrulecolor{black!20}\cmidrule{2-6}
 & \href{https://wellecks.com/dialogue\_nli/}{DNLI} \citep{welleck-etal-2019-dialogue}& Accuracy & domain &  \cmark  &  \\
 \arrayrulecolor{black!20}\cmidrule{2-6}
 & \href{https://github.com/AbhilashaRavichander/NLI\_StressTest}{Stress Test} \citep{naik-etal-2018-stress}& Accuracy & challenge, domain &  \cmark  &  \\
 \arrayrulecolor{black!20}\cmidrule{2-6}
 & \href{https://github.com/acmi-lab/counterfactually-augmented-data}{SNLI CAD} \citep{DBLP:conf/iclr/KaushikHL20}& Accuracy & challenge &  \cmark  &  \\
 \arrayrulecolor{black!20}\cmidrule{2-6}
 & \href{https://github.com/BIU-NLP/Breaking\_NLI}{Breaking NLI} \citep{glockner-etal-2018-breaking}& Accuracy & challenge &  \cmark  &  \\
 \arrayrulecolor{black!20}\cmidrule{2-6}
 & \href{https://huggingface.co/datasets/snli}{SNLI} \citep{bowman-etal-2015-large}& Accuracy & in-domain &  \cmark  &  \\
\midrule
\multirow{3}{*}{\rotatebox[origin=c]{90}{QQP}}& \href{https://github.com/lanwuwei/Twitter-URL-Corpus}{Twitter PPDB} \citep{lan-etal-2017-continuously} & Accuracy & domain &  \cmark  & Dataset not available at the link. Acquired via email. \\
\arrayrulecolor{black!20}\cmidrule{2-6}
 & \href{https://github.com/google-research-datasets/paws}{PAWS-QQP} \citep{zhang-etal-2019-paws} & Accuracy, AUC & challenge, domain &  \cmark  &  \\
 \arrayrulecolor{black!20}\cmidrule{2-6}
 & \href{https://huggingface.co/datasets/glue}{QQP}\tablefootnote{\url{https://quoradata.quora.com/First-Quora-Dataset-Release-Question-Pairs}} & Accuracy, F1 & in-domain &  \cmark  &  \\
\midrule
\multirow{3}{*}{\rotatebox[origin=c]{90}{YELP}} & \href{https://huggingface.co/datasets/imdb}{IMDB} \citep{maas-etal-2011-learning} & Accuracy & domain &  \cmark  &  \\
\arrayrulecolor{black!20}\cmidrule{2-6}
 & \href{https://huggingface.co/datasets/glue}{SST2} \citep{socher-etal-2013-recursive} & Accuracy & domain &  \cmark  &  \\
 & \href{https://huggingface.co/datasets/yelp\_polarity}{YELP} \citep{DBLP:conf/nips/ZhangZL15}& Accuracy & in-domain &  \cmark  &  \\
 \midrule
\multirow{7}{*}{\rotatebox[origin=c]{90}{SST2}} & \href{https://huggingface.co/datasets/imdb}{IMDB} \citep{maas-etal-2011-learning}& Accuracy & domain &  \cmark  &  \\
\arrayrulecolor{black!20}\cmidrule{2-6}
 & \href{https://github.com/allenai/contrast-sets/tree/main/IMDb}{IMDB contrast} \citep{gardner-etal-2020-evaluating} & Accuracy & challenge &  \cmark  &  \\
 \arrayrulecolor{black!20}\cmidrule{2-6}
 & \href{https://github.com/acmi-lab/counterfactually-augmented-data/tree/master/sentiment/combined/paired}{C-IMDB} \citep{DBLP:conf/iclr/KaushikHL20} & Accuracy & challenge &  \cmark  &  \\
 \arrayrulecolor{black!20}\cmidrule{2-6}
 & \href{https://huggingface.co/datasets/yelp\_polarity}{YELP} \citep{DBLP:conf/nips/ZhangZL15}& Accuracy & domain &  \cmark  &  \\
 \arrayrulecolor{black!20}\cmidrule{2-6}
 & \href{https://cseweb.ucsd.edu/\textasciitilde jmcauley/datasets.html\#amazon\_reviews}{AmazonReviews} \citep{DBLP:conf/www/HeM16} & Accuracy & domain &  \cmark  & Six Genres: appliances, beauty, fashion, gift\_cards, magazines, software (10k examples each) \\
 \arrayrulecolor{black!20}\cmidrule{2-6}
 & \href{https://huggingface.co/datasets/rotten\_tomatoes}{RottenTomatoes} \citep{pang-lee-2005-seeing} & Accuracy & domain &  \cmark  &  \\
 \arrayrulecolor{black!20}\cmidrule{2-6}
 & \href{https://huggingface.co/datasets/glue}{SST2} \citep{socher-etal-2013-recursive}& Accuracy & in-domain &  \cmark  &  \\
 \midrule
\multirow{8}{*}{\rotatebox[origin=c]{90}{IMDB}} & \href{https://huggingface.co/datasets/glue}{SST2} \citep{socher-etal-2013-recursive}& Accuracy & domain &  \cmark  &  \\
\arrayrulecolor{black!20}\cmidrule{2-6}
 & \href{https://huggingface.co/datasets/yelp\_polarity}{YELP} \citep{DBLP:conf/nips/ZhangZL15}& Accuracy & domain &  \cmark  &  \\
 \arrayrulecolor{black!20}\cmidrule{2-6}
 & \href{https://github.com/allenai/contrast-sets/tree/main/IMDb}{IMDB Contrast} \citep{gardner-etal-2020-evaluating}& Accuracy & challenge &  \cmark  &  \\
 \arrayrulecolor{black!20}\cmidrule{2-6}
 & \href{https://github.com/acmi-lab/counterfactually-augmented-data/tree/master/sentiment/combined/paired}{C-IMDB} \citep{DBLP:conf/iclr/KaushikHL20}& Accuracy & challenge &  \cmark  &  \\
 \arrayrulecolor{black!20}\cmidrule{2-6}
 & \href{https://alt.qcri.org/semeval2017/task4/index.php?id=data-and-tools}{Twitter Emotion} \citep{rosenthal-etal-2017-semeval} & Accuracy & domain &  \cmark  & SemEval-2017 Task - 4 - Twitter Sentiment Analysis \\
 \arrayrulecolor{black!20}\cmidrule{2-6}
 & \href{https://cseweb.ucsd.edu/\textasciitilde jmcauley/datasets.html\#amazon\_reviews}{AmazonReviews} \citep{DBLP:conf/www/HeM16}& Accuracy & domain &  \cmark  & Six Genres: appliances, beauty, fashion, gift\_cards, magazines, software (10k examples each) \\
 \arrayrulecolor{black!20}\cmidrule{2-6}
 & \href{https://huggingface.co/datasets/rotten\_tomatoes}{RottenTomatoes} \citep{pang-lee-2005-seeing}& Accuracy & domain &  \cmark  &  \\
 \arrayrulecolor{black!20}\cmidrule{2-6}
 & \href{https://huggingface.co/datasets/imdb}{IMDB} \citep{maas-etal-2011-learning}& Accuracy & in-domain &  \cmark  &  \\
 \midrule
\multirow{2}{*}{\rotatebox[origin=c]{90}{MRPC}} & \href{https://github.com/google-research-datasets/paws}{PAWS-Wiki} \citep{zhang-etal-2019-paws}& Accuracy, AUC & domain &  \cmark  &  \\
 & \href{https://huggingface.co/datasets/glue}{MRPC} \citep{dolan-brockett-2005-automatically} & Accuracy, F1 & in-domain &  \cmark  &  \\
\midrule
\multirow{3}{*}{\rotatebox[origin=c]{90}{FEVER}} & \href{https://github.com/TalSchuster/FeverSymmetric}{FEVER-Symmetric v1} \citep{schuster-etal-2019-towards} & Accuracy & challenge, domain &  \cmark  &  \\
\arrayrulecolor{black!20}\cmidrule{2-6}
 & \href{https://github.com/TalSchuster/FeverSymmetric}{FEVER-Symmetric v2} \citep{schuster-etal-2019-towards} & Accuracy & challenge, domain &  \cmark  &  \\
 \arrayrulecolor{black!20}\cmidrule{2-6}
 & \href{https://huggingface.co/datasets/pietrolesci/nli\_fever}{FEVER} \citep{DBLP:conf/aaai/NieCB19} & Accuracy & in-domain &  \cmark  & Used as NLI \\
 \midrule
\multirow{2}{*}{\rotatebox[origin=c]{90}{QNLI}} & \href{https://worksheets.codalab.org/worksheets/0xc86d3ebe69a3427d91f9aaa63f7d1e7d/}{Adversarial Squad} \citep{jia-liang-2017-adversarial} & Accuracy & domain &  \cmark  & Used in Dataset Cartography for the first time. Not Available. 
Created from the source \\
\arrayrulecolor{black!20}\cmidrule{2-6}
 & \href{https://huggingface.co/datasets/glue}{QNLI} \citep{wang-etal-2018-glue} & Accuracy & in-domain &  \cmark  &  \\
\bottomrule
\end{tabular}
}
\caption{Evaluation setups for NLI, Sentiment Classification, Paraphrase Identification, and Claim Verification tasks. The name of each dataset contains a link to its source.
}
\label{tab:tab:train_eval1}
\end{table*}

\begin{table*}
\resizebox{\textwidth}{!}{%
\begin{tabular}{llllcp{8cm}}
\toprule
\textbf{Train} & \textbf{Eval} & \textbf{Metrics} & \textbf{Evaluation Types} & \textbf{Included?} & \textbf{Remarks} \\
\midrule
\multirow{22}{*}{\large \rotatebox[origin=c]{90}{SQuAD}} & \href{https://huggingface.co/datasets/mrqa}{TriviaQA} \citep{joshi-etal-2017-triviaqa} & Exact Match, F1 & domain &  \cmark  &  \\
\arrayrulecolor{black!20}\cmidrule{2-6}
& \href{https://huggingface.co/datasets/mrqa}{NaturalQuestions} \citep{kwiatkowski-etal-2019-natural} & Exact Match, F1 & domain, challenge &  \cmark  &  \\
\arrayrulecolor{black!20}\cmidrule{2-6}
& \href{https://huggingface.co/datasets/mrqa}{NewsQA} \citep{trischler-etal-2017-newsqa} & Exact Match, F1 & domain, challenge &  \cmark  &  \\
\arrayrulecolor{black!20}\cmidrule{2-6}
& \href{https://huggingface.co/datasets/mrqa}{HotPotQA} \citep{yang-etal-2018-hotpotqa} & Exact Match, F1 & domain &  \cmark  &  \\
\arrayrulecolor{black!20}\cmidrule{2-6}
& \href{https://huggingface.co/datasets/mrqa}{SearchQA} \citep{DBLP:journals/corr/DunnSHGCC17} & Exact Match, F1 & domain &  \cmark  &  \\
\arrayrulecolor{black!20}\cmidrule{2-6}
& \href{https://huggingface.co/datasets/mrqa}{BioASQ} \citep{DBLP:journals/bmcbi/TsatsaronisBMPZ15} & Exact Match, F1 & domain &  \cmark  &  \\
\arrayrulecolor{black!20}\cmidrule{2-6}
& \href{https://huggingface.co/datasets/mrqa}{TextbookQA} \citep{DBLP:conf/cvpr/KembhaviSSCFH17} & Exact Match, F1 & domain &  \cmark  &  \\
\arrayrulecolor{black!20}\cmidrule{2-6}
& \href{https://huggingface.co/datasets/xquad}{XQuAD} \citep{DBLP:conf/acl/ArtetxeRY20} & Exact Match, F1 & domain &  \xmark  & English subset already included in SQuAD validation \\
\arrayrulecolor{black!20}\cmidrule{2-6}
& \href{https://github.com/nusnlp/paraphrasing-squad}{Non-Adversarial Paraphrased} \citep{gan-ng-2019-improving} & Exact Match, F1 & challenge &  \cmark  &  \\
\arrayrulecolor{black!20}\cmidrule{2-6}
& \href{https://github.com/nusnlp/paraphrasing-squad}{Adversarial Paraphrased} \citep{gan-ng-2019-improving} & Exact Match, F1 & challenge &  \cmark  &  \\
\arrayrulecolor{black!20}\cmidrule{2-6}
& \href{https://github.com/Alab-NII/mrc-heuristics/blob/master/subsets/squad-hard-subset.json}{SQuAD-hard} \citep{sugawara-etal-2018-makes} & Exact Match, F1 & challenge &  \cmark  &  \\
\arrayrulecolor{black!20}\cmidrule{2-6}
& \href{https://github.com/marcotcr/qa\_consistency/}{SQuAD-implications} \citep{ribeiro-etal-2019-red} & Exact Match, F1 & challenge &  \xmark  & Generating set of implications requires each model's predictions. \\
\arrayrulecolor{black!20}\cmidrule{2-6}
& \href{https://worksheets.codalab.org/worksheets/0xc86d3ebe69a3427d91f9aaa63f7d1e7d/}{AddOneSent} \citep{jia-liang-2017-adversarial} & Exact Match, F1 & challenge &  \cmark  & HF \href{https://huggingface.co/datasets/squad \_adversarial}{link} gives loading error.
This \href{https://aclanthology.org/2021.acl-srw.21.pdf}{paper} uses it wrongly. The examples come from 1000 dev examples. The original file contains these 1000 extra clean dev examples that need to be removed. 
So our count of these are 2560 and 787 instead of 3560 and 1787 as reported in the paper. \\
\arrayrulecolor{black!20}\cmidrule{2-6}
& \href{https://worksheets.codalab.org/worksheets/0xc86d3ebe69a3427d91f9aaa63f7d1e7d/}{AddSent} \citep{jia-liang-2017-adversarial} & Exact Match, F1 & challenge &  \cmark  & Same as AddOneSent %
\\
\arrayrulecolor{black!20}\cmidrule{2-6}
& \href{https://huggingface.co/datasets/quoref}{Quoref} \citep{DBLP:conf/emnlp/DasigiLMSG19} & Exact Match, F1 & challenge &  \cmark  &  \\
\arrayrulecolor{black!20}\cmidrule{2-6}
& \href{https://github.com/allenai/tailor}{QA Contrast} \citep{ross-etal-2022-tailor} & Exact Match, F1 & challenge &  \xmark  & Authors \href{https://arxiv.org/pdf/2107.07150.pdf}{here} say they "use the QA implication challenge set (Rajpurkar et al., 2016) as the
human contrast set",
but it's unclear what they are referring to. \\
\arrayrulecolor{black!20}\cmidrule{2-6}
& \href{https://quac.ai/}{QuAC} \citep{choi-etal-2018-quac} & Exact Match, F1 & domain &  \cmark  & Original dataset has multi-turn dialogues which we convert to SQuAD format using this \href{https://github.com/amazon-science/qa-dataset-converter}{tool} \\
\arrayrulecolor{black!20}\cmidrule{2-6}
& \href{https://huggingface.co/datasets/mrqa}{DROP} \citep{dua-etal-2019-drop} & Exact Match, F1 & domain &  \cmark  &  \\
\arrayrulecolor{black!20}\cmidrule{2-6}
& \href{https://huggingface.co/datasets/mrqa}{DuoRC} \citep{saha-etal-2018-duorc} & Exact Match, F1 & domain &  \cmark  &  \\
\arrayrulecolor{black!20}\cmidrule{2-6}
& \href{https://huggingface.co/datasets/mrqa}{RACE} \citep{lai-etal-2017-race} & Exact Match, F1 & domain &  \cmark  &  \\
\arrayrulecolor{black!20}\cmidrule{2-6}
& \href{https://huggingface.co/datasets/mrqa}{RelationExtraction} \citep{levy-etal-2017-zero} & Exact Match, F1 & domain &  \cmark  &  \\
\arrayrulecolor{black!20}\cmidrule{2-6}
& \href{https://github.com/facebookresearch/mlqa}{MLQA} \citep{lewis-etal-2020-mlqa} & Exact Match, F1 & domain &  \cmark  & Using only en subset \\
\arrayrulecolor{black!20}\cmidrule{2-6}
& \href{https://huggingface.co/datasets/mrqa}{SQuAD} \citep{rajpurkar-etal-2016-squad} & Exact Match, F1 & in-domain &  \cmark  &  \\
\midrule
\multirow{11}{*}{\large \rotatebox[origin=c]{90}{NewsQA}} & \href{https://huggingface.co/datasets/mrqa}{SQuAD} \citep{rajpurkar-etal-2016-squad}& Exact Match, F1 & domain &  \cmark  &  \\
\arrayrulecolor{black!20}\cmidrule{2-6}
& \href{https://huggingface.co/datasets/mrqa}{NaturalQuestions} \citep{kwiatkowski-etal-2019-natural} & Exact Match, F1 & domain &  \cmark  &  \\
\arrayrulecolor{black!20}\cmidrule{2-6}
& \href{https://huggingface.co/datasets/mrqa}{TriviaQA} \citep{joshi-etal-2017-triviaqa}& Exact Match, F1 & domain &  \cmark  &  \\
\arrayrulecolor{black!20}\cmidrule{2-6}
& \href{https://quac.ai/}{QuAC} \citep{choi-etal-2018-quac}& Exact Match, F1 & domain &  \cmark  & Original dataset has multi-turn dialogues
which we convert to SQuAD format using this \href{https://github.com/amazon-science/qa-dataset-converter}{tool} \\
\arrayrulecolor{black!20}\cmidrule{2-6}
& \href{https://huggingface.co/datasets/mrqa}{BioASQ} \citep{DBLP:journals/bmcbi/TsatsaronisBMPZ15}& Exact Match, F1 & domain &  \cmark  &  \\
\arrayrulecolor{black!20}\cmidrule{2-6}
& \href{https://huggingface.co/datasets/mrqa}{DROP} \citep{dua-etal-2019-drop}& Exact Match, F1 & domain &  \cmark  &  \\
\arrayrulecolor{black!20}\cmidrule{2-6}
& \href{https://huggingface.co/datasets/mrqa}{DuoRC} \citep{saha-etal-2018-duorc}& Exact Match, F1 & domain &  \cmark  &  \\
\arrayrulecolor{black!20}\cmidrule{2-6}
& \href{https://huggingface.co/datasets/mrqa}{RACE} \citep{lai-etal-2017-race}& Exact Match, F1 & domain &  \cmark  &  \\
\arrayrulecolor{black!20}\cmidrule{2-6}
& \href{https://huggingface.co/datasets/mrqa}{RelationExtraction} \citep{levy-etal-2017-zero}& Exact Match, F1 & domain &  \cmark  &  \\
\arrayrulecolor{black!20}\cmidrule{2-6}
& \href{https://huggingface.co/datasets/mrqa}{TextbookQA} \citep{DBLP:conf/cvpr/KembhaviSSCFH17}& Exact Match, F1 & domain &  \cmark  &  \\
\arrayrulecolor{black!20}\cmidrule{2-6}
& \href{https://huggingface.co/datasets/mrqa}{NewsQA} \citep{trischler-etal-2017-newsqa}& Exact Match, F1 & in-domain &  \cmark  &  \\
\midrule
\multirow{13}{*}{\large \rotatebox[origin=c]{90}{NaturalQuestions}} & \href{https://huggingface.co/datasets/mrqa}{TriviaQA} \citep{joshi-etal-2017-triviaqa}& Exact Match, F1 & domain &  \cmark  &  \\
& \href{https://huggingface.co/datasets/mrqa}{SQuAD} \citep{rajpurkar-etal-2016-squad}& Exact Match, F1 & domain &  \cmark  &  \\
\arrayrulecolor{black!20}\cmidrule{2-6}
& \href{https://huggingface.co/datasets/mrqa}{NewsQA} \citep{trischler-etal-2017-newsqa}& Exact Match, F1 & domain &  \cmark  &  \\
\arrayrulecolor{black!20}\cmidrule{2-6}
& \href{https://huggingface.co/datasets/mrqa}{BioASQ} \citep{DBLP:journals/bmcbi/TsatsaronisBMPZ15}& Exact Match, F1 & domain &  \cmark  &  \\
\arrayrulecolor{black!20}\cmidrule{2-6}
& \href{https://huggingface.co/datasets/mrqa}{DROP} \citep{dua-etal-2019-drop}& Exact Match, F1 & domain &  \cmark  &  \\
\arrayrulecolor{black!20}\cmidrule{2-6}
& \href{https://huggingface.co/datasets/mrqa}{DuoRC} \citep{saha-etal-2018-duorc}& Exact Match, F1 & domain &  \cmark  &  \\
\arrayrulecolor{black!20}\cmidrule{2-6}
& \href{https://huggingface.co/datasets/mrqa}{RACE} \citep{lai-etal-2017-race}& Exact Match, F1 & domain &  \cmark  &  \\
\arrayrulecolor{black!20}\cmidrule{2-6}
& \href{https://huggingface.co/datasets/mrqa}{RelationExtraction} \citep{levy-etal-2017-zero}& Exact Match, F1 & domain &  \cmark  &  \\
\arrayrulecolor{black!20}\cmidrule{2-6}
& \href{https://huggingface.co/datasets/mrqa}{TextbookQA} \citep{DBLP:conf/cvpr/KembhaviSSCFH17}& Exact Match, F1 & domain &  \cmark  &  \\
\arrayrulecolor{black!20}\cmidrule{2-6}
& \href{https://quac.ai/}{QuAC} \citep{choi-etal-2018-quac}& Exact Match, F1 & domain &  \cmark  &  \\
\arrayrulecolor{black!20}\cmidrule{2-6}
& \href{https://github.com/brmson/yodaqa}{TREC}\tablefootnote{\url{https://trec.nist.gov/data/qa/2001_qadata/main_task.html}} & Exact Match, F1 & domain &  \xmark  & Only applicable for open-domain QA
\\
\arrayrulecolor{black!20}\cmidrule{2-6}
& \href{https://nlp.cs.washington.edu/ambigqa/}{AmbigQA} \citep{min-etal-2020-ambigqa}& Exact Match, F1 & domain &  \xmark  & Only applicable for open-domain QA
\\
\arrayrulecolor{black!20}\cmidrule{2-6}
& \href{https://huggingface.co/datasets/mrqa}{NaturalQuestions} \citep{kwiatkowski-etal-2019-natural}& Exact Match, F1 & in-domain &  \cmark  &  \\
\midrule
\multirow{4}{*}{\large \rotatebox[origin=c]{90}{BoolQ}} & \href{https://github.com/allenai/contrast-sets}{BoolQ Contrast Set} \citep{gardner-etal-2020-evaluating} & Exact Match or Accuracy & challenge &  \cmark  & other name boolq\_contrast\_gardner \\
\arrayrulecolor{black!20}\cmidrule{2-6}
& \href{https://github.com/allenai/natural-perturbations}{BoolQ CAD} \citep{khashabi-etal-2020-bang} & Exact Match or Accuracy & challenge &  \cmark  &  \\
\arrayrulecolor{black!20}\cmidrule{2-6}
& \href{https://cogcomp.seas.upenn.edu/multirc/}{MultiRC} \citep{khashabi-etal-2018-looking} & Exact Match or Accuracy & domain &  \cmark  & Used only train set questions with yes/no answers \\
\arrayrulecolor{black!20}\cmidrule{2-6}
& \href{https://huggingface.co/datasets/boolq}{BoolQ} \citep{clark-etal-2019-boolq} & Exact Match or Accuracy & in-domain &  \cmark  &  \\
\bottomrule
\end{tabular}
}
\caption{Evaluation setups for Reading Comprehension and QA tasks. The name of each dataset contains a link to its source.}
\label{tab:train_eval2}
\end{table*}

\clearpage

\begin{table*}[t]
\centering
\resizebox{0.85\textwidth}{!}{
\begin{tabular}{lllccr}
\toprule
&  & \multicolumn{4}{c}{Results of a model with the max OOD-ID} \\
\cmidrule{3-6}
\textbf{Train} & \textbf{Test} & \textbf{Model} & \textbf{ID} & \textbf{OOD} & \textbf{OOD-ID} \\
\midrule
MNLI & HANS Lexical Overlap & DeBERTa-v3-Base & 89.66 & 98.48 & 8.82 \\
SNLI & HANS Lexical Overlap & T5-11B & 92.13 & 98.95 & 6.82 \\
IMDB & AmazonReviews & RoBERTa-Large & 92.43 & 94.89 & 2.46 \\
IMDB & YELP & OPT-125M & 89.14 & 91.04 & 1.90 \\
SNLI & NLI Diagnostics genitives/partitives & T5-Base & 90.11 & 91.67 & 1.56 \\
SST2 & YELP & T5-11B & 95.99 & 97.45 & 1.46 \\
FEVER & FEVER-symmetric v2 & OPT-13B & 78.75 & 79.63 & 0.88 \\
SNLI & NLI Diagnostics double negation & T5-3B & 92.16 & 92.86 & 0.70 \\
SST2 & AmazonReviews & T5-11B & 95.99 & 95.73 & -0.26 \\
IMDB & SST2 & DeBERTa-v3-Base & 91.41 & 91.09 & -0.32 \\
SST2 & RottenTomatoes & GPT-2 & 90.79 & 90.15 & -0.64 \\
SST2 & IMDB & T5-11B & 95.99 & 95.30 & -0.69 \\
MNLI & SNLI & T5-11B & 91.46 & 90.16 & -1.30 \\
IMDB & RottenTomatoes & DeBERTa-v3-Base & 91.41 & 89.56 & -1.85 \\
SNLI & NLI Diagnostics nominalization & T5-11B & 92.13 & 89.29 & -2.84 \\
\midrule
MNLI & MNLI-Hard & T5-11B & 91.46 & 88.43 & -3.03 \\
SNLI & Stress Test & T5-11B & 92.13 & 89.05 & -3.08 \\
QQP & Twitter PPDB & OPT-125M & 88.80 & 85.56 & -3.24 \\
SNLI & MNLI-Mismatched & T5-3B & 92.16 & 88.87 & -3.29 \\
SNLI & MNLI-Matched & T5-3B & 92.16 & 88.86 & -3.30 \\
SNLI & NLI Diagnostics universal & GPT-2-Medium & 88.69 & 85.19 & -3.50 \\
YELP & IMDB & T5-3B & 98.62 & 94.80 & -3.82 \\
YELP & SST2 & OPT-13B & 97.01 & 93.00 & -4.01 \\
FEVER & FEVER-symmetric v1 & OPT-13B & 78.75 & 74.48 & -4.27 \\
SNLI & NLI Diagnostics morphological negation & RoBERTa-Large & 91.57 & 87.18 & -4.39 \\
SNLI & NLI Diagnostics redundancy & DeBERTa-v3-Large & 92.24 & 87.18 & -5.06 \\
SNLI & NLI Diagnostics prepositional phrases & T5-Small & 88.40 & 83.33 & -5.07 \\
SNLI & NLI Diagnostics datives & RoBERTa-Base & 90.37 & 85.00 & -5.37 \\
IMDB & Twitter Emotion & OPT-125M & 89.14 & 81.33 & -7.81 \\
SNLI & NLI Diagnostics conjunction & T5-11B & 92.13 & 82.50 & -9.63 \\
SNLI & NLI Diagnostics upward monotone & GPT-2-large & 89.21 & 78.43 & -10.78 \\
SNLI & NLI Diagnostics ellipsis/implicits & T5-3B & 92.16 & 81.37 & -10.79 \\
QNLI & Adversarial Squad & T5-Large & 94.41 & 79.73 & -14.68 \\
MNLI & DNLI & GPT-2 & 81.66 & 66.89 & -14.77 \\
SNLI & NLI Diagnostics anaphora/coreference & T5-3B & 92.16 & 74.71 & -17.45 \\
SNLI & NLI Diagnostics lexical entailment & DeBERTa-v3-Large & 92.24 & 74.52 & -17.72 \\
SNLI & NLI Diagnostics news & DeBERTa-v3-Large & 92.24 & 74.51 & -17.73 \\
SNLI & NLI Diagnostics negation & T5-11B & 92.13 & 74.39 & -17.74 \\
MNLI & HANS Subsequence & OPT-125M & 74.39 & 56.09 & -18.30 \\
MNLI & HANS Constituent & OPT-350M & 81.76 & 62.87 & -18.89 \\
MRPC & PAWS Wiki & OPT-350M & 68.22 & 48.78 & -19.44 \\
SNLI & NLI Diagnostics conditionals & T5-3B & 92.16 & 71.88 & -20.28 \\
SNLI & NLI Diagnostics core args & DeBERTa-v3-Large & 92.24 & 71.79 & -20.45 \\
SNLI & NLI Diagnostics coordination scope & T5-3B & 92.16 & 71.67 & -20.49 \\
SNLI & NLI Diagnostics artificial & T5-11B & 92.13 & 71.33 & -20.80 \\
SNLI & HANS Constituent & DeBERTa-v3-Large & 92.24 & 71.17 & -21.07 \\
SNLI & NLI Diagnostics restrictivity & OPT-6.7B & 90.41 & 69.23 & -21.18 \\
SNLI & HANS Subsequence & T5-3B & 92.16 & 70.26 & -21.90 \\
SNLI & NLI Diagnostics quantifiers & OPT-13B & 91.38 & 69.23 & -22.15 \\
SNLI & NLI Diagnostics symmetry/collectivity & GPT-2-large & 89.21 & 66.67 & -22.54 \\
SNLI & NLI Diagnostics relative clauses & T5-3B & 92.16 & 68.75 & -23.41 \\
\bottomrule
\end{tabular}%
}
\caption{The OOD evaluation of \underline{classification} \textbf{finetuned} models (continued in Table \ref{tab:appendix_class_splits_part_2}). All results are accuracy scores. For each train-test split, we report the model with the best OOD generalization (among 19 models), defined as the highest OOD accuracy minus ID accuracy. All except a few models in the upper part of the table achieve an in-domain accuracy over 90\% while also maintaining similar OOD and ID accuracies. These splits might no longer be fitting for OOD robustness research.}
\label{tab:appendix_class_splits_part_1}
\end{table*}
\begin{table*}[t]
\centering
\resizebox{0.85\textwidth}{!}{
\begin{tabular}{lllccr}
\toprule
&  & \multicolumn{4}{c}{Results of a model with the max OOD-ID} \\
\cmidrule{3-6}
\textbf{Train} & \textbf{Test} & \textbf{Model} & \textbf{ID} & \textbf{OOD} & \textbf{OOD-ID} \\
\midrule
SNLI & NLI Diagnostics named entities & OPT-13B & 91.38 & 66.67 & -24.71 \\
SNLI & NLI Diagnostics wikipedia & DeBERTa-v3-Large & 92.24 & 67.33 & -24.91 \\
SNLI & NLI Diagnostics common sense & DeBERTa-v3-Large & 92.24 & 67.33 & -24.91 \\
SNLI & NLI Diagnostics world knowledge & OPT-13B & 91.38 & 66.42 & -24.96 \\
SNLI & NLI Diagnostics acl & DeBERTa-v3-Large & 92.24 & 66.67 & -25.57 \\
SNLI & NLI Diagnostics non-monotone & OPT-6.7B & 90.41 & 63.33 & -27.08 \\
SNLI & NLI Diagnostics existential & T5-3B & 92.16 & 65.00 & -27.16 \\
SNLI & NLI Diagnostics active/passive & OPT-125M & 86.22 & 58.82 & -27.40 \\
SNLI & NLI Diagnostics factivity & DeBERTa-v3-Large & 92.24 & 64.71 & -27.53 \\
SNLI & NLI Diagnostics temporal & DeBERTa-v3-Large & 92.24 & 64.58 & -27.66 \\
SNLI & NLI Diagnostics reddit & T5-3B & 92.16 & 64.33 & -27.83 \\
SNLI & NLI Diagnostics intersectivity & RoBERTa-Base & 90.37 & 60.87 & -29.50 \\
SNLI & DNLI & T5-Large & 91.20 & 60.35 & -30.85 \\
SNLI & NLI Diagnostics intervals/numbers & T5-11B & 92.13 & 60.53 & -31.60 \\
QQP  & PAWS QQP& T5-11B & 92.11 & 50.81 & -41.30 \\
SNLI & NLI Diagnostics disjunction & OPT-13B & 91.38 & 47.37 & -44.01 \\
SNLI & NLI Diagnostics downward monotone & T5-11B & 92.13 & 23.33 & -68.80 \\
\bottomrule
\end{tabular}%
}
\caption{The OOD evaluation of \underline{classification} \textbf{finetuned} models (continuation of Table \ref{tab:appendix_class_splits_part_1}).}
\label{tab:appendix_class_splits_part_2}
\end{table*}

\begin{table*}[t]
\centering
\resizebox{0.75\textwidth}{!}{
\begin{tabular}{lllccr}
\toprule
&  & \multicolumn{4}{c}{Results of a model with the max OOD-ID} \\
\cmidrule{3-6}
\textbf{Train} & \textbf{Test} & \textbf{Model} & \textbf{ID} & \textbf{OOD} & \textbf{OOD-ID} \\
\midrule
NewsQA & SQuAD & T5-Large & 21.78 & 87.92 & 66.14 \\
NewsQA & RelationExtraction & T5-Large & 21.78 & 79.62 & 57.84 \\
NewsQA & NaturalQuestions & T5-11B & 23.06 & 64.40 & 41.34 \\
NewsQA & BioASQ & T5-11B & 23.06 & 58.24 & 35.18 \\
NewsQA & RACE & T5-11B & 23.06 & 52.73 & 29.67 \\
NewsQA & TriviaQA & OPT-6.7B & 60.33 & 79.74 & 19.41 \\
NaturalQuestions & SQuAD & T5-11B & 71.64 & 88.83 & 17.19 \\
NaturalQuestions & RelationExtraction & T5-11B & 71.64 & 87.12 & 15.48 \\
NewsQA & DROP & T5-11B & 23.06 & 35.02 & 11.96 \\
NewsQA & DuoRC & GPT-2-XL & 59.74 & 63.11 & 3.37 \\
NaturalQuestions & TriviaQA & OPT-6.7B & 80.38 & 82.98 & 2.60 \\
SQuAD & RelationExtraction & GPT-2 & 76.13 & 74.86 & -1.27 \\
NewsQA & TextbookQA & OPT-6.7B & 60.33 & 57.56 & -2.77 \\
\midrule
NaturalQuestions & DROP & T5-11B & 71.64 & 67.47 & -4.17 \\
BoolQ & MultiRC & OPT-350M & 68.60 & 63.91 & -4.69 \\
SQuAD & NoiseQA & RoBERTa-Large & 92.46 & 86.21 & -6.25 \\
SQuAD & MLQA & OPT-350M & 82.33 & 75.24 & -7.09 \\
SQuAD & TriviaQA & OPT-13B & 91.33 & 83.56 & -7.77 \\
NaturalQuestions & BioASQ & T5-11B & 71.64 & 62.89 & -8.75 \\
BoolQ & BoolQ Contrast Set & OPT-125M & 66.54 & 57.70 & -8.84 \\
BoolQ & BoolQ CAD & T5-3B & 89.23 & 78.09 & -11.14 \\
NaturalQuestions & DuoRC & OPT-6.7B & 80.38 & 68.78 & -11.60 \\
NewsQA & QuAC & T5-11B & 23.06 & 10.99 & -12.07 \\
NaturalQuestions & TextbookQA & OPT-6.7B & 80.38 & 67.29 & -13.09 \\
NaturalQuestions & RACE & T5-11B & 71.64 & 56.64 & -15.00 \\
NaturalQuestions & NewsQA & DeBERTa-v3-Large & 56.83 & 40.60 & -16.23 \\
SQuAD & HotPotQA & OPT-6.7B & 91.79 & 73.62 & -18.17 \\
SQuAD & DuoRC & OPT-13B & 91.33 & 71.32 & -20.01 \\
SQuAD & BioASQ & OPT-6.7B & 91.79 & 68.65 & -23.14 \\
SQuAD & NewsQA & OPT-13B & 91.33 & 66.50 & -24.83 \\
SQuAD & NaturalQuestions & OPT-6.7B & 91.79 & 65.63 & -26.16 \\
SQuAD & DROP & T5-11B & 93.80 & 65.36 & -28.44 \\
SQuAD & TextbookQA & OPT-2.7B & 91.56 & 62.57 & -28.99 \\
SQuAD & SearchQA & OPT-2.7B & 91.56 & 59.94 & -31.62 \\
SQuAD & RACE & T5-11B & 93.80 & 57.05 & -36.75 \\
NaturalQuestions & QuAC & DeBERTa-v3-Large & 56.83 & 11.86 & -44.97 \\
SQuAD & QuAC & GPT-2 & 76.13 & 16.44 & -59.69 \\
\bottomrule
\end{tabular}
}
\caption{The OOD evaluation of \underline{reading comprehension} \textbf{finetuned} models. All results are F1 scores, except for models trained with BoolQ for which we report exact match (EM) following the standard practice. For each train-test split, we report the model with the best OOD generalization (among 19 models), defined as the highest OOD F1/EM minus ID F1/EM. In contrast to sentiment classification and NLI, no reading comprehension model achieves an F1/exact match score over 90\% while also maintaining similar OOD and ID accuracies; see the upper part of the table. Thus, all reading comprehension splits remain suitable.}
\label{tab:appendix_qa_splits}
\end{table*}

\begin{table*}[t]
\centering
\resizebox{0.7\textwidth}{!}{%
\begin{tabular}{lllll}
\toprule
\textbf{Train} & \textbf{Test} & \textbf{ID} & \textbf{OOD} & \textbf{OOD-ID} \\
\midrule
SNLI & NLI Diagnostics redundancy & 45.88 & 96.15 & 50.27 \\
SNLI & NLI Diagnostics genitives/partitives & 45.88 & 85.00 & 39.12 \\
FEVER & FEVER-symmetric v1 & 36.08 & 55.79 & 19.71 \\
SNLI & NLI Diagnostics restrictivity & 45.88 & 65.38 & 19.50 \\
SNLI & NLI Diagnostics morphological negation & 45.88 & 65.38 & 19.50 \\
SNLI & NLI Diagnostics nominalization & 45.88 & 64.29 & 18.41 \\
SNLI & NLI Diagnostics double negation & 45.88 & 64.29 & 18.41 \\
FEVER & FEVER-symmetric v2 & 36.08 & 53.37 & 17.29 \\
QNLI & Adversarial Squad & 55.81 & 71.29 & 15.48 \\
SNLI & NLI Diagnostics symmetry/collectivity & 45.88 & 60.71 & 14.83 \\
SST2 & YELP & 65.94 & 80.11 & 14.17 \\
SST2 & AmazonReviews & 65.94 & 75.86 & 9.92 \\
SNLI & NLI Diagnostics core args & 45.88 & 55.77 & 9.89 \\
SNLI & NLI Diagnostics named entities & 45.88 & 55.56 & 9.68 \\
SNLI & Stress Test & 67.83 & 75.34 & 7.51 \\
YELP & IMDB & 81.54 & 88.98 & 7.44 \\
SNLI & NLI Diagnostics ellipsis/implicits & 45.88 & 52.94 & 7.06 \\
SNLI & NLI Diagnostics world knowledge & 45.88 & 52.24 & 6.36 \\
SNLI & NLI Diagnostics acl & 45.88 & 50.50 & 4.62 \\
SNLI & NLI Diagnostics wikipedia & 45.88 & 50.50 & 4.62 \\
SNLI & NLI Diagnostics news & 45.88 & 50.49 & 4.61 \\
SNLI & HANS Constituent & 45.88 & 50.04 & 4.16 \\
SNLI & HANS Subsequence & 45.88 & 50.02 & 4.14 \\
SNLI & HANS Lexical Overlap & 45.88 & 50.01 & 4.13 \\
SNLI & NLI Diagnostics conditionals & 45.88 & 50.00 & 4.12 \\
SNLI & NLI Diagnostics universal & 45.88 & 50.00 & 4.12 \\
MNLI & HANS Constituent & 46.68 & 50.04 & 3.36 \\
MNLI & HANS Subsequence & 46.68 & 50.02 & 3.34 \\
MNLI & HANS Lexical Overlap & 46.68 & 50.01 & 3.33 \\
SNLI & NLI Diagnostics nan & 45.88 & 48.65 & 2.77 \\
SNLI & NLI Diagnostics prepositional phrases & 45.88 & 48.53 & 2.65 \\
IMDB & YELP & 95.10 & 97.48 & 2.38 \\
QQP & Twitter PPDB & 78.36 & 80.15 & 1.79 \\
SST2 & RottenTomatoes & 65.94 & 67.73 & 1.79 \\
MNLI & MNLI-Hard & 46.68 & 48.08 & 1.40 \\
SNLI & MNLI-Mismatched & 45.88 & 47.21 & 1.33 \\
IMDB & AmazonReviews & 95.10 & 95.94 & 0.84 \\
SNLI & MNLI-Matched & 45.88 & 46.14 & 0.26 \\
SNLI & NLI Diagnostics common sense & 45.88 & 46.00 & 0.12 \\
SST2 & IMDB & 93.23 & 93.22 & -0.01 \\
SNLI & NLI Diagnostics lexical entailment & 45.88 & 45.71 & -0.17 \\
MNLI & SNLI & 59.86 & 59.52 & -0.34 \\
SNLI & NLI Diagnostics existential & 45.88 & 45.00 & -0.88 \\
SNLI & NLI Diagnostics datives & 45.88 & 45.00 & -0.88 \\
SNLI & NLI Diagnostics conjunction & 45.88 & 45.00 & -0.88 \\
IMDB & SST2 & 95.10 & 93.69 & -1.41 \\
SNLI & NLI Diagnostics active/passive & 45.88 & 44.12 & -1.76 \\
SNLI & NLI Diagnostics relative clauses & 45.88 & 43.75 & -2.13 \\
IMDB & RottenTomatoes & 95.10 & 92.40 & -2.70 \\
\bottomrule
\end{tabular}%
}
\caption{The OOD evaluation of \textbf{few-shot in-context learning} Mistral-7B for \underline{classification} tasks (continued in Table \ref{tab:appendix_class_splits_mistral_part_2}). All results are accuracy scores. For only a few splits Mistral-7B achieves an in-domain accuracy over 90\% while also maintaining similar OOD and ID accuracies. These splits might no longer be fitting for OOD robustness research.}
\label{tab:appendix_class_splits_mistral_part_1}
\end{table*}
\begin{table*}[t]
\centering
\resizebox{0.7\textwidth}{!}{%
\begin{tabular}{lllll}
\toprule
\textbf{Train} & \textbf{Test} & \textbf{ID} & \textbf{OOD} & \textbf{OOD-ID} \\
\midrule
MRPC & PAWS-Wiki & 61.76 & 58.64 & -3.12 \\
YELP & SST2 & 98.10 & 94.95 & -3.15 \\
SNLI & NLI Diagnostics negation & 67.83 & 64.63 & -3.20 \\
SNLI & NLI Diagnostics reddit & 45.88 & 42.50 & -3.38 \\
SNLI & NLI Diagnostics upward monotone & 45.88 & 41.18 & -4.70 \\
MNLI & DNLI & 59.86 & 54.66 & -5.20 \\
SNLI & NLI Diagnostics disjunction & 45.88 & 39.47 & -6.41 \\
SNLI & NLI Diagnostics anaphora/coreference & 45.88 & 37.93 & -7.95 \\
SNLI & NLI Diagnostics artificial & 45.88 & 37.33 & -8.55 \\
SNLI & NLI Diagnostics intersectivity & 45.88 & 36.96 & -8.92 \\
SNLI & NLI Diagnostics quantifiers & 45.88 & 36.54 & -9.34 \\
SNLI & DNLI & 45.88 & 36.43 & -9.45 \\
QQP & PAWS-QQP & 40.67 & 30.28 & -10.39 \\
SNLI & NLI Diagnostics downward monotone & 45.88 & 33.33 & -12.55 \\
SNLI & NLI Diagnostics temporal & 45.88 & 31.25 & -14.63 \\
SNLI & NLI Diagnostics coordination scope & 45.88 & 30.00 & -15.88 \\
SNLI & NLI Diagnostics intervals/numbers & 45.88 & 28.95 & -16.93 \\
SNLI & NLI Diagnostics factivity & 45.88 & 26.47 & -19.41 \\
SNLI & NLI Diagnostics non-monotone & 45.88 & 23.33 & -22.55 \\
IMDB & Twitter Emotion & 95.10 & 70.59 & -24.51 \\
\bottomrule
\end{tabular}%
}
\caption{The OOD evaluation of \textbf{few-shot in-context learning} Mistral-7B for \underline{classification} tasks (continuation of Table \ref{tab:appendix_class_splits_part_1}).}
\label{tab:appendix_class_splits_mistral_part_2}
\end{table*}

\begin{table*}[t]
\centering
\resizebox{0.65\textwidth}{!}{%
\begin{tabular}{lllll}
\toprule
\textbf{Train} & \textbf{Test} & \textbf{ID} & \textbf{OOD} & \textbf{OOD - ID} \\
\midrule
NaturalQuestions & TextbookQA & 31.94 & 66.65 & 34.71 \\
NaturalQuestions & TriviaQA & 31.94 & 65.79 & 33.85 \\
NewsQA & RelationExtraction & 60.77 & 90.52  & 29.75\\
NewsQA & SQuAD & 60.77 &  89.51 &  28.74\\
NewsQA & TriviaQA & 60.77 & 85.69 & 24.92 \\
NaturalQuestions & SQuAD & 31.94 & 54.28 & 22.34 \\
NewsQA & TextbookQA & 60.77 & 81.76 & 20.99\\
NewsQA & BioASQ & 60.77 & 80.24 & 19.47\\
NaturalQuestions & BioASQ & 31.94 & 48.40 & 16.46 \\
NaturalQuestions & RelationExtraction & 66.86 & 82.53 & 15.67 \\
NewsQA & NaturalQuestions & 60.77 & 74.82 & 14.05\\
SQuAD & TextbookQA & 54.28 & 66.65 & 12.37 \\
SQuAD & TriviaQA & 54.28 & 65.79 & 11.51 \\
NaturalQuestions & DROP & 31.94 & 42.16 & 10.22 \\
NaturalQuestions & NewsQA & 31.94 & 42.08 & 10.14 \\
NaturalQuestions & DuoRC & 31.94 & 41.85 & 9.91 \\
SQuAD & HotPotQA & 54.28 & 63.11 & 8.83 \\
NewsQA & DuoRC & 60.77 & 69.29 & 8.52\\
SQuAD & SearchQA & 54.28 & 60.36 & 6.08 \\
NewsQA & DROP & 60.77 & 65.40 & 4.63\\
NewsQA & RACE & 60.77 & 63.78 & 3.01 \\
NaturalQuestions & RACE & 31.94 & 34.08 & 2.14 \\
BoolQ & BoolQ & 83.52 & 83.52 & 0.00 \\
SQuAD & RelationExtraction & 84.01 & 82.53 & -1.48 \\
SQuAD & NoiseQA & 54.28 & 50.44 & -3.84 \\
SQuAD & BioASQ & 54.28 & 48.40 & -5.88 \\
BoolQ & MultiRC & 83.52 & 76.38 & -7.14 \\
SQuAD & MLQA & 54.28 & 46.12 & -8.16 \\
SQuAD & DROP & 54.28 & 42.16 & -12.12 \\
SQuAD & NewsQA & 54.28 & 42.08 & -12.20 \\
SQuAD & DuoRC & 54.28 & 41.85 & -12.43 \\
BoolQ & BoolQ CAD & 83.52 & 69.38 & -14.14 \\
SQuAD & NaturalQuestions & 84.01 & 66.86 & -17.15 \\
NaturalQuestions & QuAC & 31.94 & 12.72 & -19.22 \\
SQuAD & RACE & 54.28 & 34.08 & -20.20 \\
BoolQ & BoolQ Contrast Set & 83.52 & 52.48 & -31.04 \\
SQuAD & QuAC & 54.28 & 12.72 & -41.56\\
\bottomrule
\end{tabular}%
}
\caption{The OOD evaluation of \textbf{few-shot in-context learning} Mistral-7B for \underline{reading comprehension} tasks. All results are F1 scores, except for models trained with BoolQ for which we report exact match (EM) following the standard practice.}
\label{tab:appendix_qa_splits_mistral}
\end{table*}

\clearpage

\begin{table}[H]
\centering
\resizebox{0.9\columnwidth}{!}{%
\begin{tabular}{llrr}
\toprule
& \textbf{Test} (\textbf{Category}) & \textbf{Accuracy} & \textbf{Diff} \\
\midrule
\multirow{4}{*}{\rotatebox{90}{\small \emph{Sentiment Class.}}}  & C-IMDB & 92.6 & 1.2 \\
\arrayrulecolor{black!20}\cmidrule{2-4}
& IMDB Contrast (all)  & 93.7 & -4.3\\
& IMDB Contrast  (contrast) & 92.8 & 1.0\\
& IMDB Contrast  (original) & 94.5 & -0.7 \\
\arrayrulecolor{black}\midrule
\multirow{14}{*}{\rotatebox{90}{\small \emph{Natural Language Inference}}}  & \cellcolor{mycolor} ANLI (r1) & \cellcolor{mycolor} 48.0 & \cellcolor{mycolor} 11.7\\
& \cellcolor{mycolor} ANLI (r2) & \cellcolor{mycolor} 44.9 & \cellcolor{mycolor} 14.8\\
& \cellcolor{mycolor} ANLI (r3) & \cellcolor{mycolor} 45.6 & \cellcolor{mycolor} 14.1\\
\arrayrulecolor{black!20}\cmidrule{2-4}
& Breaking NLI  & 77.8 & -18.1 \\
\arrayrulecolor{black!20}\cmidrule{2-4}
& \cellcolor{mycolor}HANS (all)& \cellcolor{mycolor} 56.2 & \cellcolor{mycolor} 3.5\\
& \cellcolor{mycolor}HANS (constituent) & \cellcolor{mycolor} 55.5 & \cellcolor{mycolor} 4.2\\
& HANS (lexical overlap) & 58.9 & 0.8 \\
& \cellcolor{mycolor}HANS (subsequence) & \cellcolor{mycolor} 54.3 & \cellcolor{mycolor} 5.4\\
\arrayrulecolor{black!20}\cmidrule{2-4}
& \cellcolor{mycolor}MNLI-hard (val matched) & \cellcolor{mycolor} 55.0 & \cellcolor{mycolor} 4.7\\
& MNLI-hard (val mismatched) & 57.1 & 2.6\\
\arrayrulecolor{black!20}\cmidrule{2-4}
& \cellcolor{mycolor}NLI Diagnostics (min--max) & \cellcolor{mycolor} 32.1--71.4 & \cellcolor{mycolor} 27.6 / -11.7\\
\arrayrulecolor{black!20}\cmidrule{2-4}
& \cellcolor{mycolor}Stress Test (min--max) & \cellcolor{mycolor} 37.9--76.2 & \cellcolor{mycolor} 21.8 / 16.5 \\
\arrayrulecolor{black!20}\cmidrule{2-4}
& \cellcolor{mycolor} SNLI CAD  & \cellcolor{mycolor} 59.9 & \cellcolor{mycolor} 7.9\\
\arrayrulecolor{black!20}\cmidrule{2-4}
& \cellcolor{mycolor}SNLI-hard & \cellcolor{mycolor} 63.2  & \cellcolor{mycolor}4.6\\
\arrayrulecolor{black}\midrule
& \cellcolor{mycolor} PAWS-QQP  & \cellcolor{mycolor} 57.2& \cellcolor{mycolor} 21.2\\
\arrayrulecolor{black}\midrule
\multirow{11}{*}{\rotatebox{90}{\small \emph{Reading Comprehension}}} & AddOneSent &	74.2 &	\underline{9.9} \\
&  \cellcolor{mycolor}AddSent &	\cellcolor{mycolor} 74.1	& \cellcolor{mycolor} 9.9\\ 
&  \cellcolor{mycolor} Adversarial Paraphrased &	\cellcolor{mycolor} 76.5		& \cellcolor{mycolor} 7.5\\ 
& \cellcolor{mycolor} BoolQ CAD &	\cellcolor{mycolor} 69.9 & \cellcolor{mycolor} 14.6	 \\
& \cellcolor{mycolor} BoolQ Contrast Set &	\cellcolor{mycolor} 52.5	& \cellcolor{mycolor} 31.0\\
& \cellcolor{mycolor} MultiRC & \cellcolor{mycolor} 76.4		& \cellcolor{mycolor} 7.1\\
& \cellcolor{mycolor} NaturalQuestions &	\cellcolor{mycolor} 66.9	& \cellcolor{mycolor} 17.2	\\
& \cellcolor{mycolor} NewsQA &	\cellcolor{mycolor} 56.2	& \cellcolor{mycolor} 27.9\\ 
& Non-Adversarial Paraphrased &	92.6	& 1.4 \\
& \cellcolor{mycolor} Quoref & \cellcolor{mycolor} 71.5	& \cellcolor{mycolor} 12.5	 \\
& SQuAD-hard &	80.5	& 3.5	\\ 
\arrayrulecolor{black}\bottomrule
\end{tabular}%
}
\caption{The challenge set performance of in-context learning with Mistral-7B (8-shots). The last column shows the difference between performance in-domain and in the challenge set (higher means poorer generalization). Shaded rows highlight datasets that remain challenging for associated models.}
\label{tab:icl_challenge_set_results}
\end{table}

\begin{table}[H]
\centering
\resizebox{0.9\columnwidth}{!}{%
\begin{tabular}{lrr}
\toprule
\textbf{Test} (\textbf{Category}) & \textbf{Accuracy} & \textbf{Diff} \\
\midrule

\rowcolor{mycolor} ANLI (r1) & 70.1 & 8.2\\
\rowcolor{mycolor} ANLI (r2) & 58.9 & 19.4\\
\rowcolor{mycolor} ANLI (r3) & 52.7 & 25.6\\
\arrayrulecolor{black!20}\midrule
Breaking NLI  & 78.3 & 0.0 \\
\arrayrulecolor{black!20}\midrule
HANS (all)& 92.0 & -13.7\\
HANS (constituent) & 79.1 & -0.8\\
HANS (lexical overlap) & 99.0 & -20.7 \\
HANS (subsequence) & 97.8 & -19.5\\
\arrayrulecolor{black!20}\midrule
MNLI-hard (val matched) & 79.3 & -1.0\\
MNLI-hard (val mismatched) & 78.8 & -0.5\\
\arrayrulecolor{black!20}\midrule
\rowcolor{mycolor}NLI Diagnostics (min--max) & 40.0--100.0 & 38.3 / -21.7\\
\arrayrulecolor{black!20}\midrule
\rowcolor{mycolor}Stress Test (min--max) & 43.2--82.5 & 35.1 / -4.2 \\
\arrayrulecolor{black!20}\midrule
SNLI CAD  & 76.2 & 2.1\\
\arrayrulecolor{black!20}\midrule
SNLI-hard & 74.1 & 4.2\\
\arrayrulecolor{black}\bottomrule
\end{tabular}%
}
\caption{The challenge set accuracy for the T5-11B model trained on WANLI~\citep{liu-etal-2022-wanli}. %
}
\label{tab:challenge_set_t5_11b_wanli}
\end{table}

\begin{table*}[t]
\centering
\resizebox{0.8\textwidth}{!}{%
\begin{tabular}{llll}
\toprule
\textbf{Test} & \textbf{Category} & \textbf{Sub-Category} & \textbf{Accuracy} \\
\midrule
NLI Diagnostics &  &  & 95.0 \\
\rowcolor{mycolor} NLI Diagnostics & acl & Domain & 67.7\\
\rowcolor{mycolor} NLI Diagnostics & artificial & Domain & 73.3\\
\rowcolor{mycolor} NLI Diagnostics & news & Domain & 75.0\\
\rowcolor{mycolor} NLI Diagnostics & reddit & Domain & 65.2\\
\rowcolor{mycolor} NLI Diagnostics & wikipedia & Domain & 70.5\\
\rowcolor{mycolor} NLI Diagnostics & common sense & Knowledge & 70.0\\
\rowcolor{mycolor} NLI Diagnostics & world knowledge & Knowledge & 66.4\\
\rowcolor{mycolor} NLI Diagnostics & factivity & Lexical Semantics & 68.6\\
\rowcolor{mycolor} NLI Diagnostics & lexical entailment & Lexical Semantics & 76.4\\
NLI Diagnostics & morphological negation & Lexical Semantics & 91.0 \\
\rowcolor{mycolor} NLI Diagnostics & named entities & Lexical Semantics & 70.4\\
\rowcolor{mycolor} NLI Diagnostics & quantifiers & Lexical Semantics & 82.7\\
NLI Diagnostics & redundancy & Lexical Semantics & 88.5 \\
\rowcolor{mycolor} NLI Diagnostics & symmetry/collectivity & Lexical Semantics & 73.8\\
\rowcolor{mycolor} NLI Diagnostics & conditionals & Logic & 75.0\\
NLI Diagnostics & conjunction & Logic & 87.5 \\
\rowcolor{mycolor} NLI Diagnostics & disjunction & Logic & 47.4\\
NLI Diagnostics & double negation & Logic & 94.1 \\
\rowcolor{mycolor} NLI Diagnostics & downward monotone & Logic & 30.0\\
\rowcolor{mycolor} NLI Diagnostics & existential & Logic & 73.3\\
\rowcolor{mycolor} NLI Diagnostics & intervals/numbers & Logic & 63.2\\
\rowcolor{mycolor} NLI Diagnostics & negation & Logic & 76.8\\
\rowcolor{mycolor} NLI Diagnostics & non-monotone & Logic & 63.3\\
\rowcolor{mycolor} NLI Diagnostics & temporal & Logic & 71.9\\
NLI Diagnostics & universal & Logic & 94.4 \\
\rowcolor{mycolor} NLI Diagnostics & upward monotone & Logic & 79.4\\
\rowcolor{mycolor} NLI Diagnostics & active/passive & Predicate-Argument Structure & 62.8\\
\rowcolor{mycolor} NLI Diagnostics & anaphora/coreference & Predicate-Argument Structure & 74.7\\
\rowcolor{mycolor} NLI Diagnostics & coordination scope & Predicate-Argument Structure & 72.5\\
\rowcolor{mycolor} NLI Diagnostics & core args & Predicate-Argument Structure & 76.3\\
NLI Diagnostics & datives & Predicate-Argument Structure & 85.0 \\
\rowcolor{mycolor} NLI Diagnostics & ellipsis/implicits & Predicate-Argument Structure & 81.4\\
NLI Diagnostics & genitives/partitives & Predicate-Argument Structure & 95.0 \\
\rowcolor{mycolor} NLI Diagnostics & intersectivity & Predicate-Argument Structure & 63.0\\
NLI Diagnostics & nominalization & Predicate-Argument Structure & 92.9 \\
NLI Diagnostics & prepositional phrases & Predicate-Argument Structure & 86.8 \\
\rowcolor{mycolor} NLI Diagnostics & relative clauses & Predicate-Argument Structure & 68.8\\
\rowcolor{mycolor} NLI Diagnostics & restrictivity & Predicate-Argument Structure & 69.2\\
\midrule
Stress Test & validation matched &  & 90.6 \\
Stress Test & validation matched & Antonym & 89.1 \\
Stress Test & validation matched & Length\_Mismatch & 89.5 \\
\rowcolor{mycolor}Stress Test & validation matched & Negation & 76.8 \\
Stress Test & validation matched & Spelling Error (contentword\_swap\_perturbed) & 89.7\\
Stress Test & validation matched & Spelling Error (functionword\_swap\_perturbed) & 90.6 \\
Stress Test & validation matched & Spelling Error (keyboard) & 90.1 \\
Stress Test & validation matched & Spelling Error (swap) & 90.2 \\
Stress Test & validation matched & Word\_Overlap & 83.1 \\
Stress Test & validation mismatched &  & 90.5 \\
Stress Test & validation mismatched & Antonym & 88.4 \\
Stress Test & validation mismatched & Length Mismatch & 90.2 \\
\rowcolor{mycolor}Stress Test & validation mismatched & Negation & 77.4 \\
Stress Test & validation mismatched & Spelling Error (contentword\_swap\_perturbed) & 89.6\\
Stress Test & validation mismatched & Spelling Error (functionword\_swap\_perturbed) & 90.5 \\
Stress Test & validation mismatched & Spelling Error (keyboard) & 89.5 \\
Stress Test & validation mismatched & Spelling Error (swap) & 90.5 \\
Stress Test & validation mismatched & Word Overlap & 82.9\\
\bottomrule
\end{tabular}%
}
\caption{The breakdown of individual tests in NLI Diagnostics and Stress Test.}
\label{tab:nli_diagnostics_stress_test}
\end{table*}

\begin{figure}[!ht]
    \centering
    \begin{minipage}[]{\columnwidth}
        \centering
        \begin{subfigure}{\columnwidth}
            \includegraphics[width=\linewidth]{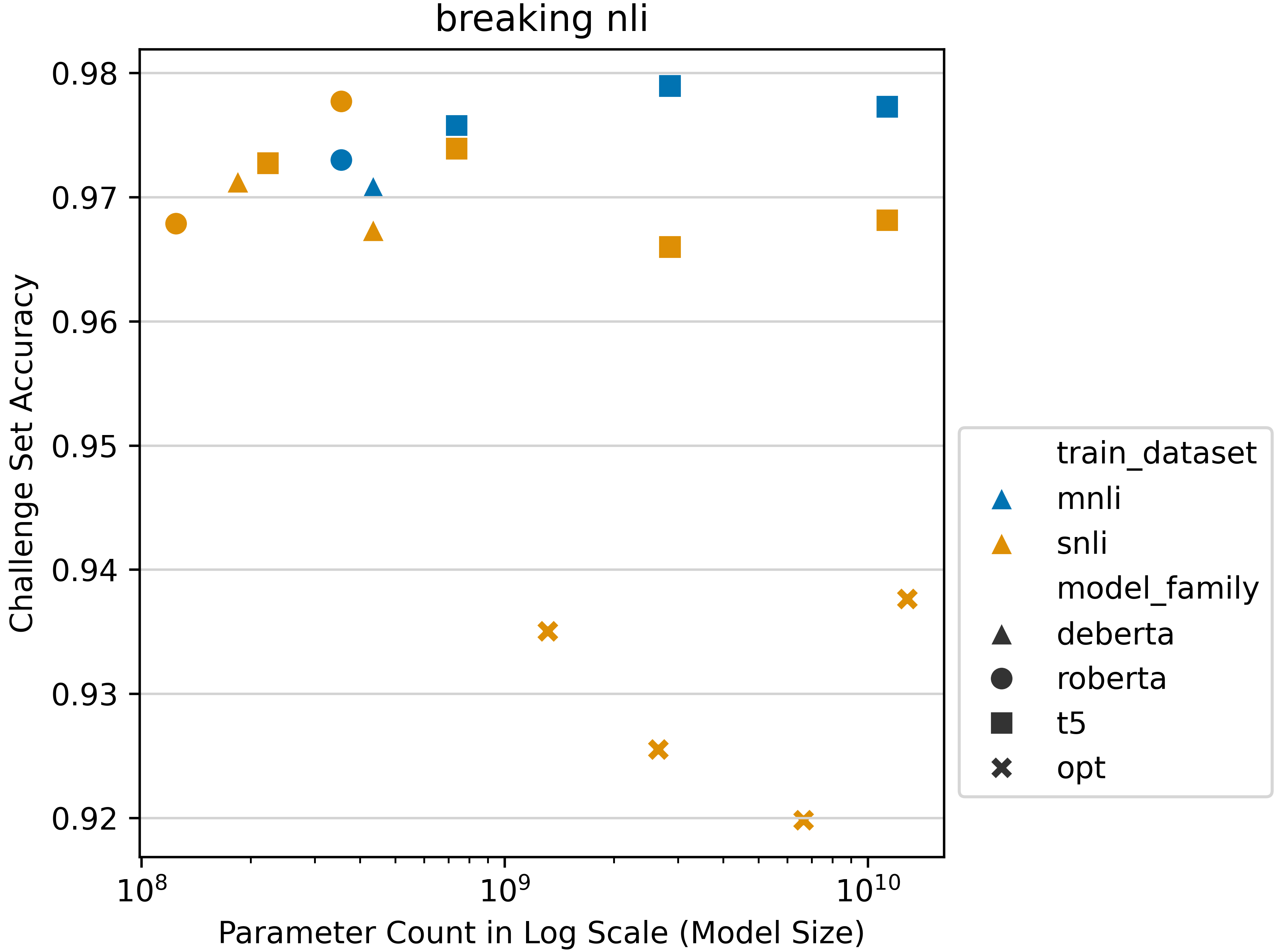}
        \end{subfigure}
    \end{minipage}
    \begin{minipage}[]{\columnwidth}
        \centering
        \begin{subfigure}{\columnwidth}
            \includegraphics[width=\linewidth]{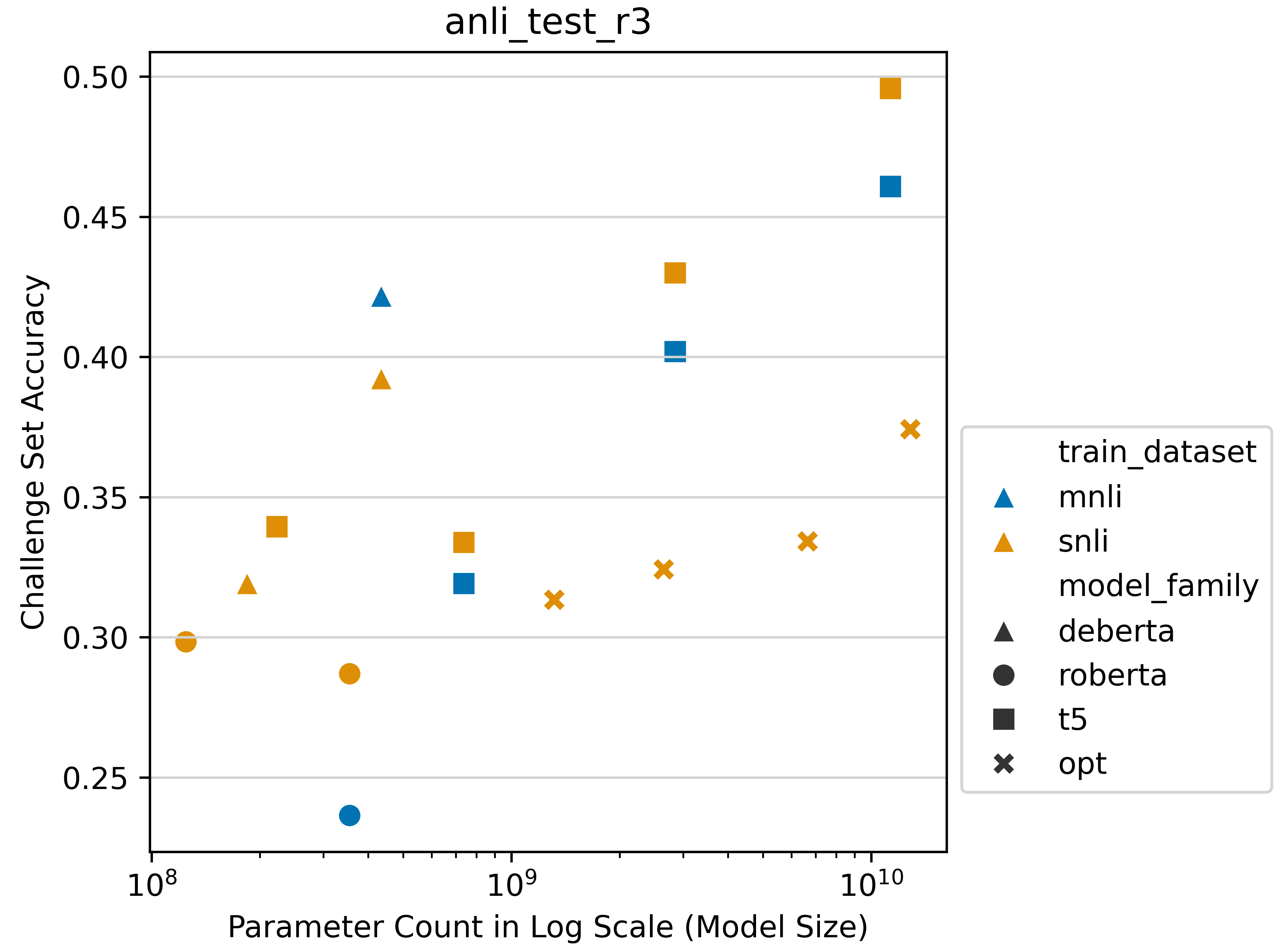}
        \end{subfigure}
    \end{minipage}
    \caption{Breaking NLI and ANLI (r3) accuracy of all models with the in-domain accuracy accuracy of 85\% or more. Breaking NLI shows that models of different types and sizes can get high challenge set accuracy, while ANLI shows that none of them can.}
    \label{fig:challenge_main}
\end{figure}

\clearpage

\begin{table*}[t]
\centering
\resizebox{\textwidth}{!}{%
\begin{tabular}{lllp{20cm}}
\toprule
\multicolumn{4}{l}{\textbf{Test Id; Capability; Test Type; \emph{Test Description}: Q1: \texttt{\{question\}} Q2: \texttt{\{question\}} (\{label\}) }} \\
\midrule
1  & \multirow{11}{*}{\rotatebox[origin=c]{90}{Vocabulary}}\ & MFT  & \emph{Add an adjective (modifier)}:  Q1: Is Adam Ward a historian? Q2: Is Adam Ward an aspiring historian? (not duplicates)  \\
\arrayrulecolor{black!20}\cmidrule{3-4}
2  &  & MFT  & \emph{Different adjectives}: Q1: Is Jason Price an immigrant? Q2: Is Jason Price Indian? (not duplicates)   \\
\arrayrulecolor{black!20}\cmidrule{3-4}
3  & \ & MFT  & \emph{Different animals}:   Q1: Can I feed my dog cereal? Q2: Can I feed my snake cereal? (not duplicates)   \\
\arrayrulecolor{black!20}\cmidrule{3-4}
4  &  & MFT  & \emph{Add irrelevant modifiers (examples with animals)}: Q1: Is that monkey up on the table? Q2: Is that monkey truly up on the table? (duplicates)\\
\arrayrulecolor{black!20}\cmidrule{3-4}
5  & & MFT  & \emph{Add irrelevant modifiers (examples with people)}: Q1: Is Melissa responding to Christina? Q2: Is Melissa really responding to Christina? (duplicates)\\
\arrayrulecolor{black!20}\cmidrule{3-4}
6  &  & MFT  & \emph{Different irrelevant preamble}: Q1: My pet cat eats soy. Is it normal for animals to eat soy? Q2: My pet monkey eats soy. Is it normal for animals to eat soy? (duplicates) \\
\arrayrulecolor{black!20}\cmidrule{3-4}
7  &  & MFT  & \emph{Preamble is relevant (different injuries)}: Q1: I hurt my hip last time I played football. Is this a common injury? Q2: I hurt my thigh last time I played football. Is this a common injury? (not duplicates) \\
\arrayrulecolor{black}\midrule
8  & \multirow{10}{*}{\rotatebox[origin=c]{90}{Taxonomy}}   & MFT  & \emph{How can I become more \{synonym\}?}: Q1: How can I become more religious? Q2: How can I become more spiritual?  (duplicates) \\
\arrayrulecolor{black!20}\cmidrule{3-4}
9  &    & INV  & \texttt{(question, f(question))} \emph{ where }\texttt{f(question)}\emph{ replaces synonyms?}: Q1: I am a 32 year old single man, doing a govt job in India, not happy with my job and life, nothing much in my bank account, what should I do? Q2: I am a 32 year old single man, doing a govt job in India, not joyful with my job and life, nothing much in my bank account, what should I do? (duplicates) \\
\arrayrulecolor{black!20}\cmidrule{3-4}
10 &    & INV  & \emph{Replace synonyms in pairs of duplicates from the dev set}: Q1: What is the secret of happy life? Q2: What's the bloody secret of a happy life? (duplicates) || Q1: What is the secret of joyful life? Q2: What's the bloody secret of a happy life?  (duplicates) \\
\arrayrulecolor{black!20}\cmidrule{3-4}
11 & & MFT  & \emph{How can I become more X $\neq$ How can I become less X}: Q1: How can I become less secular? Q2: How can I become more secular? (not duplicates) \\
\arrayrulecolor{black!20}\cmidrule{3-4}
12 &    & MFT  & \emph{How can I become more X = How can I become less antonym(X)}: Q1: How can I become less hopeful? Q2: How can I become more hopeless?  (not duplicates) \\
\arrayrulecolor{black}\midrule
13 & \multirow{11}{*}{\rotatebox[origin=c]{90}{Robustness}}  & INV  & \emph{Add one typo:} Q1: What are the best tisp for early-stage startups? Q2: What are your best tips for very early stage startups? (duplicates) || Q1: What are the best tips for early-stage startups? Q2: What are your best tips for very early stage statrups? (duplicates) || .. \\
\arrayrulecolor{black!20}\cmidrule{3-4}
14 &  & INV  & \emph{Contractions}: Q1: What are the qualifications for being an FBI or CIA agent? Q2: What does it take to become an FBI agent? (not duplicates) || Q1: What're the qualifications for being an FBI or CIA agent? Q2: What does it take to become an FBI agent? (not duplicates)  || ... \\
\arrayrulecolor{black!20}\cmidrule{3-4}
15 &  & DIR  & \texttt{(q, paraphrase(q))}: Q1: Do you think you can use another opertator's SIM in Jio SIM slot after using Jio SIM? Q2: Can you use another opertator's SIM in Jio SIM slot after using Jio SIM? (duplicates)  \\
\arrayrulecolor{black!20}\cmidrule{3-4}
16 & & INV  & \emph{Product of \texttt{paraphrases(q1) * paraphrases(q2)}}: Q1: If you want to publish poetry on Quora, what should you do? Q2: Do you think you can post your poetry on Quora? (not duplicates)||  Q1: In order to publish poetry on Quora, what should you do? Q2: Can you post my poetry on Quora? (not duplicates) || ...    \\
\arrayrulecolor{black}\midrule 
17 & \multirow{26}{*}{\rotatebox[origin=c]{90}{NER}}         & MFT  & \emph{Same adjectives, different people}:   Q1: Is Samuel Rogers Australian? Q2: Is Joshua James Australian? (not duplicates)   \\
\arrayrulecolor{black!20}\cmidrule{3-4}
18 &         & MFT  & \emph{Same adjectives, different people v2:} Q1: Is Eric Wilson Jewish? Q2: Is Victoria Wilson Jewish? (not duplicates) \\
\arrayrulecolor{black!20}\cmidrule{3-4}
19 &         & MFT  & \emph{Same adjectives, different people v3}: Q1: Is Olivia Edwards Muslim? Q2: Is Olivia Reyes Muslim? (not duplicates)  \\
\arrayrulecolor{black!20}\cmidrule{3-4}
20 &         & INV  & \emph{Change same name in both questions:} Q1: Did Jesus keep the sabbath? Q2: When Jesus died on the cross did he do away with keeping the seventh year sabbath? (not duplicates) || Q1: Did Kyle keep the sabbath? Q2: When Kyle died on the cross did he do away with keeping the seventh year sabbath? (not duplicates) \\
\arrayrulecolor{black!20}\cmidrule{3-4}
21 &         & INV  & \emph{Change same location in both questions}: Q1: Why does the caste system persist in India? Q2: Do you support the caste system in India? ... (not duplicates)  \\
\arrayrulecolor{black!20}\cmidrule{3-4}
22 &         & INV  & \emph{Change same number in both questions}: Q1: How can I invest \$100 into myself? Q2: What is the best way to invest \$100 in todays market? (not duplicates) || Q1: How can I invest \$103 into myself? Q2: What is the best way to invest \$103 in todays market? ... (not duplicates) \\
\arrayrulecolor{black!20}\cmidrule{3-4}
23 &         & DIR  & \emph{Change first name in one of the questions}: Q1: What does Hillary Clinton think of high-skill immigration? Q2: What is Hillary Clinton’s stance on high skilled immigration? (duplicates) || Q1: What does Hillary Clinton think of high-skill immigration Q2: What is Diana Clinton’s stance on high skilled immigration? (not duplicates) || ... \\
\arrayrulecolor{black!20}\cmidrule{3-4}
24 &         & DIR  & \emph{Change first and last name in one of the questions}: Q1: Would Hillary get women's vote just because she's a female? Q2: Are there a lot of women who will vote for Hillary Clinton just because she is a woman? (duplicates) || Q1: Would Brooke get women's vote just because she's a female? Q2: Are there a lot of women who will vote for Hillary Clinton just because she is a woman? (not duplicates) || ... \\
\arrayrulecolor{black!20}\cmidrule{3-4}
25 &         & DIR  & \emph{Change location in one of the questions}: Q1: Why did India sign the Indus Water Treaty? Q2: Why did India signed Indus water treaty? (duplicates) || Q1: Why did Nauru sign the Indus Water Treaty? Q2: Why did India signed Indus water treaty? (not duplicates) || ... \\
\arrayrulecolor{black!20}\cmidrule{3-4}
26 &         & DIR  & \emph{Change numbers in one of the questions}: Q1: What do you think of abolishing 500 and 1000 Rupee Currency notes by the Indian Government? Q2: Was the decision by the Indian Government to demonetize 500 and 1000 notes right or is it a big scam? (duplicates) || Q1: What do you think of abolishing 500 and 931 Rupee Currency notes by the Indian Government? Q2: Was the decision by the Indian Government to demonetize 500 and 1000 notes right or is it a big scam? (not duplicates) || ... \\
\arrayrulecolor{black!20}\cmidrule{3-4}
27 &         & DIR  & \emph{Keep entities, fill in with gibberish}: Q1: What would have happened if Hitler hadn't declared war on the United States after Pearl Harbor? Q2: What would have happened if the United States split in two after the revolutionary war? (not duplicates) || Q1: What would have happened if the United States split in two after the revolutionary war? Q2: What divided the United States in two after the revolutionary war? (not duplicates) || ... \\
\arrayrulecolor{black}\bottomrule
\end{tabular}
}
\caption{Checklists tests 1--27 for QQP.}
\label{tab:checklist_tests_1_to_32}
\end{table*}

\begin{table*}[t]
\centering
\resizebox{\textwidth}{!}{%
\begin{tabular}{lllp{20cm}}
\toprule
\multicolumn{4}{l}{\textbf{Test Id; Capability; Test Type; \emph{Test Description}: Q1: \texttt{\{question\}} Q2: \texttt{\{question\}} (\{label\}) }} \\
\midrule
28 & \multirow{9}{*}{\rotatebox[origin=c]{90}{Temporal}}   & MFT  & \emph{Is person X $\neq$ Did person use to be X}: Q1: Is James Russell an actress? Q2: Did James Russell use to be an actress? (not duplicates) \\
\arrayrulecolor{black!20}\cmidrule{3-4}
29 &    & MFT  & \emph{Is person X $\neq$ Is person becoming X}: Q1: Is Taylor Long an investigator? Q2: Is Taylor Long becoming an investigator? (not duplicates)  \\
30 & & MFT & \emph{What was person's life before becoming X $\neq$ [...] after becoming X}: Q1: What was Kyle Ross's life before becoming an academic? Q2: What was Kyle Ross's life after becoming an academic? (not duplicates)
\\
\arrayrulecolor{black!20}\cmidrule{3-4}
31 &    & MFT  & \emph{Do you have to X your dog before Y it $\neq$ [...] after Y it}: Q1: Do you have to weigh your dog before naming it? Q2: Do you have to weigh your dog after naming it? (not duplicates)  \\
\arrayrulecolor{black!20}\cmidrule{3-4}
32 &    & MFT  & \emph{Is it \{ok, ...\} to \{smoke, ...\} after $\neq$ before}: Q1: Is it reasonable to text before 7pm? Q2: Is it reasonable to text after 7pm? (not duplicates)\\
\arrayrulecolor{black}\midrule
33 & \multirow{8}{*}{\rotatebox[origin=c]{90}{Negation}}   & MFT  & \emph{How can I become a X person $\neq$ [...] a person who is not X}: Q1: How can I become an invisible person? Q2: How can I become a person who is not invisible? (not duplicates) \\
34 & & MFT &
\emph{Is it \{ok, ...\} to \{smoke, ...\} in country $\neq$ [...] not to [...]} Q1: Is it socially acceptable to preach in Tanzania? Q2: Is it socially acceptable not to preach in Tanzania? (not duplicates) \\
\arrayrulecolor{black!20}\cmidrule{3-4}
35 &    & MFT  & \emph{What are things a \{noun\} should worry about $\neq$ [...]  not worry about}: Q1:  What are things an escort should worry about? Q2: What are things an escort should not worry about? (not duplicates) \\
\arrayrulecolor{black!20}\cmidrule{3-4}
36 &    & MFT  & \emph{How can I become a X person = [...] a person who is not antonym(X)}: Q1: How can I become a smart person? Q2: How can I become a person who is not stupid? (duplicates)  \\
\arrayrulecolor{black}\midrule
37 & \multirow{5}{*}{\rotatebox[origin=c]{90}{Coref}}      & MFT  & \emph{Simple coref (he and she)}: Q1: If Olivia and Donald were alone, do you think he would reject her? Q2: If Olivia and Donald were alone, do you think she would reject him? (not duplicates) \\
\arrayrulecolor{black!20}\cmidrule{3-4}
38 &       & MFT  & \emph{Simple coref (his and her)}: Q1: If George and Jasmine were married, would his family be happy? Q2: If George and Jasmine were married, would Jasmine's family be happy? (not duplicates) || Q1: If George and Jasmine were married, would her family be happy? Q2: If George and Jasmine were married, would George's family be happy? (not duplicates)  \\
\arrayrulecolor{black}\midrule
39 & \multirow{11}{*}{\rotatebox[origin=c]{90}{SRL}}        & MFT  & \emph{Who do X think = Who is the [...] according to X} Q1:Who do critics think is the brightest boxer in the world? Q2: Who is the brightest boxer in the world according to critics? (duplicates) \\
\arrayrulecolor{black!20}\cmidrule{3-4}
40 &         & MFT  & \emph{Order does not matter for comparison}: Q1: Are dwarves warmer than men? Q2: Are men warmer than dwarves? (not duplicates) \\
\arrayrulecolor{black!20}\cmidrule{3-4}
41 &         & MFT  & \emph{Order does not matter for symmetric relations}: Q1: Is Hannah engaged to Isabella? Q2: Is Isabella engaged to Hannah? (duplicates)  \\
\arrayrulecolor{black!20}\cmidrule{3-4}
42 &         & MFT  & \emph{Order does matter for asymmetric relations}: Q1: Is Elizabeth beating Adam? Q2: Is Adam beating Elizabeth? (not duplicates)   \\
\arrayrulecolor{black!20}\cmidrule{3-4}
43 &         & MFT  & \emph{Traditional SRL: active / passive swap}: Q1: Did Samuel miss the estate? Q2: Was the estate missed by Samuel? (duplicates)   \\
\arrayrulecolor{black!20}\cmidrule{3-4}
44 &         & MFT  & \emph{Traditional SRL: wrong active / passive swap} Q1: Did Michelle like the car? Q2: Was Michelle liked by the car? (not duplicates)  \\
\arrayrulecolor{black!20}\cmidrule{3-4}
45 &         & MFT  & \emph{Traditional SRL: active / passive swap with people}: Q1: Does Mary remember Adam? Q2: Is Adam remembered by Mary? (duplicates) \\
\arrayrulecolor{black!20}\cmidrule{3-4}
46 &         & MFT  & \emph{Traditional SRL: wrong active / passive swap with people}: Q1: Does Michelle trust Angela? Q2: Is Michelle trusted by Angela? (not duplicates)  \\
\arrayrulecolor{black}\midrule
47 & \multirow{15}{*}{\rotatebox[origin=c]{90}{Logic}}        & MFT  & \emph{A or B is not the same as C and D}: Q1: Is Emily Fisher an actress or an investor? Q2: Is Emily Fisher simultaneously an auditor and an organizer? (not duplicates)  \\
\arrayrulecolor{black!20}\cmidrule{3-4}
48 &       & MFT  & \emph{A or B is not the same as A and B}: Q1: Is Taylor King an educator or an accountant? Q2: Is Taylor King simultaneously an educator and an accountant? (not duplicates)  \\
\arrayrulecolor{black!20}\cmidrule{3-4}
49 &       & MFT  & \emph{A and / or B is the same as B and / or A}: Q1: Is Jennifer Flores an engineer and an editor? Q2: Is Jennifer Flores an editor and an engineer? (duplicates) \\
\arrayrulecolor{black!20}\cmidrule{3-4}
50 &       & MFT  & \emph{a \texttt{\{nationality\}} \texttt{\{profession\}} = a \texttt{\{profession\}} and \texttt{\{nationality\}}}: Q1: Is Christina Nguyen a French nurse? Q2: Is Christina Nguyen a nurse and French? (duplicates) \\
\arrayrulecolor{black!20}\cmidrule{3-4}
51 &       & MFT  & \emph{Reflexivity: \texttt{(q,q)} should be duplicate}: Q1: What does the following symbol mean  Q2: What does the following symbol mean ? (duplicates)  \\
\arrayrulecolor{black!20}\cmidrule{3-4}
52 &       & INV  & \emph{Symmetry: \texttt{f(a,b) = f(b,a)}}: Q1: Which colleges come under the GMAT? Q2: Which all colleges come under GMAT in india? (not duplicates) || Q1: Which all colleges come under GMAT in india? Q2: Which colleges come under the GMAT? (not duplicates)  \\
\arrayrulecolor{black!20}\cmidrule{3-4}
53 &       & DIR  & \emph{Testing implications}: Q1: Why was Albert Einstein considered an atheist? Q2: Do atheists look down on Albert Einstein because he was religious? (not duplicates) || Q1: Why was Albert Einstein considered an atheist? Q2: Was Albert Einstein an atheist? (not duplicates) || ...\\
\arrayrulecolor{black}\bottomrule
\end{tabular}%
}
\caption{Checklists tests 28--53 for QQP.}
\label{tab:checklist_tests_33_to_53}
\end{table*}

\begin{table*}[htb]
\centering
\resizebox{\textwidth}{!}{%
\begin{tabular}{lrrrrr}
\toprule
& \makecell[c]{\textbf{RoBERTa-L}} & \makecell[c]{\textbf{DeBERTa-L}} & \makecell[c]{\textbf{OPT-6.7B}} & \makecell[c]{\textbf{GPT-2-XL}} & \makecell[c]{\textbf{T5-11B}}  \\
 \midrule
\textbf{\% Total Tests with 95+\% accuracy} &  45.3 & 45.3 & 34.0 & 26.4 & 45.3 \\
\quad\textbf{\rotatebox[origin=c]{180}{$\Lsh$} \& No 3+\% drops from scaling} & 45.3 & 45.3 & 24.5 & 26.4 & 32.1  \\
\quad\textbf{\rotatebox[origin=c]{180}{$\Lsh$} \& Equally robust across model sizes} & 32.1 & 34.0 & 13.2 & 22.6 & 26.4 \\
\textbf{\% 10+\% gains from scaling, no 3+\% drops} & 28.3 & 20.8 & 9.4 & 7.5 & 15.1  \\
\arrayrulecolor{black!20}\midrule
\textbf{\% Total Tests with \textless{}60\% accuracy} & 28.3 & 22.6 & 32.1 & 35.8 & 15.1  \\
\textbf{\% Major scaling complications (10+\% drops)} & 1.9 & 1.9 & 37.7 & 18.9 & 22.6 \\
\arrayrulecolor{black}\bottomrule
\end{tabular}%
}
\caption{Analysis of CheckList results for identifying duplicate questions (QQP) with best models from each type. %
}
\label{tab:checklist_summary}
\end{table*}

\begin{figure*}[t]
  \centering
  \begin{subfigure}{\textwidth}
    \centering
    \includegraphics[width=\textwidth]{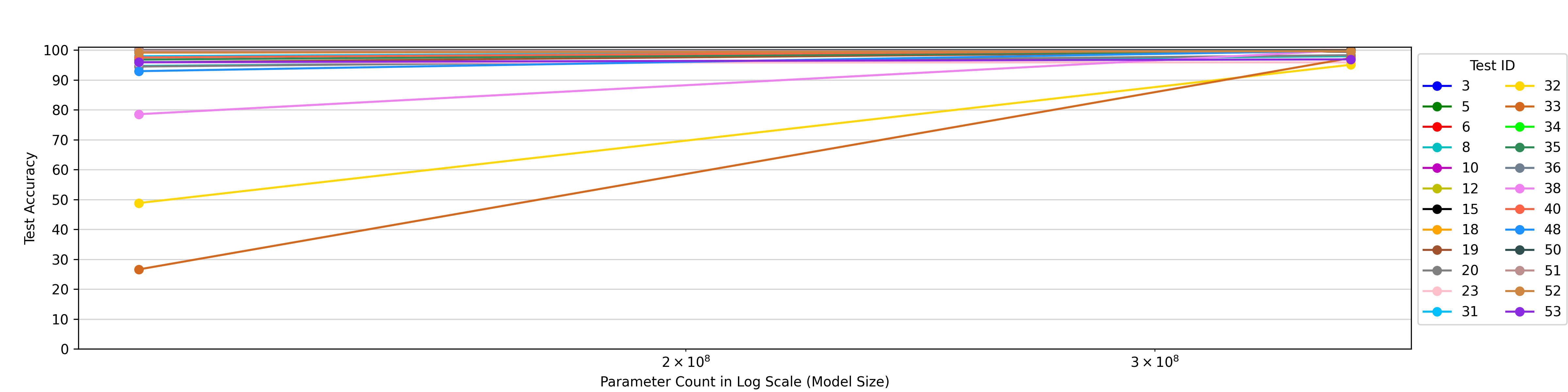}
    \caption{Tests for which the large model gets an accuracy of 95\% or more.}
    \label{fig:95_plus_roberta}
  \end{subfigure}
  \begin{subfigure}{\textwidth}
    \centering
    \includegraphics[width=\textwidth]{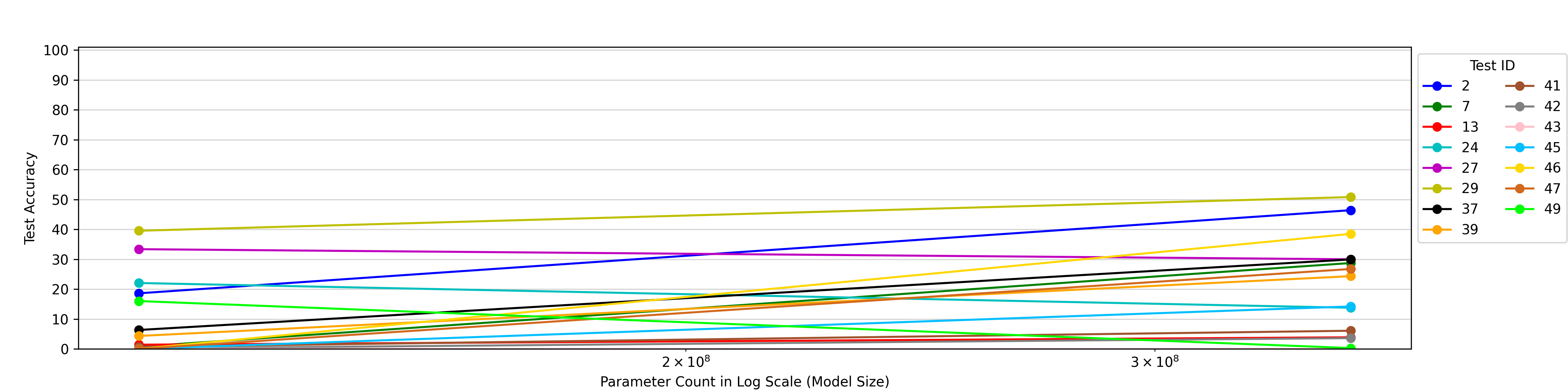}
    \caption{Tests for which the large model gets an accuracy of 60\% or less.}
    \label{fig:less_than_60_roberta}
  \end{subfigure}
  \begin{subfigure}{\textwidth}
    \centering
    \includegraphics[width=\textwidth]{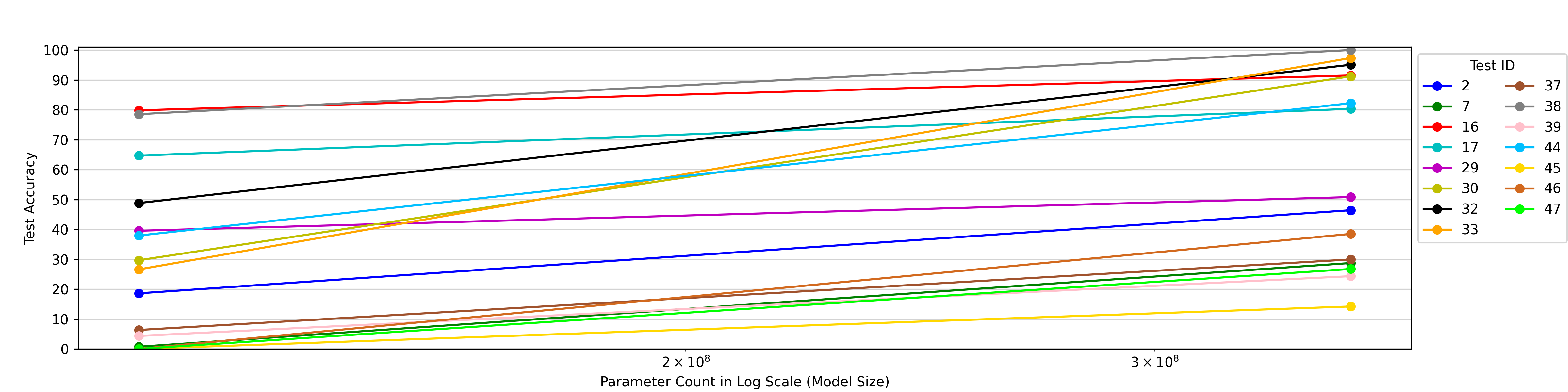}
    \caption{Test with 10+\% gains obtained by scaling to the largest model version without 3+\% drops.}
    \label{fig:scaling_impr_roberta}
  \end{subfigure}
  \begin{subfigure}{\textwidth}
    \centering
    \includegraphics[width=\textwidth]{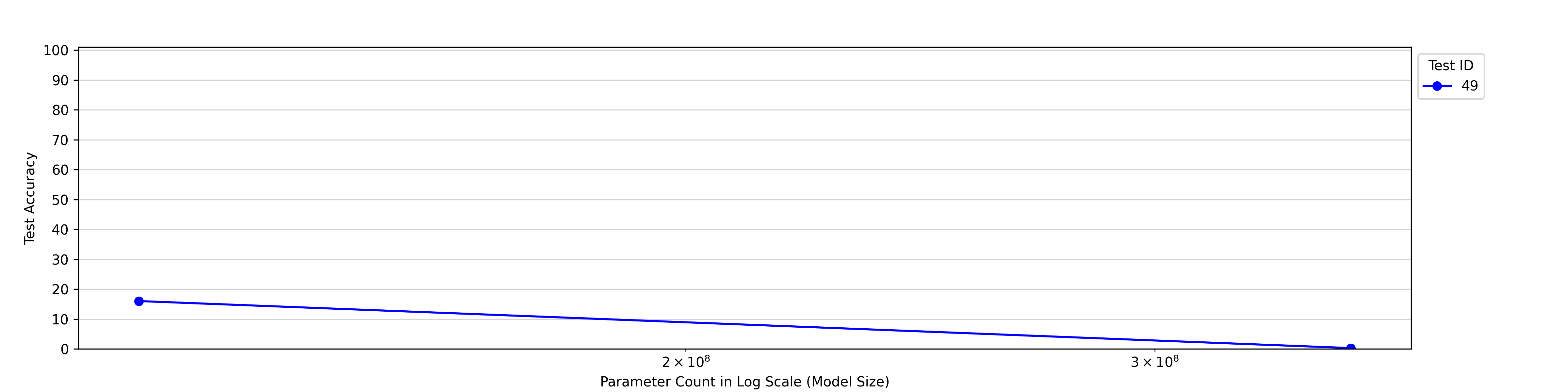}
    \caption{Tests with scaling complications (10+\% drops).}
    \label{fig:scaling_complications_roberta}
  \end{subfigure}
  \caption{RoBERTa}
  \label{fig:checklist_figures_roberta}
\end{figure*}

\begin{figure*}[t]
  \centering
  \begin{subfigure}{\textwidth}
    \centering
    \includegraphics[width=\textwidth]{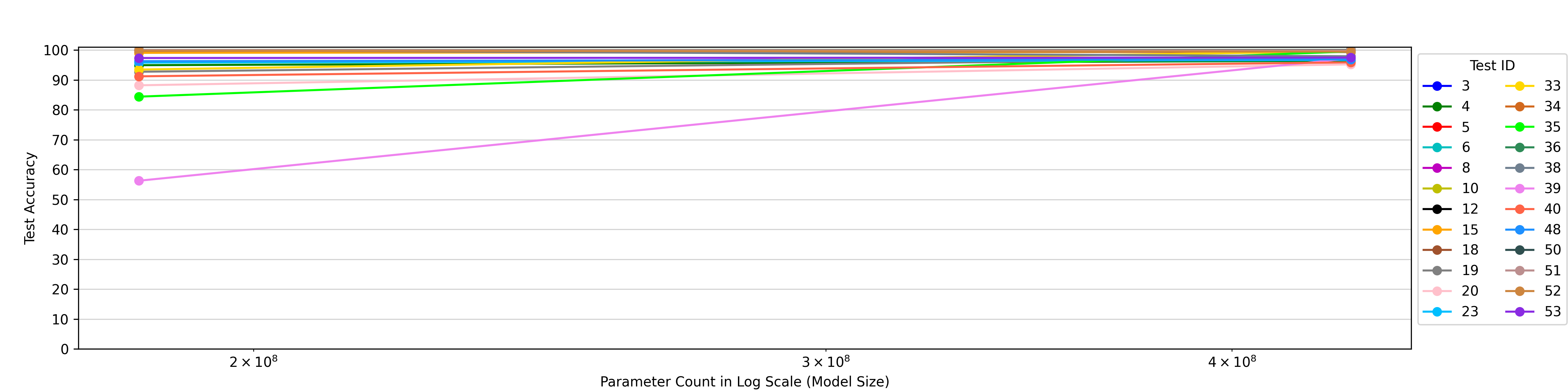}
    \caption{Tests for which the large model gets an accuracy of 95\% or more.}
    \label{fig:95_plus_deberta}
  \end{subfigure}
  \begin{subfigure}{\textwidth}
    \centering
    \includegraphics[width=\textwidth]{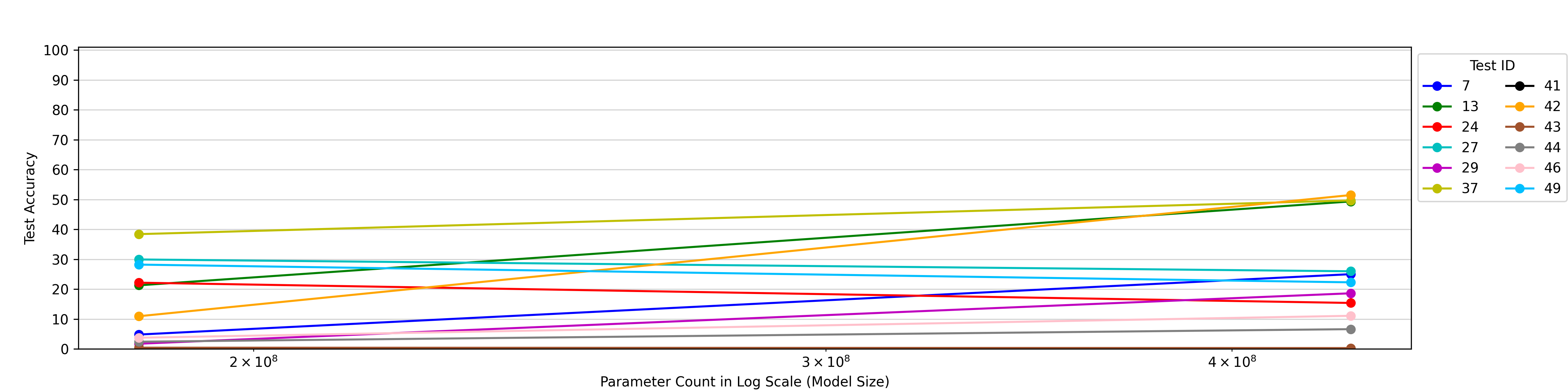}
    \caption{Tests for which the large model gets an accuracy of 60\% or less.}
    \label{fig:less_than_60_deberta}
  \end{subfigure}
  \begin{subfigure}{\textwidth}
    \centering
    \includegraphics[width=\textwidth]{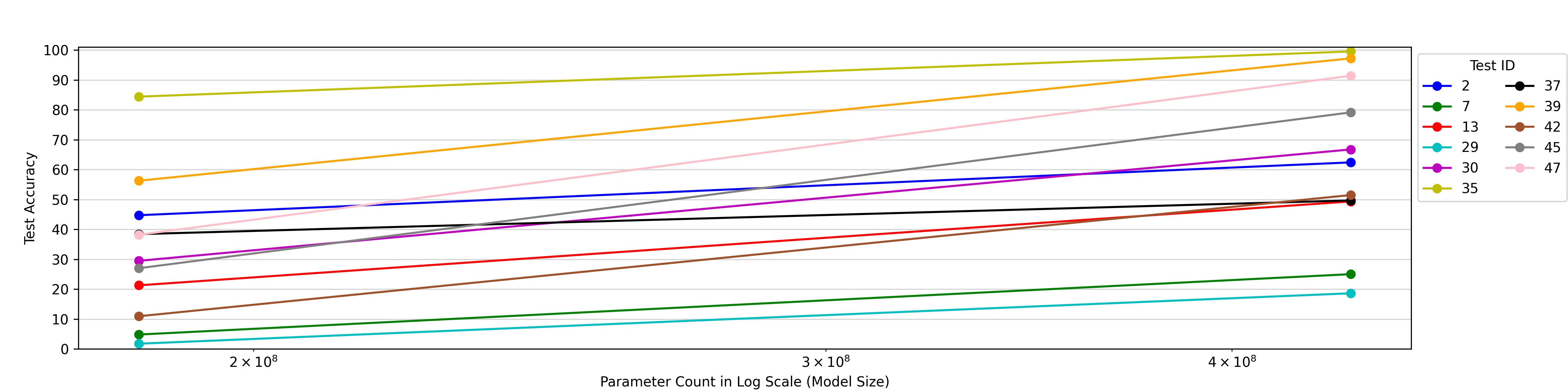}
    \caption{Test with 10+\% gains obtained by scaling to the largest model version without 3+\% drops.}
    \label{fig:scaling_impr_deberta}
  \end{subfigure}
  \begin{subfigure}{\textwidth}
    \centering
    \includegraphics[width=\textwidth]{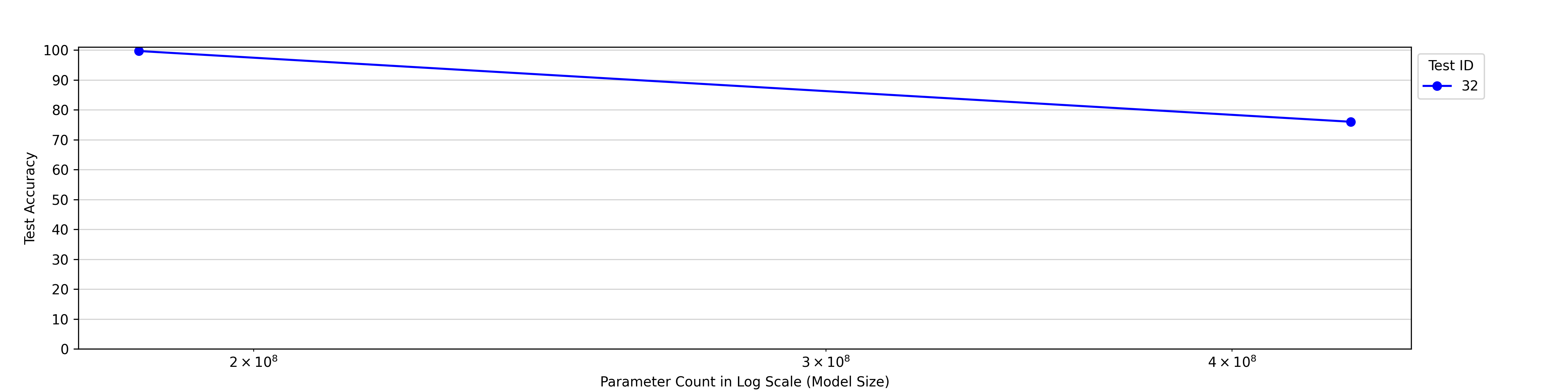}
    \caption{Tests with scaling complications (10+\% drops).}
    \label{fig:scaling_complications_deberta}
  \end{subfigure}
  \caption{DeBERTa}
  \label{fig:checklist_figures_deberta}
\end{figure*}

\begin{figure*}[t]
  \centering
  \begin{subfigure}{\textwidth}
    \centering
    \includegraphics[width=\textwidth]{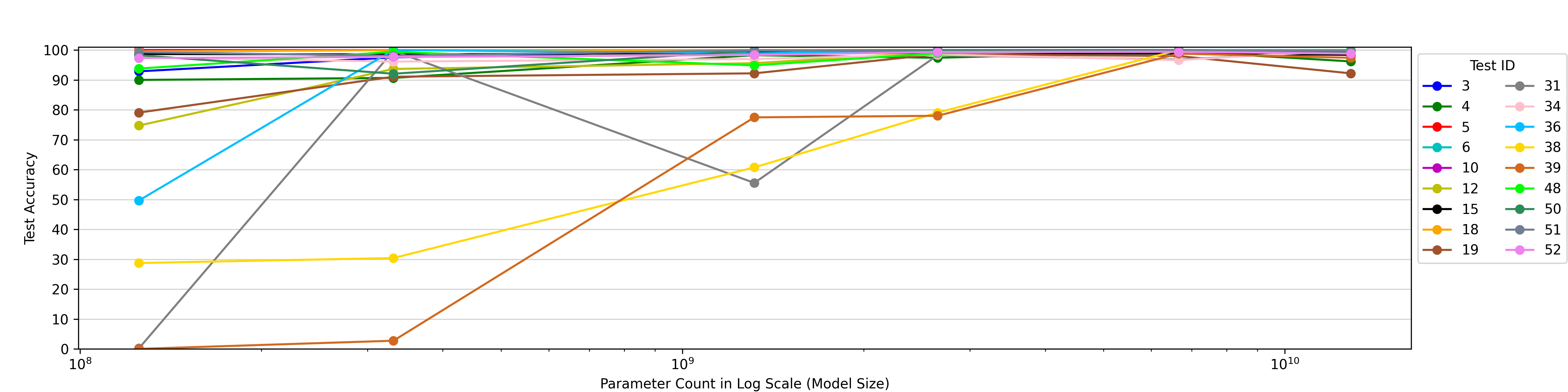}
    \caption{Tests for which the 6.7B model gets an accuracy of 95\% or more.}
    \label{fig:95_plus_opt}
  \end{subfigure}
  \begin{subfigure}{\textwidth}
    \centering
    \includegraphics[width=\textwidth]{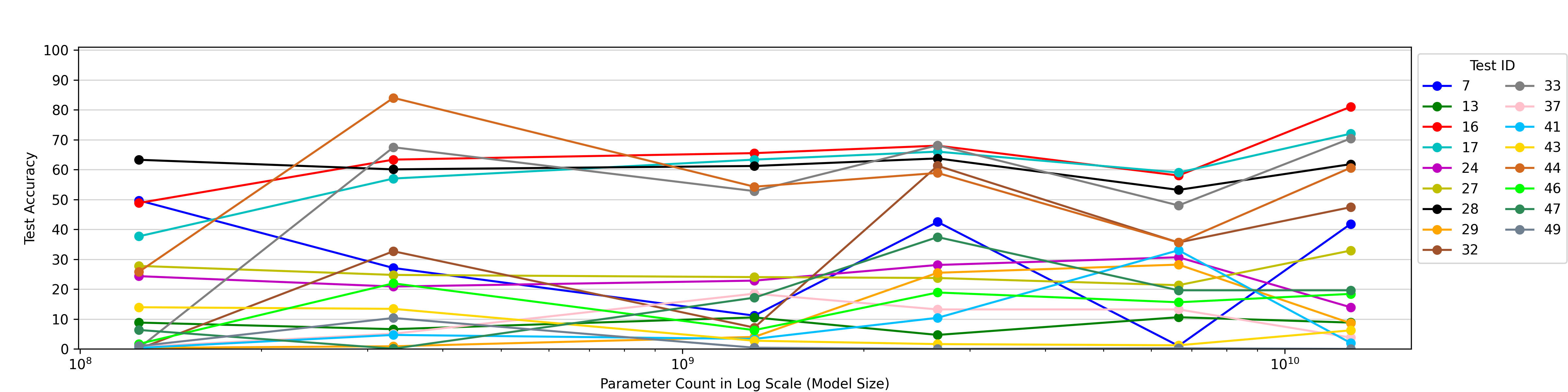}
    \caption{Tests for which the 6.7B model gets an accuracy of 60\% or less.}
    \label{fig:less_than_60_opt}
  \end{subfigure}
  \begin{subfigure}{\textwidth}
    \centering
    \includegraphics[width=\textwidth]{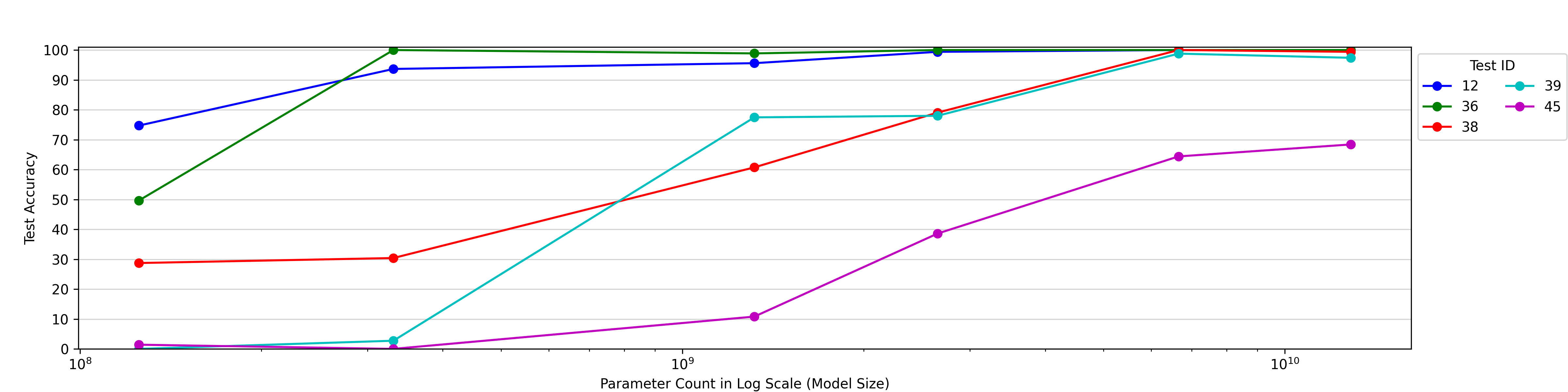}
    \caption{Test with 10+\% gains obtained by scaling to the largest model version without 3+\% drops.}
    \label{fig:scaling_impr_opt}
  \end{subfigure}
  \begin{subfigure}{\textwidth}
    \centering
    \includegraphics[width=\textwidth]{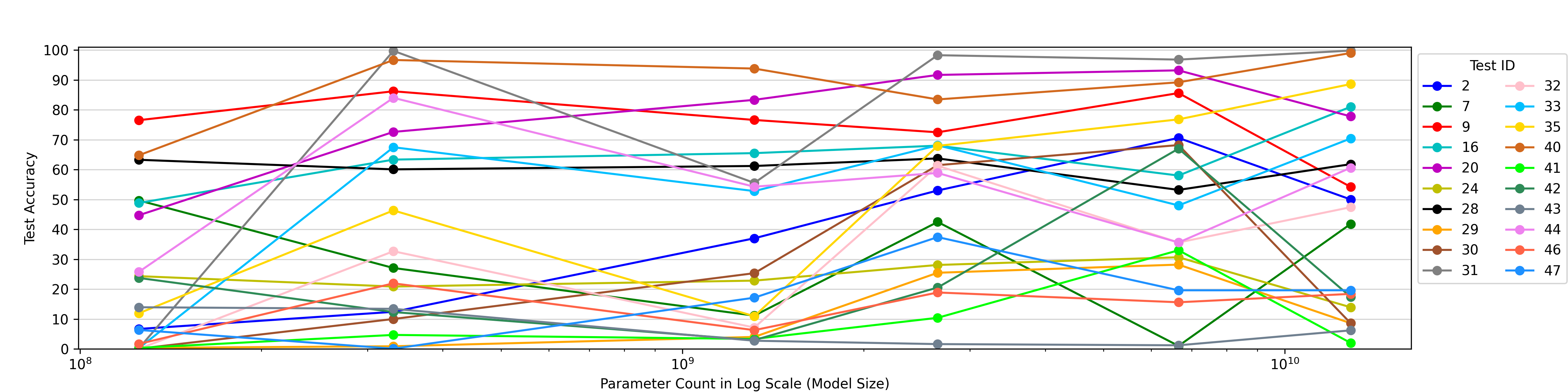}
    \caption{Tests with scaling complications (10+\% drops).}
    \label{fig:scaling_complications_opt}
  \end{subfigure}
  \caption{OPT}
  \label{fig:checklist_figures_opt}
\end{figure*}

\begin{figure*}[t]
  \centering
  \begin{subfigure}{\textwidth}
    \centering
    \includegraphics[width=\textwidth]{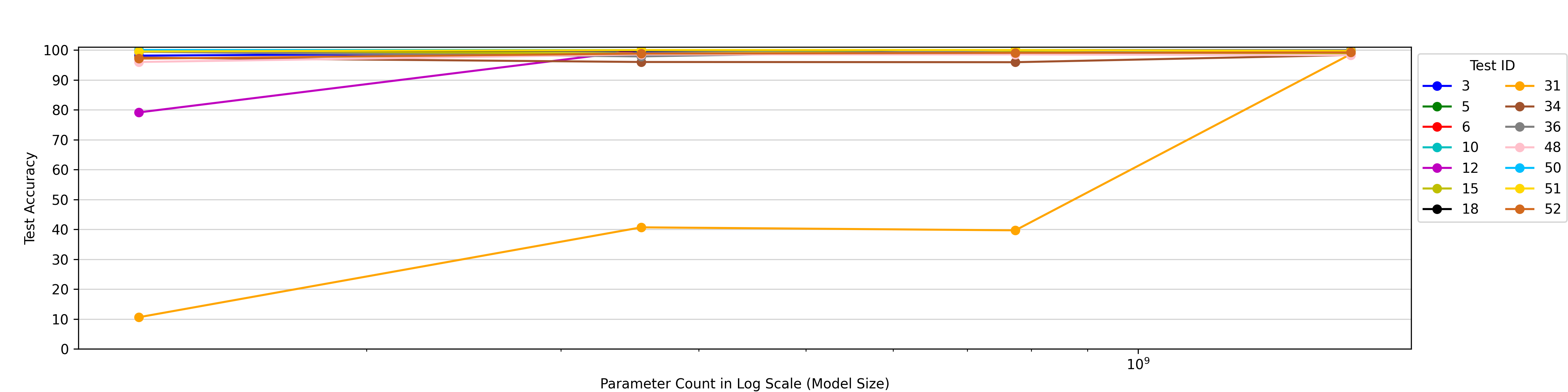}
    \caption{Tests for which the XL model gets an accuracy of 95\% or more.}
    \label{fig:95_plus_gpt}
  \end{subfigure}
  \begin{subfigure}{\textwidth}
    \centering
    \includegraphics[width=\textwidth]{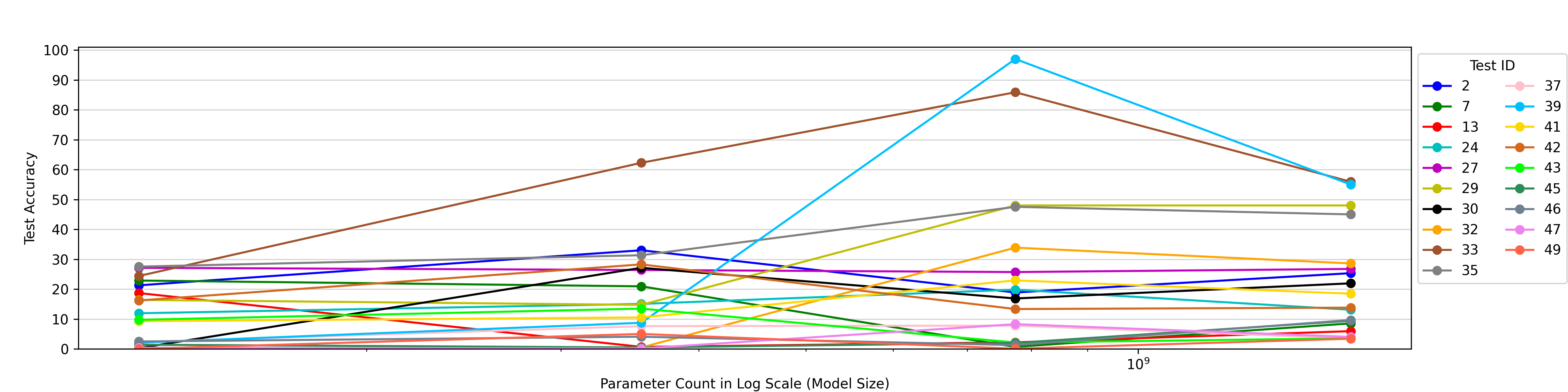}
    \caption{Tests for which the XL model gets an accuracy of 60\% or less.}
    \label{fig:less_than_60_gpt}
  \end{subfigure}
  \begin{subfigure}{\textwidth}
    \centering
    \includegraphics[width=\textwidth]{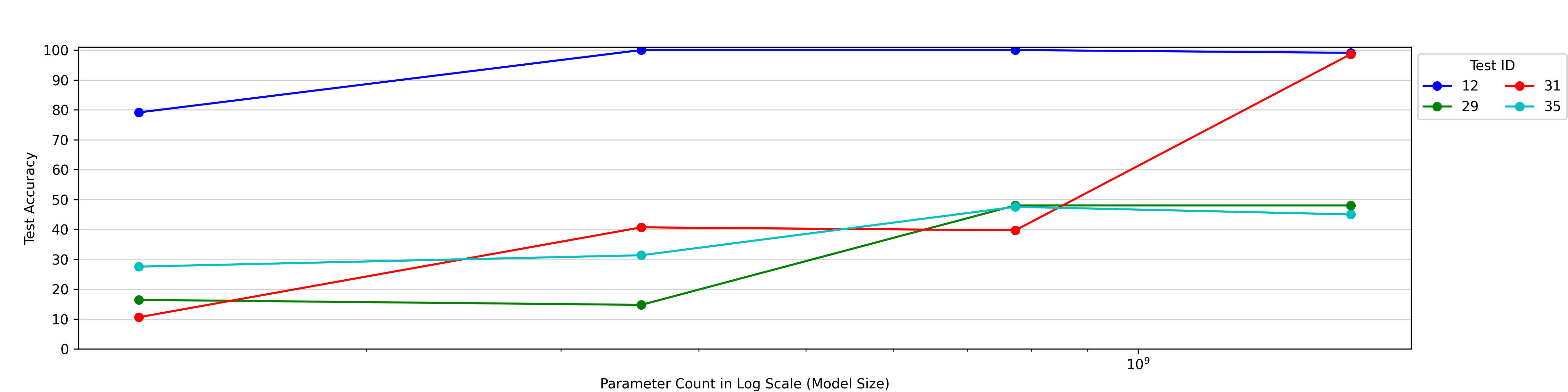}
    \caption{Test with 10+\% gains obtained by scaling to the largest model version without 3+\% drops.}
    \label{fig:scaling_impr_gpt}
  \end{subfigure}
  \begin{subfigure}{\textwidth}
    \centering
    \includegraphics[width=\textwidth]{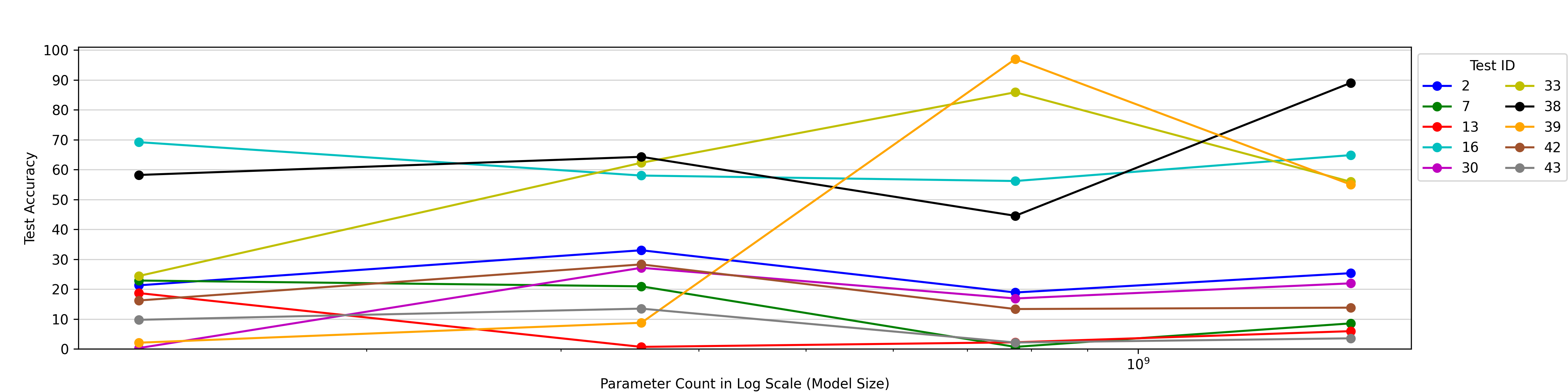}
    \caption{Tests with scaling complications (10+\% drops).}
    \label{fig:scaling_complications_gpt}
  \end{subfigure}
  \caption{GPT-2}
  \label{fig:checklist_figures_gpt}
\end{figure*}

\begin{figure*}[t]
  \centering
  \begin{subfigure}{\textwidth}
    \centering
    \includegraphics[width=\textwidth]{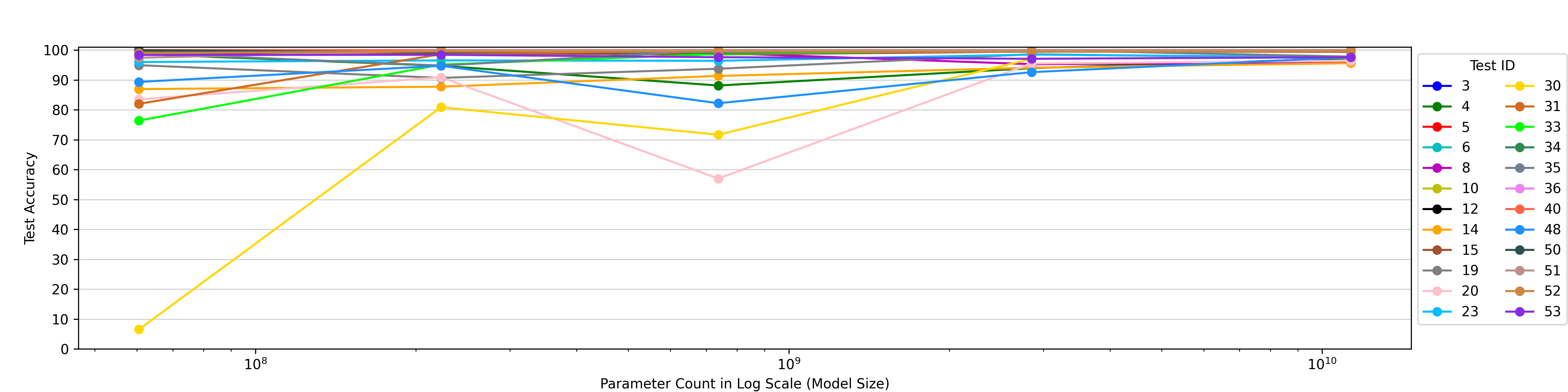}
    \caption{Tests for which the 11B model gets an accuracy of 95\% or more.}
    \label{fig:95_plus_t5}
  \end{subfigure}
  \begin{subfigure}{\textwidth}
    \centering
    \includegraphics[width=\textwidth]{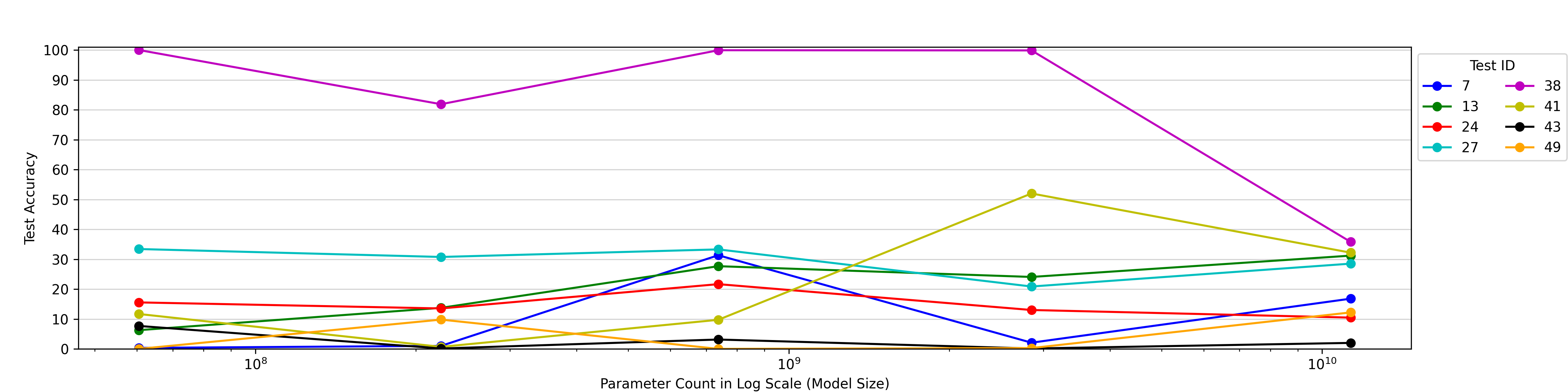}
    \caption{Tests for which the 11B model gets an accuracy of 60\% or less.}
    \label{fig:less_than_60_t5}
  \end{subfigure}
  \begin{subfigure}{\textwidth}
    \centering
    \includegraphics[width=\textwidth]{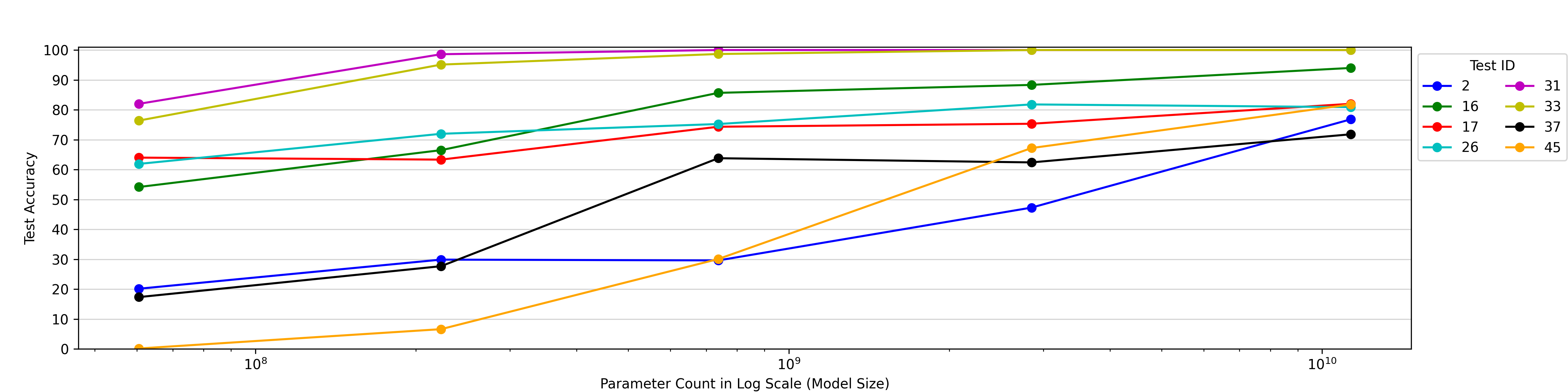}
    \caption{Test with 10+\% gains obtained by scaling to the largest model version without 3+\% drops.}
    \label{fig:scaling_impr_t5}
  \end{subfigure}
  \begin{subfigure}{\textwidth}
    \centering
    \includegraphics[width=\textwidth]{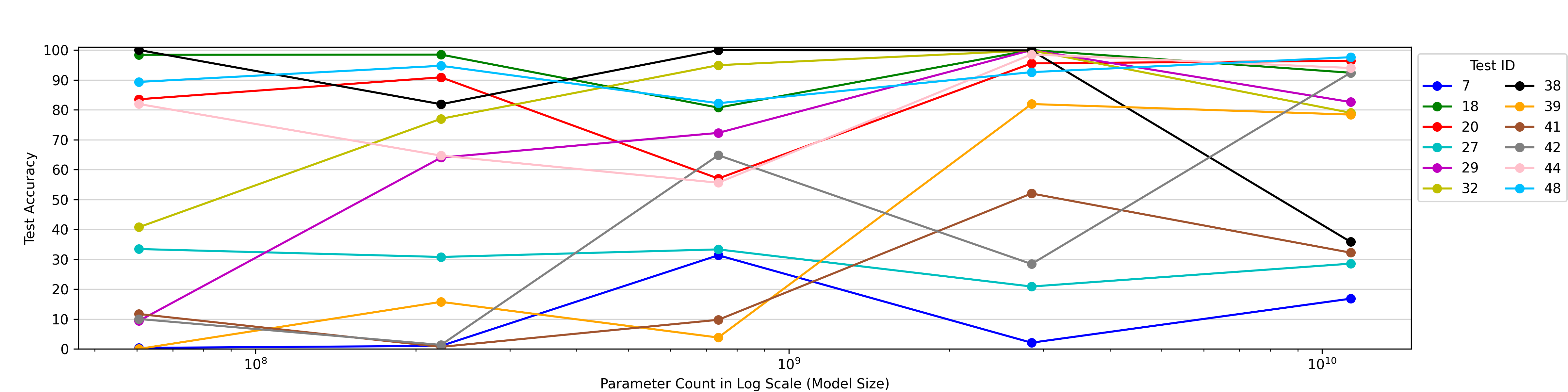}
    \caption{Tests with scaling complications (10+\% drops).}
    \label{fig:scaling_complications_t5}
  \end{subfigure}
  \caption{T5}
  \label{fig:checklist_figures_t5}
\end{figure*}

\begin{table*}[]
\footnotesize
\begin{tabular}{@{}p{.12\linewidth}p{.9\linewidth}@{}}
\toprule
\textbf{Task} &
  \textbf{CoT Prompt} \\
\toprule
CondaQA &
  \begin{tabular}[c]{@{}p{1\linewidth}@{}}\textbf{Passage:} \\ The end of the long-held animosity between Moscow and Beijing was marked by the visit to China by Soviet General Secretary Mikhail Gorbachev in 1989. After the 1991 demise of the Soviet Union, China's relations with Russia and the former states of the Soviet Union became more amicable as the conflicting ideologies of the two vast nations no longer stood in the way. A new round of bilateral agreements was signed during reciprocal head of state visits. As in the early 1950s with the Soviet Union, Russia has again become an important source of military technology for China, as well as for raw materials and trade. Friendly relations with Russia have been an important advantage for China, complementing its strong ties with the U.S.\\ \textbf{Question:} Can China rely on both the US and Russia as supportive allies?\\ \textbf{Give the rationale before answering.} If a country has either friendly relations or strong ties with another country, one can expect the other country is their ally. So the answer is \emph{YES.}\\ \#\#\#\\ ... \\ \#\#\#\\ \textbf{Passage:} Trams have operated continuously in Melbourne since 1885 (the horse tram line in Fairfield opened in 1884, but was at best an irregular service). Since then they have become a distinctive part of Melbourne's character and feature in tourism and travel advertising. Melbourne's cable tram system opened in 1885, and expanded to one of the largest in the world, with of double track. The first electric tram line opened in 1889, but closed only a few years later in 1896. In 1906 electric tram systems were opened in St Kilda and Essendon, marking the start of continuous operation of Melbourne's electric trams.\\ \textbf{Question:} If I wanted to take a horse tram in 1884, could I look up the next tram on a schedule?\\ \textbf{Give the rationale before answering.}\end{tabular} \\ \midrule
BoolQ &
  \begin{tabular}[c]{@{}p{1\linewidth}@{}}\textbf{Passage:} Love Island Australia is an Australian dating reality show based on the British series Love Island. The series is presented by Sophie Monk and narrated by Eoghan McDermott. The series began airing on 9Go! and 9Now on 27 May 2018. The final aired on 5 July 2018, with Grant Crapp and Tayla Damir winning and sharing the \$50,000 prize money. Eden Dally and Erin Barnett finished as runners up.\\ \textbf{Question:} is there a prize for love island australia\\ \textbf{Give the rationale before answering.} There is a \$50000 prize money. So the answer is \emph{True.}\\ \#\#\# \\ \dots \\ \#\#\#\\ \textbf{Passage:} Conscription in the United States, commonly known as the draft, has been employed by the federal government of the United States in five conflicts: the American Revolution, the American Civil War, World War I, World War II, and the Cold War (including both the Korean War and the Vietnam War). The third incarnation of the draft came into being in 1940 through the Selective Training and Service Act. It was the country's first peacetime draft. From 1940 until 1973, during both peacetime and periods of conflict, men were drafted to fill vacancies in the United States Armed Forces that could not be filled through voluntary means. The draft came to an end when the United States Armed Forces moved to an all-volunteer military force. However, the Selective Service System remains in place as a contingency plan; all male civilians between the ages of 18 and 25 are required to register so that a draft can be readily resumed if needed. United States Federal Law also provides for the compulsory conscription of men between the ages of 17 and 45 and certain women for militia service pursuant to Article I, Section 8 of the United States Constitution and 10 U.S. Code § 246.\\ \textbf{Question:} was there a draft in the revolutionary war\\ \textbf{Give the rationale before answering.}\end{tabular} \\ \midrule
PERSPECTRUM &
  \begin{tabular}[c]{@{}p{1\linewidth}@{}}\textbf{Perspective:} Interfering with a species for cosmetic reasons is poor practice\\ \textbf{Claim:} The breeding of white tigers in captivity should be banned. \\ Does the perspective support or undermine the claim? \\ Answer with: supports, undermines or not a valid perspective.\\ \textbf{Give the rationale before answering.} Breeding tigers in captivity is interfering with the species. So the answer is \emph{supports.}\\ \#\#\# \\ \dots \\ \#\#\#\\ \textbf{Perspective:} I would just be so upset that that guy cheated and stole that moment away from me.\\ \textbf{Claim:} should cheaters be given a second chance. \\ Does the perspective support or undermine the claim? \\ Answer with: supports, undermines or not a valid perspective.\\ \textbf{Give the rationale before answering.}\end{tabular} \\ 
  \bottomrule
  \end{tabular}
\caption{The exact prompts \textbf{with CoT} used for each contrast set (continued in Tables \ref{tab:cot-contrast2}-\ref{tab:cot-contrast4}).}
\label{tab:cot-contrast1}
\end{table*}

\begin{table*}
\footnotesize
\begin{tabular}{@{}p{.1\linewidth}p{.9\linewidth}@{}}
\toprule
\textbf{Task} &
  \textbf{CoT Prompt} \\
\toprule
ROPES &
  \begin{tabular}[c]{@{}p{1\linewidth}@{}}\textbf{Passage:} Hemophilia is the name of a group of hereditary diseases that affect the body's ability to control blood clotting. Hemophilia is caused by a lack of clotting factors in the blood. Clotting factors are normally released by platelets. Since people with hemophilia cannot produce clots, any cut can put a person at risk of bleeding to death. The risk of internal bleeding is also increased in hemophilia, especially into muscles and joints. This disease affected the royal families of Europe. Greg and Bobby are coworkers who are currently eating lunch together on their break. They haven't known each other very long and are having a casual conversation when Bobby mentions that he is a hemophiliac. Greg, unsure of what exactly that means, looks confused and taken aback. Bobby, noticing the bewilderment on Greg's face, takes a moment to explain the details of his condition to Greg.\textbf{\textbf{Question:}} Does Greg or Bobby have a lower platelet count?\\ \textbf{Give the rationale before answering.} Bobby is hemophiliac which would mean that he lacks clotting factors normally released by platelets. So the answer is \emph{Bobby.}\\ \#\#\# \\ \dots \\ \#\#\#\\ \textbf{Passage:} The theory of evolution by natural selection was proposed at about the same time by both Charles Darwin and Alfred Russel Wallace, shown in Figure below , and was set out in detail in Darwin's 1859 book On the Origin of Species . Natural selection is a process that causes heritable traits that are helpful for survival and reproduction to become more common, and harmful traits, or traits that are not helpful or advantageous for survival to become more rare in a population of organisms. This occurs because organisms with advantageous traits are more "fit" to survive in a particular environment and have "adapted" to the conditions of that environment. These individuals will have greater reproductive success than organisms less fit for survival in the environment. This will lead to an increase in the number of organisms with the advantageous trait(s) over time. Over many generations, adaptations occur through a combination of successive, small, random changes in traits, and natural selection of those variants best-suited for their environment. Natural selection is one of the cornerstones of modern biology.\\ \textbf{\textbf{Question:}} The tree frogs range covers 2 different habitats, One has a lot of trees and the other a lot of tall green grass with few if any trees. Tree frogs can either be green or brown. Which color tree frog has an advantage living in trees?\\ \textbf{Give the rationale before answering.}\end{tabular} \\ \midrule
DROP &
  \begin{tabular}[c]{@{}p{1\linewidth}@{}}\textbf{Passage:} Švitrigaila was losing his influence in the Slavic principalities and could no longer resist Poland and Sigismund. On 4 September 1437 he attempted to reconcile with Poland: he would rule the lands that still backed him , and after his death these territories would pass to the King of Poland. However, under strong protest from Sigismund, the Polish Senate declined to ratify the treaty . In 1438 Švitrigaila withdrew to Moldavia. The reign of Sigismund Kęstutaitis was brief — he was assassinated in 1440. Švitrigaila returned from exile in 1442 and ruled Lutsk until his death a decade later. Jogaila's son Casimir IV Jagiellon, born in 1426, received approval as a hereditary hospodar from Lithuania's ruling families in 1440. This event is seen by the historians Jerzy Lukowski and Hubert Zawadzki as marking the end of the succession dispute.\\ \textbf{Question:} Who was ruler first, Sigismund of Casimir IV Jagiellon?\\ \textbf{Give the rationale before answering.} Casimir IV Jagiellon became the ruler only after the assination of Sigismund Kęstutaitis. So the answer is \emph{Sigismund Kęstutaitis.}\\ \#\#\# \\ \dots \\ \#\#\#\\ \textbf{Passage:} The origins of al-Qaeda can be traced back to the Soviet war in Afghanistan . The United States, the United Kingdom, Saudi Arabia, Pakistan, and the People's Republic of China supported the Islamist Afghan mujahadeen guerillas against the military forces of the Soviet Union and the Democratic Republic of Afghanistan. A small number of "Afghan Arab" volunteers joined the fight against the Soviets, including Osama bin Laden, but there is no evidence they received any external assistance. In May 1996 the group World Islamic Front for Jihad Against Jews and Crusaders , sponsored by bin Laden , started forming a large base of operations in Afghanistan, where the Islamist extremist regime of the Taliban had seized power earlier in the year. In February 1998, Osama bin Laden signed a fatwā, as head of al-Qaeda, declaring war on the West and Israel. Earlier in August 1996, Bin Laden declared jihad against the United States. Later in May 1998 al-Qaeda released a video declaring war on the U.S. and the West. On 7 August 1998, al-Qaeda struck the U.S. embassies in Kenya and Tanzania, killing 224 people, including 12 Americans. In retaliation, U.S. President Bill Clinton launched Operation Infinite Reach, a bombing campaign in Sudan and Afghanistan against targets the U.S. asserted were associated with WIFJAJC, although others have questioned whether a pharmaceutical plant in Sudan was used as a chemical warfare facility. The plant produced much of the region's antimalarial drugs and around 50\% of Sudan's pharmaceutical needs. The strikes failed to kill any leaders of WIFJAJC or the Taliban. Next came the 2000 millennium attack plots, which included an attempted bombing of Los Angeles International Airport. On 12 October 2000, the USS Cole bombing occurred near the port of Yemen, and 17 U.S. Navy sailors were killed.\\ \textbf{Question:} How many years was it between when Bin Laden declared a war against the United States and when he became leader of al-Qaeda?\\ \textbf{Give the rationale before answering.}\end{tabular} \\ \midrule
\end{tabular}
\caption{The exact prompts \textbf{with CoT} used for each contrast set (continuation of Table \ref{tab:cot-contrast1}).}
\label{tab:cot-contrast2}
\end{table*}

\begin{table*}
\footnotesize
\begin{tabular}{@{}p{.1\linewidth}p{.9\linewidth}@{}}
\toprule
\textbf{Task} &
  \textbf{CoT Prompt} \\
\toprule  

IMDb &
  \begin{tabular}[c]{@{}p{1\linewidth}@{}}\textbf{Review:} "Garde à Vue has to be seen a number of times in order to understand the sub-plots it contains. If you're not used to french wordy films, based upon conversation and battle of wits rather than on action, don't even try to watch it. You'll only obtain boredom to death, and reassured opinion that french movies are not for you.\textless{}br /\textgreater{}\textless{}br /\textgreater{}Garde à Vue is a wordy film, essentially based upon dialogs (written by Audiard by the way)and it cruelly cuts the veil of appearances.\textless{}br /\textgreater{}\textless{}br /\textgreater{}Why does Maître Martineau (Serrault) prefer to be unduly accused of being a child murderer rather than telling the truth ? Because at the time of the murder he was with a 18 years old girl with which he has a 8-years sexual relation. His wife knows it, she's jealous of it and he prefers to be executed (in 1980 in France, there was still death penalty)rather than unveiling the sole ""pure and innocent"" aspect of his pitiful life."\\ \textbf{What is the sentiment of this review:} Positive or Negative?\\ \textbf{Give the rationale before answering.} The reviewer describes the movie to be complex, one where the audience needs to pay close attention to the dialogues and not just visuals to understand the nuanced story. So the answer is \emph{Positive.}\\ \#\#\# \\ \dots \\ \#\#\#\\ \textbf{Review:} May 2004, Wonderland is fairly new in the UK. Brilliant film of a brutal true story. If you know LA from the early 80's, you will appreciate how well it is captured. The use of the elements which make up its gritty cinematic style is original, amplifying the experience and bringing the viewer very close to actually being there. The use of a disjointed 'Pulp Fiction' style time line allows exploration of the uncertainty concerning what really happened, while the direction and performances of the cast command attention, especially Val Kilmer as John Holmes; an Oscar for sure if I were handing them out.\\ \textbf{What is the sentiment of this review:} Positive or Negative?\\ \textbf{Give the rationale before answering.}\end{tabular} \\ \midrule

MATRES &
  \begin{tabular}[c]{@{}p{1\linewidth}@{}}\textbf{Passage:} Dr. Barnett Slepian was  *killed*   in his kitchen by a sniper's bullet last fall . Investigators said Friday  they  *found*   a rifle buried  near his home in the Buffalo suburb of Amherst.\\ \textbf{Question:} When did the event *killed* happen in relation to the event *found*: before, after, simultaneously, or is it vague?\\ \textbf{Give the rationale before answering.} The investigation into the killing, during which the finding took place, was started only after the killing. So the answer is \emph{before.}\\ \#\#\# \\ \dots    \\ \#\#\#\\ \textbf{Passage:} A book of memoirs and photographs from the climb by Mr Lowe , which he worked on with Dr Lewis-Jones , is due to be published in May . He  *said* : " Lowe was a brilliant , kind fellow who never sought the limelight ... and 60 years on from Everest his achievements deserve wider recognition . " He was  *involved* in two of the most important explorations of the 20th Century ... yet remained a humble , happy man right to the end ... an inspirational lesson to us all . "\\ \textbf{Question:} When did the event *said* happen in relation to the event *involved*: before, after, simultaneously, or is it vague?\\ \textbf{Give the rationale before answering.}\end{tabular} \\ \bottomrule
\end{tabular}
\caption{The exact prompts \textbf{with CoT} used for each contrast set (continuation of Table \ref{tab:cot-contrast2}).}
\label{tab:cot-contrast3}
\end{table*}

\begin{table*}
\footnotesize
\begin{tabular}{@{}p{.1\linewidth}p{.9\linewidth}@{}}
\toprule
\textbf{Task} &
  \textbf{CoT Prompt} \\
\toprule  
  MCTACO &
  \begin{tabular}[c]{@{}p{1\linewidth}@{}}\textbf{Passage:} He succeeds James A. Taylor, who stepped down as chairman, president and chief executive in March for health reasons. How often are chief executives typically replaced?\\ Is this the answer: every 10 weeks?\\ \textbf{Give the rationale before answering.} Most chief executives keep their positions for years. So the answer is \emph{no.}\\ \#\#\# \\ \dots \\ \#\#\#\\ \textbf{Passage:} Roberta Adams skipped the thick how-to guide on child-custody forms and sat down at a computer at the Lamoreaux Justice Center in Orange on Wednesday. When did Roberta arrive at the Lamoreaux Justice Center?\\ Is this the answer: 10:45 PM?\\ \textbf{Give the rationale before answering.}\end{tabular} \\ \midrule
Quoref &
  \begin{tabular}[c]{@{}p{1\linewidth}@{}}\textbf{Passage:} In the year 1978, Gracie Bowen, a 15-year-old tomboy who lives in South Orange, New Jersey, is crazy about soccer, as are her three brothers and their former soccer star father. Although Gracie wants to join her brothers and neighbor Kyle in the nightly practices her father runs, she is discouraged by everyone except her older brother, Johnny. Johnny, Gracie and Kyle attend Columbia High School, where Johnny is the captain and star player for the varsity soccer team. After missing a shot at the end of a game, the despondent Johnny drives off with a friend's car and dies in a traffic accident. Struggling with grief, Gracie decides that she wants to replace her brother on the team. Her father does not believe that girls should play soccer, telling her she is neither tough nor talented enough. Her mother is a nurse who lacks the competitive drive of the rest of her family and fears for Gracie's safety. Her mother later tells Gracie that she would have liked to become a surgeon, but that option had not been available to her as a woman. Rejected and depressed, Gracie begins to rebel; she stops doing her schoolwork, is caught cheating on an exam, and experiments with wild and self-destructive behavior. She is finally caught by her father almost having sex with a guy she met near the docks after telling her friend, "I want to do something that I've never done before." This serves as a wake-up call for her parents, particularly her father. He quits his job to work with her on her soccer training.\\ \textbf{Question:} What happens to the sibling that supports Gracie's interest in soccer?\\ \textbf{Give the rationale before answering.} Garcie's older brother Johnny who was supportive of her interest in soccer died in a car accident. So the answer is \emph{dies in a traffic accident.}\\ \#\#\# \\ \dots \\ \#\#\#\\ \textbf{Passage:} Bath once had an important manufacturing sector, particularly in crane manufacture, furniture manufacture, printing, brass foundries, quarries, dye works and Plasticine manufacture, as well as many mills. Significant Bath companies included Stothert \& Pitt, Bath Cabinet Makers and Bath \& Portland Stone. Nowadays, manufacturing is in decline, but the city boasts strong software, publishing and service-oriented industries, being home to companies such as Future plc and London \& Country mortgage brokers. The city's attraction to tourists has also led to a significant number of jobs in tourism-related industries. Important economic sectors in Bath include education and health (30,000 jobs), retail, tourism and leisure (14,000 jobs) and business and professional services (10,000 jobs). Major employers are the National Health Service, the city's two universities, and the Bath and North East Somerset Council, as well as the Ministry of Defence although a number of MOD offices formerly in Bath have recently moved to Bristol. Growing employment sectors include information and communication technologies and creative and cultural industries where Bath is one of the recognised national centres for publishing, with the magazine and digital publisher Future plc employing around 650 people. Others include Buro Happold (400) and IPL Information Processing Limited (250). The city boasts over 400 retail shops, half of which are run by independent specialist retailers, and around 100 restaurants and cafes primarily supported by tourism.\\ \textbf{Question:} half of what are run by independent specialists?\\ \textbf{Give the rationale before answering.}\end{tabular} \\ \midrule

\end{tabular}
\caption{The exact prompts \textbf{with CoT} used for each contrast set (continuation of Table \ref{tab:cot-contrast3}).}
\label{tab:cot-contrast4}
\end{table*}

\begin{table*}[]
\footnotesize
\begin{tabular}{@{}p{.12\linewidth}p{.9\linewidth}@{}}
\toprule
\textbf{Task} &
  \textbf{Prompt} \\
\toprule
CondaQA &
  \begin{tabular}[c]{@{}p{1\linewidth}@{}}\textbf{Passage:} \\ The end of the long-held animosity between Moscow and Beijing was marked by the visit to China by Soviet General Secretary Mikhail Gorbachev in 1989. After the 1991 demise of the Soviet Union, China's relations with Russia and the former states of the Soviet Union became more amicable as the conflicting ideologies of the two vast nations no longer stood in the way. A new round of bilateral agreements was signed during reciprocal head of state visits. As in the early 1950s with the Soviet Union, Russia has again become an important source of military technology for China, as well as for raw materials and trade. Friendly relations with Russia have been an important advantage for China, complementing its strong ties with the U.S.\\ \textbf{Question:} Can China rely on both the US and Russia as supportive allies?\\ \textbf{The answer is} \emph{YES.}\\ \#\#\#\\ ... \\ \#\#\#\\ \textbf{Passage:} Trams have operated continuously in Melbourne since 1885 (the horse tram line in Fairfield opened in 1884, but was at best an irregular service). Since then they have become a distinctive part of Melbourne's character and feature in tourism and travel advertising. Melbourne's cable tram system opened in 1885, and expanded to one of the largest in the world, with of double track. The first electric tram line opened in 1889, but closed only a few years later in 1896. In 1906 electric tram systems were opened in St Kilda and Essendon, marking the start of continuous operation of Melbourne's electric trams.\\ \textbf{Question:} If I wanted to take a horse tram in 1884, could I look up the next tram on a schedule?\\ \end{tabular} \\ \midrule
BoolQ &
  \begin{tabular}[c]{@{}p{1\linewidth}@{}}\textbf{Passage:} Love Island Australia is an Australian dating reality show based on the British series Love Island. The series is presented by Sophie Monk and narrated by Eoghan McDermott. The series began airing on 9Go! and 9Now on 27 May 2018. The final aired on 5 July 2018, with Grant Crapp and Tayla Damir winning and sharing the \$50,000 prize money. Eden Dally and Erin Barnett finished as runners up.\\ \textbf{Question:} is there a prize for love island australia\\ \textbf{The answer is} \emph{True.}\\ \#\#\# \\ \dots \\ \#\#\#\\ \textbf{Passage:} Conscription in the United States, commonly known as the draft, has been employed by the federal government of the United States in five conflicts: the American Revolution, the American Civil War, World War I, World War II, and the Cold War (including both the Korean War and the Vietnam War). The third incarnation of the draft came into being in 1940 through the Selective Training and Service Act. It was the country's first peacetime draft. From 1940 until 1973, during both peacetime and periods of conflict, men were drafted to fill vacancies in the United States Armed Forces that could not be filled through voluntary means. The draft came to an end when the United States Armed Forces moved to an all-volunteer military force. However, the Selective Service System remains in place as a contingency plan; all male civilians between the ages of 18 and 25 are required to register so that a draft can be readily resumed if needed. United States Federal Law also provides for the compulsory conscription of men between the ages of 17 and 45 and certain women for militia service pursuant to Article I, Section 8 of the United States Constitution and 10 U.S. Code § 246.\\ \textbf{Question:} was there a draft in the revolutionary war\\\end{tabular} \\ \midrule
PERSPECTRUM &
  \begin{tabular}[c]{@{}p{1\linewidth}@{}}Interfering with a species for cosmetic reasons is poor practice\\ The breeding of white tigers in captivity should be banned. \\ Does the perspective support or undermine the claim? \\ Answer with: supports, undermines or not a valid perspective.\\ \textbf{The answer is} \emph{supports.}\\ \#\#\# \\ \dots \\ \#\#\#\\ I would just be so upset that that guy cheated and stole that moment away from me.\\ should cheaters be given a second chance. \\ Does the perspective support or undermine the claim? \\ Answer with: supports, undermines or not a valid perspective.\\ \end{tabular} \\ 
  \bottomrule
  \end{tabular}
\caption{The exact prompts \textbf{without CoT} used for each contrast set (continued in Tables \ref{tab:nocot-contrast2}-\ref{tab:nocot-contrast4}).}
\label{tab:nocot-contrast1}
\end{table*}

\begin{table*}
\footnotesize
\begin{tabular}{@{}p{.1\linewidth}p{.9\linewidth}@{}}
\toprule
\textbf{Task} &
  \textbf{Prompt} \\
\toprule
ROPES &
  \begin{tabular}[c]{@{}p{1\linewidth}@{}}\textbf{Passage:} Hemophilia is the name of a group of hereditary diseases that affect the body's ability to control blood clotting. Hemophilia is caused by a lack of clotting factors in the blood. Clotting factors are normally released by platelets. Since people with hemophilia cannot produce clots, any cut can put a person at risk of bleeding to death. The risk of internal bleeding is also increased in hemophilia, especially into muscles and joints. This disease affected the royal families of Europe. Greg and Bobby are coworkers who are currently eating lunch together on their break. They haven't known each other very long and are having a casual conversation when Bobby mentions that he is a hemophiliac. Greg, unsure of what exactly that means, looks confused and taken aback. Bobby, noticing the bewilderment on Greg's face, takes a moment to explain the details of his condition to Greg. \\ \textbf{Question:} Does Greg or Bobby have a lower platelet count?\\ \textbf{The answer is} \emph{Bobby.}\\ \#\#\# \\ \dots \\ \#\#\#\\ \textbf{Passage:} The theory of evolution by natural selection was proposed at about the same time by both Charles Darwin and Alfred Russel Wallace, shown in Figure below , and was set out in detail in Darwin's 1859 book On the Origin of Species . Natural selection is a process that causes heritable traits that are helpful for survival and reproduction to become more common, and harmful traits, or traits that are not helpful or advantageous for survival to become more rare in a population of organisms. This occurs because organisms with advantageous traits are more "fit" to survive in a particular environment and have "adapted" to the conditions of that environment. These individuals will have greater reproductive success than organisms less fit for survival in the environment. This will lead to an increase in the number of organisms with the advantageous trait(s) over time. Over many generations, adaptations occur through a combination of successive, small, random changes in traits, and natural selection of those variants best-suited for their environment. Natural selection is one of the cornerstones of modern biology.\\ \textbf{\textbf{Question:}} The tree frogs range covers 2 different habitats, One has a lot of trees and the other a lot of tall green grass with few if any trees. Tree frogs can either be green or brown. Which color tree frog has an advantage living in trees?\\ \end{tabular} \\ \midrule
DROP &
  \begin{tabular}[c]{@{}p{1\linewidth}@{}}\textbf{Passage:} Švitrigaila was losing his influence in the Slavic principalities and could no longer resist Poland and Sigismund. On 4 September 1437 he attempted to reconcile with Poland: he would rule the lands that still backed him , and after his death these territories would pass to the King of Poland. However, under strong protest from Sigismund, the Polish Senate declined to ratify the treaty . In 1438 Švitrigaila withdrew to Moldavia. The reign of Sigismund Kęstutaitis was brief — he was assassinated in 1440. Švitrigaila returned from exile in 1442 and ruled Lutsk until his death a decade later. Jogaila's son Casimir IV Jagiellon, born in 1426, received approval as a hereditary hospodar from Lithuania's ruling families in 1440. This event is seen by the historians Jerzy Lukowski and Hubert Zawadzki as marking the end of the succession dispute.\\ \textbf{Question:} Who was ruler first, Sigismund of Casimir IV Jagiellon?\\ \textbf{The answer is} \emph{Sigismund Kęstutaitis.}\\ \#\#\# \\ \dots \\ \#\#\#\\ \textbf{Passage:} The origins of al-Qaeda can be traced back to the Soviet war in Afghanistan . The United States, the United Kingdom, Saudi Arabia, Pakistan, and the People's Republic of China supported the Islamist Afghan mujahadeen guerillas against the military forces of the Soviet Union and the Democratic Republic of Afghanistan. A small number of "Afghan Arab" volunteers joined the fight against the Soviets, including Osama bin Laden, but there is no evidence they received any external assistance. In May 1996 the group World Islamic Front for Jihad Against Jews and Crusaders , sponsored by bin Laden , started forming a large base of operations in Afghanistan, where the Islamist extremist regime of the Taliban had seized power earlier in the year. In February 1998, Osama bin Laden signed a fatwā, as head of al-Qaeda, declaring war on the West and Israel. Earlier in August 1996, Bin Laden declared jihad against the United States. Later in May 1998 al-Qaeda released a video declaring war on the U.S. and the West. On 7 August 1998, al-Qaeda struck the U.S. embassies in Kenya and Tanzania, killing 224 people, including 12 Americans. In retaliation, U.S. President Bill Clinton launched Operation Infinite Reach, a bombing campaign in Sudan and Afghanistan against targets the U.S. asserted were associated with WIFJAJC, although others have questioned whether a pharmaceutical plant in Sudan was used as a chemical warfare facility. The plant produced much of the region's antimalarial drugs and around 50\% of Sudan's pharmaceutical needs. The strikes failed to kill any leaders of WIFJAJC or the Taliban. Next came the 2000 millennium attack plots, which included an attempted bombing of Los Angeles International Airport. On 12 October 2000, the USS Cole bombing occurred near the port of Yemen, and 17 U.S. Navy sailors were killed.\\ \textbf{Question:} How many years was it between when Bin Laden declared a war against the United States and when he became leader of al-Qaeda?\\ \end{tabular} \\ \midrule
\end{tabular}
\caption{The exact prompts \textbf{without CoT} used for each contrast set (continuation of Table \ref{tab:nocot-contrast1}).}
\label{tab:nocot-contrast2}
\end{table*}

\begin{table*}
\footnotesize
\begin{tabular}{@{}p{.1\linewidth}p{.9\linewidth}@{}}
\toprule
\textbf{Task} &
  \textbf{Prompt} \\
\toprule  

IMDb &
  \begin{tabular}[c]{@{}p{1\linewidth}@{}}\textbf{Review:} "Garde à Vue has to be seen a number of times in order to understand the sub-plots it contains. If you're not used to french wordy films, based upon conversation and battle of wits rather than on action, don't even try to watch it. You'll only obtain boredom to death, and reassured opinion that french movies are not for you.\textless{}br /\textgreater{}\textless{}br /\textgreater{}Garde à Vue is a wordy film, essentially based upon dialogs (written by Audiard by the way)and it cruelly cuts the veil of appearances.\textless{}br /\textgreater{}\textless{}br /\textgreater{}Why does Maître Martineau (Serrault) prefer to be unduly accused of being a child murderer rather than telling the truth ? Because at the time of the murder he was with a 18 years old girl with which he has a 8-years sexual relation. His wife knows it, she's jealous of it and he prefers to be executed (in 1980 in France, there was still death penalty)rather than unveiling the sole ""pure and innocent"" aspect of his pitiful life."\\ \textbf{What is the sentiment of this review:} Positive or Negative?\\ \textbf{The answer is} \emph{Positive.}\\ \#\#\# \\ \dots \\ \#\#\#\\ \textbf{Review:} May 2004, Wonderland is fairly new in the UK. Brilliant film of a brutal true story. If you know LA from the early 80's, you will appreciate how well it is captured. The use of the elements which make up its gritty cinematic style is original, amplifying the experience and bringing the viewer very close to actually being there. The use of a disjointed 'Pulp Fiction' style time line allows exploration of the uncertainty concerning what really happened, while the direction and performances of the cast command attention, especially Val Kilmer as John Holmes; an Oscar for sure if I were handing them out.\\ \textbf{What is the sentiment of this review:} Positive or Negative?\\ \end{tabular} \\ \midrule

MATRES &
  \begin{tabular}[c]{@{}p{1\linewidth}@{}}\textbf{Passage:} Dr. Barnett Slepian was  *killed*   in his kitchen by a sniper's bullet last fall . Investigators said Friday  they  *found*   a rifle buried  near his home in the Buffalo suburb of Amherst.\\ \textbf{Question:} When did the event *killed* happen in relation to the event *found*: before, after, simultaneously, or is it vague?\\ \textbf{The answer is} \emph{before.}\\ \#\#\# \\ \dots    \\ \#\#\#\\ \textbf{Passage:} A book of memoirs and photographs from the climb by Mr Lowe , which he worked on with Dr Lewis-Jones , is due to be published in May . He  *said* : " Lowe was a brilliant , kind fellow who never sought the limelight ... and 60 years on from Everest his achievements deserve wider recognition . " He was  *involved* in two of the most important explorations of the 20th Century ... yet remained a humble , happy man right to the end ... an inspirational lesson to us all . "\\ \textbf{Question:} When did the event *said* happen in relation to the event *involved*: before, after, simultaneously, or is it vague?\\ \end{tabular} \\ \bottomrule
\end{tabular}
\caption{The exact prompts \textbf{without CoT} used for each contrast set (continuation of Table \ref{tab:nocot-contrast2}).}
\label{tab:nocot-contrast3}
\end{table*}

\begin{table*}
\footnotesize
\begin{tabular}{@{}p{.1\linewidth}p{.9\linewidth}@{}}
\toprule
\textbf{Task} &
  \textbf{Prompt} \\
\toprule  
  MCTACO &
  \begin{tabular}[c]{@{}p{1\linewidth}@{}}\textbf{Passage:} He succeeds James A. Taylor, who stepped down as chairman, president and chief executive in March for health reasons. How often are chief executives typically replaced?\\ Is this the answer: every 10 weeks?\\ \textbf{The answer is} \emph{no.}\\ \#\#\# \\ \dots \\ \#\#\#\\ \textbf{Passage:} Roberta Adams skipped the thick how-to guide on child-custody forms and sat down at a computer at the Lamoreaux Justice Center in Orange on Wednesday. When did Roberta arrive at the Lamoreaux Justice Center?\\ Is this the answer: 10:45 PM?\\ \end{tabular} \\ \midrule
Quoref &
  \begin{tabular}[c]{@{}p{1\linewidth}@{}}\textbf{Passage:} In the year 1978, Gracie Bowen, a 15-year-old tomboy who lives in South Orange, New Jersey, is crazy about soccer, as are her three brothers and their former soccer star father. Although Gracie wants to join her brothers and neighbor Kyle in the nightly practices her father runs, she is discouraged by everyone except her older brother, Johnny. Johnny, Gracie and Kyle attend Columbia High School, where Johnny is the captain and star player for the varsity soccer team. After missing a shot at the end of a game, the despondent Johnny drives off with a friend's car and dies in a traffic accident. Struggling with grief, Gracie decides that she wants to replace her brother on the team. Her father does not believe that girls should play soccer, telling her she is neither tough nor talented enough. Her mother is a nurse who lacks the competitive drive of the rest of her family and fears for Gracie's safety. Her mother later tells Gracie that she would have liked to become a surgeon, but that option had not been available to her as a woman. Rejected and depressed, Gracie begins to rebel; she stops doing her schoolwork, is caught cheating on an exam, and experiments with wild and self-destructive behavior. She is finally caught by her father almost having sex with a guy she met near the docks after telling her friend, "I want to do something that I've never done before." This serves as a wake-up call for her parents, particularly her father. He quits his job to work with her on her soccer training.\\ \textbf{Question:} What happens to the sibling that supports Gracie's interest in soccer?\\ \textbf{The answer is} Garcie's older brother Johnny who was supportive of her interest in soccer died in a car accident. So the answer is \emph{dies in a traffic accident.}\\ \#\#\# \\ \dots \\ \#\#\#\\ \textbf{Passage:} Bath once had an important manufacturing sector, particularly in crane manufacture, furniture manufacture, printing, brass foundries, quarries, dye works and Plasticine manufacture, as well as many mills. Significant Bath companies included Stothert \& Pitt, Bath Cabinet Makers and Bath \& Portland Stone. Nowadays, manufacturing is in decline, but the city boasts strong software, publishing and service-oriented industries, being home to companies such as Future plc and London \& Country mortgage brokers. The city's attraction to tourists has also led to a significant number of jobs in tourism-related industries. Important economic sectors in Bath include education and health (30,000 jobs), retail, tourism and leisure (14,000 jobs) and business and professional services (10,000 jobs). Major employers are the National Health Service, the city's two universities, and the Bath and North East Somerset Council, as well as the Ministry of Defence although a number of MOD offices formerly in Bath have recently moved to Bristol. Growing employment sectors include information and communication technologies and creative and cultural industries where Bath is one of the recognised national centres for publishing, with the magazine and digital publisher Future plc employing around 650 people. Others include Buro Happold (400) and IPL Information Processing Limited (250). The city boasts over 400 retail shops, half of which are run by independent specialist retailers, and around 100 restaurants and cafes primarily supported by tourism.\\ \textbf{Question:} half of what are run by independent specialists?\\ \end{tabular} \\ \midrule

\end{tabular}
\caption{The exact prompts \textbf{without CoT} used for each contrast set (continuation of Table \ref{tab:nocot-contrast3}).}
\label{tab:nocot-contrast4}
\end{table*}

\clearpage

\begin{table*}[]
\tiny
\begin{tabular}{p{.3\linewidth}p{.3\linewidth}llll}
\toprule
\textbf{Original} & \textbf{Perturbed} & \textbf{Model} & \textbf{Prediction} & \textbf{GPT-4} & \textbf{Label} \\
\midrule
Oil exports flow as strike woes ease A general strike in Nigeria, which has raised fears over oil supply from the world \#39;s seventh-largest exporter, will likely end its first phase on Thursday  quot;all going well quot;, union leaders said. & Oil exports flow as strike woes ease A massive strike in Nigeria, which has raised fears over oil supply from the world \#39;s seventh-largest exporter, will likely end its first phase on Thursday  quot;all going well quot;, union leaders said. & OPT-13B & Business & Business & World \\ \midrule
Veritas Keeps Reaching into Its Wallet By acquiring KVault, which makes e-mail-archiving software, it aims to erode EMC \#39;s lead and rebuild investors \#39; confidence. & Veritas Keeps Reaching into Its Wallet By acquiring ups, which makes e-mail-archiving software, it aims to erode EMC \#39;s lead and rebuild investors \#39; confidence. & T5-3B & Business & Business & Sci/Tech \\ \midrule
Thailand Shows No Easy War Against Wildlife Crime (Reuters) Reuters - With an AK-47 assault rifle slung over\textbackslash his shoulder, Sompong Prajobjan roamed one of Thailand's lush\textbackslash national parks for more than a decade. & Thailand Shows No Easy War Against mass Crime (Reuters) Reuters - With an AK-47 assault rifle slung over\textbackslash his shoulder, Sompong Prajobjan roamed one of Thailand's lush\textbackslash national parks for more than a decade. & OPT-6.7B & World & World & Sci/Tech \\ \midrule
Google stock falls as share lockups expire SAN FRANCISCO, - Shares of Google Inc. fell as much as 6.5 percent Tuesday, as selling restrictions were lifted on 39 million shares held by employees and early investors in the newly public Web search company. & Google stock falls as share lockups expire la FRANCISCO, - Shares of Google Inc. fell as much as 6.5 percent Tuesday, as selling restrictions were lifted on 39 million shares held by employees and early investors in the newly public Web search service. & GPT-2-Large & Sci/Tech & Business & Business \\ \midrule
VZ Wireless Slams National 411 Directory WASHINGTON -- Verizon Wireless, the nation \#39;s largest wireless carrier, clashed with other cellular carriers on Tuesday, telling a US Senate committee that a proposal for a national wireless telephone directory is a  quot;terrible idea quot; and that the proposal & network Wireless Slams National 411 Directory WASHINGTON -- Verizon Wireless, the nation \#39;s largest wireless carrier, clashed with other cellular carriers on Tuesday, telling a US Senate committee that a proposal for a national wireless telephone directory is a  quot;terrible idea quot; and that the proposal & OPT-125M & Sci/Tech & Sci/Tech & Business \\ \midrule
Douglas, Fraser, and blue await you Jim McLeod has a great day job, but a seasonal sideline is his  \#39; \#39;tree \#39; \#39; calling. Throughout the year, he \#39;s president and owner of a software company called InfoCode Corp. & Douglas, Fraser, and blue await you Jim McLeod has a regular day job, but a seasonal sideline is his  \#39; \#39;special \#39; \#39; calling. Throughout the year, he \#39;s president and owner of a software firm called InfoCode Corp. & GPT-2-Medium & Sci/Tech & Business & Business \\ \midrule
Pa. Golfer Cleared of Not Yelling 'Fore' (AP) AP - A golfer plunked in the face by an errant ball was unable to convince a jury that the man who hit him was negligent for failing to yell "Fore!" & Pa. Golfer Cleared of Not Yelling 'golf (AP) pa - A golfer plunked in the face by an errant ball was able to convince a jury that the man who assaulted him was negligent for failing to yell "golf!" & RoBERTa-Base & World & Sports & Sports \\ \midrule
Intel lauds milestone in shrinking chips Contradicting fears that the semiconductor industry \#39;s pace of development is slowing, Intel Corp has announced that it has achieved a milestone in shrinking the size of transistors that will power its next-generation chips. & Intel lauds milestone in growing chips Contradicting fears that the semiconductor industry \#39;s pace of development is slowing, Intel Corp has announced that it has achieved a milestone in shrinking the size of transistors that will power its next-generation chips. & OPT-1.3B & Business & Sci/Tech & Sci/Tech \\ \midrule
Earthquake Rocks Indonesia's Bali, One Dead  BALI, Indonesia (Reuters) - A powerful earthquake rocked  Indonesia's premier tourist island of Bali Wednesday, killing  one person, injuring at least two and triggering some panic,  officials said. & Earthquake Rocks Indonesia's Bali, One Dead  bali, Indonesia (Reuters) - A powerful earthquake rocked  Indonesia's premier tourist island of Bali Wednesday, killing  one person, injuring at least two and triggering some panic,  officials said. & T5-Large & Sci/Tech & World & World \\ \midrule
Surviving Biotech's Downturns Charly Travers offers advice on withstanding the volatility of the biotech sector. & Surviving Biotech's Downturns Charly s offers advice on withstanding the volatility of the biotech sector. & T5-Base & Sci/Tech & Business & Business \\
\bottomrule
\end{tabular}
\caption{Examples of AGNews attacked by BAE.}
\label{tab:agnews-bae}
\end{table*}

\begin{table*}[]
\tiny 
\begin{tabular}{p{.3\linewidth}p{.3\linewidth}llll}
\toprule
\textbf{Original} & \textbf{Perturbed} & \textbf{Model} & \textbf{Prediction} & \textbf{GPT-4} & \textbf{Label}  \\
\midrule
Jordan prince loses succession Jordan \#39;s Prince Hamzah says he is conceding to the wish of King Abdullah II to strip him of his crown as heir to the throne.  quot;I obey the command of my elder brother out of my loyalty, love & Jordan princ loses successioz JBordan \#q39;s Pricne Hamzah asys he is conceidng to the wish of Kng Abdullha II to strip him of his crwn as hier to the thorne.  qwot;I obZey the ommand of my elder brother out of my loyalKy, love & OPT-13B & Sports & World & World \\ \midrule
Trust Digital Gets CEO, Cash Influx Trust Digital Inc., a McLean software company, is getting a new chief executive and \textbackslash \$3.1 million in new investments as it tries to expand its business making security software for wireless devices. \&lt;FONT face="verdana,MS Sans Serif,arial,helvetica" size="-2"\textbackslash \space color="\#666666"\&gt;\&lt;B\&gt;-The Washington Post\&lt;/B\&gt;\&lt;/FONT\&gt; & Trust Digiutal GetP COE, Cash Influx Trust Dgiital Inc., a McLean software cmopany, is getting a new chief executive and \textbackslash \$3.1 million in new investments as it ries to expand its business making security software for wireless devices. \&lt;FONT face="verdana,MS Sans Serif,arial,helvetica" size="-2"\textbackslash \space color="\#666666"\&gt;\&lt;B\&gt;-The Washington Post\&lt;/B\&gt;\&lt;/FONT\&gt; & T5-3B & Business & Business & Sci/Tech \\ \midrule
PeopleSoft sweetens employee compensation Business software maker PeopleSoft Friday said it was boosting compensation packages for all employees except its chief executive in a move that would raise the & PeopleSoft sweetens employee ompensation Busienss oftware maker PeoplSeoft Fridya maid it was Zboosting comensation pakcages for all employees except its cihef eGecutive in a Aove that wound raise the & OPT-6.7B & Sci/Tech & Business & Business \\ \midrule
Second Acts Former House speaker Thomas M. Finneran is the new president of the Massachusetts Biotechnology Council, a trade group that counts more than 400 members, including Genzyme Corp. and Biogen Idec Inc., the two largest biotechnology companies in the state. Its previous president left under pressure earlier this year, and some members say they chose Finneran, who quit his legislative post ... & Second Acts Former House speaker Thmoas M. Fineran is the new presqident of the Massachusemts Biotechnolgy Council, a trade group that counts more than 400 members, including Genzyme Corp. and Biogen Idec InOc., the two largest biotechnology cmpanies in the stabe. Its pDrevious rpesident left under pressure earlMer this year, and some members say they chose Finneran, who quit his legislative post ... & GPT-2-Large & World & Business & Business \\ \midrule
Putin Says Russia Working on New Nuclear Systems (Reuters) Reuters - Russia is working on new nuclear missile\textbackslash systems that other powers do not have in order to protect\textbackslash itself against future security challenges, President Vladimir\textbackslash Putin said Wednesday. & Putin Says Russia Working on New Nuclear Systems (Reuers) Reuters - Russia is working on new nuclera missile\textbackslash systems that other powers do not have in order to protect\textbackslash itself against future security challenges, Poresident Vladimir\textbackslash Putin said Wednvesday. & OPT-125M & Sci/Tech & World & World \\ \midrule
Challenger disappoints with writedown The Kerry Packer-backed Challenger Financial Services Group has reported its first net loss since incorporating, impacted by a massive writedown of goodwill. & Challenger disappoints with writedown The KFrry Packer-babked ChallengeBr Financial eSrvices Gnoup has reporetd its first net lxss since incorporating, impacted by a massive writedown of goodwill. & GPT-2-Medium & Sci/Tech & Business & Business \\ \midrule
Hewlett-Packard buys Synstar Palo Alto-based Hewlett-Packard Co. has bought IT services company Synstar plc, of Bracknell, England for about \textbackslash \$293.3 million. Synstar has some 1,500 customers across Europe, selling IT support for various computer platforms. & Hewlett-Packard ubys Synstar Palo Alto-based Hewlett-Packard Co. has bought IT services company Synstar plc, of Bracknell, England for about \textbackslash \$293.3 million. Synstar has some 1,500 customers across Europe, selling IT support for various computer platforms. & RoBERTa-Base & Sci/Tech & Business & Business \\ \midrule
Alaska \#39;s summer tourism pegged at 1.4 million visitors The number of summer visitors to Alaska rose from the year before, prompting the president of the Alaska Travel Industry Association to say tourism appeared to be back on track since leveling off after the 2001 terrorist attacks. & Alaska \#39;s summer tourims pegged at 1.4 million visitors The number of summer visitors to Alaska rose from the year before, prompting the president of the Alaska TraPel IndAstry Association to say tourism appeared to be back on track since leveling off after the 2001 terrorist attacks. & OPT-1.3B & Sci/Tech & World & Business \\ \midrule
Two top Nokia executives resign HELSINKI: Nokia said the respected head of its networks unit had resigned and another top networks official left in the second major departure of top management in two weeks at the world \#39;s largest mobile phone maker. & Two top Nokia executives resign HELSINKI: Nokia said the respected head of its networks unit had resigned and another top networks official left in the second major dparture of top management in two weeks at the world \#39;s largest mobile phone aker. & T5-Large & Sci/Tech & Business & Business \\ \midrule
ATA: Customers Won \#39;t Be Affected By Bankruptcy INDIANAPOLIS -- ATA says it will honor all tickets and maintain its full schedule, after filing for Chapter 11 bankruptcy Tuesday. & ATA: Cusomers Won \#39;t Be Affected By Bankruptcy INDIANAPOLIS -- ATEA says it will honor all tickets and maintain its full scheduel, after filing for hCapter k1 bankruptcy Tuesday. & T5-Base & Sports & Business & Business \\
\bottomrule
\end{tabular}
\caption{Examples of AGNews attacked by DeepWordBug.}
\label{tab:agnews-deepwordbug}
\end{table*}

\begin{table*}[]
\tiny
\begin{tabular}{p{.3\linewidth}p{.3\linewidth}llll}
\toprule
\textbf{Original} & \textbf{Perturbed} & \textbf{Model} & \textbf{Prediction} & \textbf{GPT-4} & \textbf{Label}  \\
\midrule
Sun looks for props with new server, storage hardware Sun Microsystems will hold its quarterly product launch this week, unleashing a raft of new hardware offerings spanning servers to storage. & Sun search for property with new server, store hardware Sun Microsystems will admit its quarterly product launch this week, unleashing a passel of newfangled hardware offerings spanning servers to memory. & OPT-13B & Business & Sci/Tech & Sci/Tech \\ \midrule
Caymas to open with security gateways Security start-up Caymas Systems launches Monday with products to protect the flow of corporate data. & Caymas to clear with surety gateways protection start-up Caymas organisation plunge Monday with merchandise to protect the menstruate of corporal data. & T5-3B & Business & Business & Sci/Tech \\ \midrule
U.S. Asks Laos to Check Massacre Report (AP) AP - The State Department said Monday it is taking seriously allegations that Laotian military forces may have massacred children of the country's Hmong ethnic minority. & U.entropy. Asks Laos to learn slaughter study (AP) AP - The commonwealth Department said Monday it is consider seriously allegations that Laotian military force-out may have massacred tyke of the country's Hmong pagan minority. & OPT-6.7B & Sci/Tech & World & World \\ \midrule
EU Move on Cyprus Eases Way for Turkey Deal  BRUSSELS (Reuters) - The European Union and Turkey inched  toward a historic agreement on starting membership talks on  Friday as EU leaders softened their demands on the crucial  sticking point of Cyprus. & EU Move on Cyprus Eases Way for Turkey Deal  BRUSSELS (Reuters) - The European Union and Turkey inched  toward a historic agreement on starting membership talks on  Friday as europium leaders softened their demands on the crucial  sticking point of Cyprus. & GPT-2-Large & Business & World & World \\ \midrule
Smith setback for Windies West Indies have been forced to make a second change to their Champions Trophy squad because of injury. Dwayne Smith is suffering from a shoulder problem and has been replaced by Ryan Hinds. & metalworker blow for Windies Occident indie have been thrust to puddle a 2nd commute to their fighter prize team because of trauma. Dwayne metalworker is brook from a berm job and has been replaced by Ryan hind. & OPT-125M & Sci/Tech & Sports & Sports \\ \midrule
Hewlett-Packard buys Synstar Palo Alto-based Hewlett-Packard Co. has bought IT services company Synstar plc, of Bracknell, England for about \textbackslash \$293.3 million. Synstar has some 1,500 customers across Europe, selling IT support for various computer platforms. & Hewlett-Packard buys Synstar Palo Alto-based Hewlett-Packard Co. has corrupt IT services society Synstar plc, of Bracknell, England for about \textbackslash \$293.3 million. Synstar has some 1,500 customers across Europe, selling IT support for various computer platforms. & GPT-2-Medium & Sci/Tech & Business & Business \\ \midrule
Liu brings China 4th gold in weightlifting at Athens Games ATHENS, Aug. 19 (Xinhuanet) -- Chinese Hercules Liu Chunhong Thursday lifted three world records on her way to winning the women \#39;s 69kg gold medal at the Athens Olympics, the fourth of the power sport competition for China. & Liu brings chinaware quaternary amber in weightlifting at Athens Games ATHENS, Aug. 19 (Xinhuanet) -- Chinese Hercules Liu Chunhong Thursday lifted 3 domain records on her way to winning the womanhood \#39;s 69kg Au medal at the Athens Olympics, the quarter of the tycoon athletics competition for chinaware. & RoBERTa-Base & Sci/Tech & Sports & Sports \\ \midrule
Gibbs doubts Skins were tipping plays If the Cleveland Browns knew what plays the Washington Redskins were going to run before the ball was snapped Sunday, a review of the game tape 24 hours later revealed scant evidence of it. & Gibbs dubiety tegument were tap gaming If the Cleveland embrown knew what gambling the WA Injun were going to operate before the testis was snap Sunday, a inspection of the crippled videotape 24 hours belated revealed light evidence of it. & OPT-1.3B & Business & Sports & Sports \\ \midrule
England's Lawyers Try to Get Photos Thrown Out Lawyers for Pfc. Lynndie R. England sought Wednesday to throw out evidence at the heart of the Abu Ghraib prison scandal -- the now-infamous photos showing her smiling and pointing at naked Iraqi detainees. & England's Lawyers taste to stimulate Photos project KO'd Lawyers for perfluorocarbon. Lynndie R. England sought Wed to throw out evidence at the kernel of the Abu Ghraib prison scandal -- the now-infamous photos showing her smiling and pointing at naked Iraqi detainees. & T5-Large & Sci/Tech & World & World \\ \midrule
MLB: NY Yankees 7, Minnesota 6 (12 inn.) Hideki Matsui drove in Derek Jeter with a 12th-inning sacrifice fly Wednesday night, giving the New York Yankees a dramatic, 7-6 win over Minnesota. & MLB: NY Yankees heptad, Minnesota 6 (12 inn.) Hideki Matsui get in Derek Jeter with a 12th-inning sacrifice fly Wed Nox, collapse the New York Yankees a dramatic, 7-6 profits over Minnesota. & T5-Base & World & Sports & Sports \\
\bottomrule
\end{tabular}
\caption{Examples of AGNews attacked by PWWS.}
\label{tab:agnews-pwws}
\end{table*}

\begin{table*}[]
\tiny
\begin{tabular}{p{.3\linewidth}p{.3\linewidth}llll}
\toprule
\textbf{Original} & \textbf{Perturbed} & \textbf{Model} & \textbf{Prediction} & \textbf{GPT-4} & \textbf{Label}  \\
\midrule
Panama flooding kills nine people At least nine people - seven of them children - have died in flooding in the capital of Panama. The authorities say at least 13 people are still missing after heavy rainfall caused rivers to break their banks. & Panama flooding murder nine people At least ninth citizens - seventh of them kid - have die in floods in the capital of Panama. De authority sy at laest 13 people are again lacking after he avy rainall caused stream to rupture their banque. & OPT-13B & Business & World & World \\ \midrule
Hit TV series 24 goes from small screen to smaller screen (AFP) AFP - The hit US television show  quot;24 quot; is going from the small screen to the smaller after 20th Century Fox and Vodaphone struck a groundbreaking deal to distribute the drama on mobile telephones. & Hit TV series 24 goes from small screen to smaller screen (AFP) AFP - The hit US television show  quot;24 quot; is going from the small screen to the smaller after 20th Century Fox and Vodaphone struck a groundbreaking deal to distribute the drama on moible telephones. & T5-3B & World & Sci/Tech & Sci/Tech \\ \midrule
Cali Cartel Boss Sent to U.S. on Drug Charges  BOGOTA, Colombia (Reuters) - The former boss of the Cali  drug cartel, who once controlled most of the world's cocaine  trade, was sent to the United States on Friday to face  trafficking and money laundering charges. & Cali Car tel Boss Sent to wu.S. on Drug Charges  BOGOTA, Colombia (Reuters) - The former boss of the Ca li  drug cartel, who once controlled most of the w orld's cocaine  trade, was sent to the United States on Friday to ace  trafficking and money laundering charges. & OPT-6.7B & Business & World & World \\ \midrule
BOND REPORT CHICAGO (CBS.MW) - Treasurys remained solidly lower Wednesday in the wake of election results that had President Bush ahead of Democratic challenger John Kerry. & BONDS APPRISE CHICAGO (CAS.TURBINES) - Treasurys remained solidly lwoer Wednesday in the wake of election results that had President Bush ahead of Democratic challenger John Kerry. & GPT-2-Large & World & Business & Business \\ \midrule
Iran shuts reformist websites WEBSITES CLOSE to Iran \#39;s leading reformist party have been blocked by religious hardliners in the police bureau of public morals. & Iran shuts reformist sites WEBSITES CLOSE to Iran \#39;s leading reformist party have been blocked by religious hardliners in the police bureau of public morals. & OPT-125M & World & World & Sci/Tech \\ \midrule
Two top Nokia executives resign HELSINKI: Nokia said the respected head of its networks unit had resigned and another top networks official left in the second major departure of top management in two weeks at the world \#39;s largest mobile phone maker. & Two top Nokia executives resign HELSINKI: Nokia said the respected head of its networks unity had resigned and another top networks official left in the second major departure of top management in two weeks at the monde \#39;s largest mobile phone maker. & GPT-2-Medium & Sci/Tech & Business & Business \\ \midrule
Two top Nokia executives resign HELSINKI: Nokia said the respected head of its networks unit had resigned and another top networks official left in the second major departure of top management in two weeks at the world \#39;s largest mobile phone maker. & Two top Nokia executives resign HELSINKI: Nokia said the respected head of its networks drives had resigned and another top networks official left in the second major departure of supremo management in two weeks at the world \#39;s largest mobile phone maker. & RoBERTa-Base & Sci/Tech & Business & Business \\ \midrule
Scandal won \#39;t go away ATHENS -- It was telling yesterday that the majority of the dozens of journalists who asked questions and attended a news conference into a Greek doping scandal were mostly Canadian. Question after question came from Canadians. We were all there, I think, ... & Scandal won \#39;t go away ATHENS -- It was telling yesterday that the majority of the dozens of journalists who asked questions and attended a news conference into a Greek fortification scandal were principally Population. Questions after ques tion cmae from Canadian. We were all there, I think, ... & OPT-1.3B & World & Sports & Sports \\ \midrule
Plane of NASCAR Team Hendrick Missing (AP) AP - A plane carrying members of the Hendrick Motorsports organization was missing Sunday after losing contact with the Federal Aviation Administration on its way to a NASCAR race, and a search was underway for the aircraft. & Plane of NASCAR Team Hendrick Missing (AP) AP - A plane carrying members of the Hendrick Motorsports organization was missing Sunday after losing contact with the Federal Aviation Administration on its way to a NASAR race, and a search was underway for the aircraft. & T5-Large & Sci/Tech & Sports & Sports \\ \midrule
Paula \#39;s going for gold PAULA RADCLIFFE has decided she WILL run in tonight \#39;s 10,000m race at the Athens Olympics. Today \#39;s dramatic decision comes just days after Britain \#39;s star long-distance runner was left weeping at the roadside after pulling up in the Olympic marathon. & Paula \#3;s going for golden PAOLA RADCLIFFE has deicded she WILL r un in to night \#39;s 1O,000m race at the Athens Olympics. Today \#39;s dramatic decision comes just days after Britain \#39;s star long-distance runner was left weeping at the brooch after pulling up in the Olympic marathon. & T5-Base & Sci/Tech & Sports & Sports \\
\bottomrule
\end{tabular}
\caption{Examples of AGNews attacked by TextBugger.}
\label{tab:agnews-textbugger}
\end{table*}

\begin{table*}[]
\tiny
\begin{tabular}{p{.3\linewidth}p{.3\linewidth}llll}
\toprule
\textbf{Original} & \textbf{Perturbed} & \textbf{Model} & \textbf{Prediction} & \textbf{GPT-4} & \textbf{Label}  \\
\midrule
Bush enforces new US diplomacy (AFP) AFP - President George W. Bush has rewritten US foreign policy during four years at the White House, with the war on terror now taking priority and doubt cast on some traditional alliances. & Bush enforces roman AMERICANS extraterrestrials (AFP) AFP - President George W. Buch enjoys rewritten US spacemen policy during four annum at the Blanco Household, with the war on fear now assuming supremacy and duda colada on some usual partner. & OPT-13B & Sci/Tech & World & World \\ \midrule
Dependent species risk extinction The global extinction crisis is worse than thought, because thousands of  quot;affiliated quot; species also at risk do not figure in calculations. & Subordinated species risk extinction The global extinction crisis is worse than thought, for thousands of  quot;associated quot; species also at menace fait not figure in calculations. & T5-3B & World & Sci/Tech & Sci/Tech \\ \midrule
Heineken Profit Dips But Repeats Outlook  AMSTERDAM (Reuters) - Dutch brewer Heineken posted a 4.5  percent fall in core profit for the first half on Wednesday, at  the low end of expectations as a weak dollar and sluggish  markets hurt business. & Heineken Profit Dips But Repeated Predictions  COPENHAGEN (Newsday) - Bassi blackbird Heineken seconded a 4.5  percent dip in keys winnings for the first half on Sunday, at  the tenuous termination of aspirations as a insufficient money and apathetic  marketplace defaced business. & OPT-6.7B & Sports & Business & Business \\ \midrule
MLB, Va. Officials Meet Chicago White Sox owner Jerry Reinsdorf led a team of negotiators from Major League Baseball in a three-hour meeting Wednesday with the leaders of the Virginia Baseball Stadium Authority. & MLB, Does. Officials Meet Boston Pai Sox owner Jerry Reinsdorf boosted a computers of negotiators from Major Society Hardball in a three-hour assembly Wednesday with the fuhrer of the Ginny Mitt Stadium Authority. & GPT-2-Large & Business & Sports & Sports \\ \midrule
Check Boeing sops, Airbustells US Check Boeing sops, Airbustells \&lt;b\&gt;...\&lt;/b\&gt; European aircraft maker Airbus on Thursday criticised a US move to take a fight about subsidies to the World Trade Organisation (WTO), saying it showed its rivals unwillingness to address its own subsidies. & Check Boeing sops, Airbustells V Controls Boeing sops, Airbustells \&lt;b\&gt;...\&lt;/b\&gt; European airforce setters Airliner on Jue critic a V resettled to bears a faces about financed to the Globo Negotiation Arranged (WTO), tells it found its hopefuls unwilling to tackled its distinctive awards. & OPT-125M & World & Business & Business \\ \midrule
Stocks Open Lower; Inflation Data Weighs  NEW YORK (Reuters) - U.S. stocks opened lower on Tuesday  after a government report showing a much larger-than-expected  rise in U.S. producer prices in October raised inflation  concerns. & Arsenals Open Lower; Blowing Data Peso  NOUVEAU BRONX (Reuters) - wu.ies. stocks opener least on Tuesday  after a government communique depicting a much larger-than-expected  escalating in oder.ies. farmers expenditure in Janeiro referred inflation  implicated. & GPT-2-Medium & Sports & Business & Business \\ \midrule
Stocks Open Lower; Inflation Data Weighs  NEW YORK (Reuters) - U.S. stocks opened lower on Tuesday  after a government report showing a much larger-than-expected  rise in U.S. producer prices in October raised inflation  concerns. & Stockpiles Open Lower; Inflation Data Weighs  NEW YORK (Reuters) - ni.r. sharing unblocked lower on Sonntag  after a government communique showcases a much larger-than-expected  rise in hu.p. ranchers prices in October raised inflation  feared. & RoBERTa-Base & World & Business & Business \\ \midrule
Dreams of perfect market are fine, as long as they don't come true In these times of financial wrongdoing and subsequent systemic changes, it's only natural to wonder what a perfect investment world would look like. & Reve of perfect market are awesome, as longer as they don't entered realistic Under these times of monetary malfunction and subsequent systemic modifications, it's only natural to wonder what a perfect capital world would visualise like. & OPT-1.3B & Sci/Tech & Business & Business \\ \midrule
Liu brings China 4th gold in weightlifting at Athens Games ATHENS, Aug. 19 (Xinhuanet) -- Chinese Hercules Liu Chunhong Thursday lifted three world records on her way to winning the women \#39;s 69kg gold medal at the Athens Olympics, the fourth of the power sport competition for China. & Liu establishes China 4th kim in gymnastics at Athens Games GRECO, Jun. 19 (Xinhuanet) -- Chinese Hercules Liang Chunhong Hoy lifted three world records on her arteries to attaining the feminine \#39;s 69kg gold decoration at the Poseidon Olympics, the fourth of the electricity sport hostilities for China. & T5-Large & World & Sports & Sports \\ \midrule
Packers lose Flanagan for the season Green Bay Packers Pro Bowl center Mike Flanagan will undergo surgery on his left knee and miss the rest of the season. Coach Mike Sherman made the announcement after practice Friday, meaning for the second & Slaughtering waste Mcgrath for the season Environmentalist Golfo Packers Pro Goblet clinics Geraldo Conner hope suffer surgeons on his leave hips and fails the rest of the seasons. Buses Michaela Sherman realised the publicity after reality Hoy, signify for the second & T5-Base & World & Sports & Sports \\
\bottomrule
\end{tabular}
\caption{Examples of AGNews attacked by TextFooler.}
\label{tab:ag-textfooler}
\end{table*}

\begin{table*}[t]
\tiny
\begin{tabular}{p{.3\linewidth}p{.3\linewidth}llll}
\toprule
\textbf{Original} & \textbf{Perturbed} & \textbf{Model} & \textbf{Prediction} & \textbf{GPT-4} & \textbf{Label}  \\
\midrule
\begin{tabular}[c]{@{}p{1\linewidth}@{}}\emph{Premise:} Lifetime Extension of SCR De-NOx Catalysts Using SCR-Tech's High Efficiency Ultrasonic Regeneration Process\\ \emph{Hypothesis:} There is a 10 year extension of SCR De-NOx catalysts.\end{tabular} & \begin{tabular}[c]{@{}p{1\linewidth}@{}}\emph{Premise:} ongoing Extension of SCR De-NOx Catalysts Using SCR-Tech's High Efficiency Ultrasonic Regeneration Process\\ \emph{Hypothesis:} There is a 10 year extension of SCR De-NOx catalysts.\end{tabular} & GPT-2 & Neutral & Neutral & Contradiction \\ \midrule
\begin{tabular}[c]{@{}p{1\linewidth}@{}}\emph{Premise:} Vrenna looked it and smiled.\\ \emph{Hypothesis:} Vreanna wore a pleased expression when she saw it.\end{tabular} & \begin{tabular}[c]{@{}p{1\linewidth}@{}}\emph{Premise:} Vrenna looked it and shuddered.\\ \emph{Hypothesis:} Vreanna wore a pleased expression when she saw it.\end{tabular} & DeBERTa-v3-Base & Contradiction & Contradiction & Entailment \\ \midrule
\begin{tabular}[c]{@{}p{1\linewidth}@{}}\emph{Premise:} There is a good restaurant in the village, in addition to a well-stocked mini-market for self-catering visitors.\\ \emph{Hypothesis:} The village has nowhere to dine.\end{tabular} & \begin{tabular}[c]{@{}p{1\linewidth}@{}}\emph{Premise:} There is a good restaurant in the village, in addition to a well-stocked mini-market for self-catering visitors.\\ \emph{Hypothesis:} another village has nowhere to dine.\end{tabular} & T5-Small & Neutral & Contradiction & Contradiction \\ \midrule
\begin{tabular}[c]{@{}p{1\linewidth}@{}}\emph{Premise:} Several of the organizations had professional and administrative staffs that provided analytical capabilities and facilitated their members' participation in the organization's activities.\\ \emph{Hypothesis:} Organizations didn't care about members' participation.\end{tabular} & \begin{tabular}[c]{@{}p{1\linewidth}@{}}\emph{Premise:} Several of the organizations had professional and administrative staffs that provided analytical capabilities and facilitated their members' participation in the organization's activities.\\ \emph{Hypothesis:} others didn't care about members' participation.\end{tabular} & OPT-350M & Neutral & Contradiction & Contradiction \\ \midrule
\begin{tabular}[c]{@{}p{1\linewidth}@{}}\emph{Premise:} He appropriated for the State much of the personal fortunes of the princes, but found it harder to curtail the power of land-owners who had extensive contacts with the more conservative elements in his Congress Party.\\ \emph{Hypothesis:} He was able to take much of the princes' individual fortunes for the State, but it was more difficult to wrest power from land owners in contact with the conservative elements of the Congress Party.\end{tabular} & \begin{tabular}[c]{@{}p{1\linewidth}@{}}\emph{Premise:} He appropriated for the country much of the personal fortunes of the princes, but found it harder to curtail the power of owners who had extensive contacts with the more conservative elements in his congress country.\\ \emph{Hypothesis:} He was able to take much of the princes' individual fortunes for the State, but it was more difficult to wrest power from land owners in contact with the conservative elements of the Congress Party.\end{tabular} & T5-Small & Neutral & Entailment & Entailment \\ \midrule
\begin{tabular}[c]{@{}p{1\linewidth}@{}}\emph{Premise:} But when he was persuaded by divers means to help us, he gave up after one week, declaring it beyond his powers.\\ \emph{Hypothesis:} He decided it was too difficult because he was distracted by other topics.\end{tabular} & \begin{tabular}[c]{@{}p{1\linewidth}@{}}\emph{Premise:} But when he was persuaded by divers means to help us, he gave up after one week, declaring it beyond his powers.\\ \emph{Hypothesis:} He decided it was too easy because he was distracted by other topics.\end{tabular} & DeBERTa-v3-Large & Contradiction & Contradiction & Neutral \\ \midrule
\begin{tabular}[c]{@{}p{1\linewidth}@{}}\emph{Premise:} Most of the dances are suggestive of ancient courtship rituals, with the man being forceful and arrogant, the woman shyly flirtatious.\\ \emph{Hypothesis:} Majority of the dances are influenced by hip hop.\end{tabular} & \begin{tabular}[c]{@{}p{1\linewidth}@{}}\emph{Premise:} many of the dances are suggestive of ancient courtship rituals, with the man being forceful and arrogant, the woman shyly flirtatious.\\ \emph{Hypothesis:} Majority of the dances are influenced by hip hop.\end{tabular} & GPT-2 & Neutral & Contradiction & Contradiction \\ \midrule
\begin{tabular}[c]{@{}p{1\linewidth}@{}}\emph{Premise:} exactly and when i'm sitting here on the sofa cross-stitching and all of a sudden somebody a man's got their hand on my door knob it's like uh like oh no and so i don't i don't like that and i guess the only way to prevent it would be just to pass a city ordinance to prevent that or\\ \emph{Hypothesis:} I don't ever sit on my sofa and do cross-stitch.\end{tabular} & \begin{tabular}[c]{@{}p{1\linewidth}@{}}\emph{Premise:} exactly and when get sitting here on the sofa cross-stitching and all of a sudden somebody a man's got their hand on my door knob it's like uh like oh no and so i don't i don't like that and i guess the only way to prevent it would be just to pass a city ordinance to prevent that or\\ \emph{Hypothesis:} I don't ever sit on my sofa and do cross-stitch.\end{tabular} & OPT-2.7B & Neutral & Contradiction & Contradiction \\ \midrule
\begin{tabular}[c]{@{}p{1\linewidth}@{}}\emph{Premise:} Do not talk.\\ \emph{Hypothesis:} Don't speak until they all leave.\end{tabular} & \begin{tabular}[c]{@{}p{1\linewidth}@{}}\emph{Premise:} Do not talk.\\ \emph{Hypothesis:} please speak until they all leave.\end{tabular} & GPT-2-Medium & Contradiction & Contradiction & Neutral \\ \midrule
\begin{tabular}[c]{@{}p{1\linewidth}@{}}\emph{Premise:} The University of Nevada-Las Vegas boasts a student population over 23,000 (though, like most of the people in Las Vegas, they are commuters).\\ \emph{Hypothesis:} Most of the students of The University of Nevada are commuters.\end{tabular} & \begin{tabular}[c]{@{}p{1\linewidth}@{}}\emph{Premise:} The University of Nevada-Las Vegas boasts a student population over 23,000 (though, like most of the people in Las Vegas, they are foreigners).\\ \emph{Hypothesis:} Most of the students of The University of Nevada are commuters.\end{tabular} & OPT-6.7B & Neutral & Neutral & Entailment \\
\bottomrule
\end{tabular}
\caption{Examples of MNLI attacked by BAE.}
\label{tab:mnli-bae}
\end{table*}

\begin{table*}[]
\tiny
\begin{tabular}{p{.3\linewidth}p{.3\linewidth}llll}
\toprule
\textbf{Original} & \textbf{Perturbed} & \textbf{Model} & \textbf{Prediction} & \textbf{GPT-4} & \textbf{Label}  \\
\midrule
\begin{tabular}[c]{@{}p{1\linewidth}@{}}\emph{Premise:} The draft treaty was Tommy's bait.\\ \emph{Hypothesis:} Tommy took the bait of the treaty.\end{tabular} & \begin{tabular}[c]{@{}p{1\linewidth}@{}}\emph{Premise:} The draft treaty was Tommy's bait.\\ \emph{Hypothesis:} Tommy took the bjit of the treaty.\end{tabular} & OPT-13B & Contradiction & Contradiction & Neutral \\ \midrule
\begin{tabular}[c]{@{}p{1\linewidth}@{}}\emph{Premise:} The anthropologist Napoleon Chagnon has shown that Yanomamo men who have killed other men have more wives and more offspring than average guys.\\ \emph{Hypothesis:} Yanomamo men who kill other men have better chances at getting more wives.\end{tabular} & \begin{tabular}[c]{@{}p{1\linewidth}@{}}\emph{Premise:} The anthropologist Napoleon Chagnon has shown that YEnomamo men who have killed other men have more wies and more offspring than average guys.\\ \emph{Hypothesis:} Yanomamo men who kill other men have bettler chances at gzetting more wives.\end{tabular} & T5-3B & Neutral & Entailment & Entailment \\ \midrule
\begin{tabular}[c]{@{}p{1\linewidth}@{}}\emph{Premise:} The providers worked with the newly created Legal Assistance to the Disadvantaged Committee of the Minnesota State Bar Association (MSBA) to create the Minnesota Legal Services Coalition State Support Center and the position of Director of Volunteer Legal Services, now the Access to Justice Director at the Minnesota State Bar Association.\\ \emph{Hypothesis:} The Access to Justice Director was formerly called the Director of Volunteer Legal Services.\end{tabular} & \begin{tabular}[c]{@{}p{1\linewidth}@{}}\emph{Premise:} The providers worked with the newly created Legal Assistance to the Disadvantaged Committee of the Minnesota State Bar Association (MSBA) to create the Minnesota Legal Services Coalition State Support Center and the position of Director of Voluteer Legal Services, now the Access to Justice Director at the Minnesota State Bar Association.\\ \emph{Hypothesis:} The Access to Justice Director was formerly called the Director of Volunteer Legal Services.\end{tabular} & OPT-6.7B & Neutral & Entailment & Entailment \\ \midrule
\begin{tabular}[c]{@{}p{1\linewidth}@{}}\emph{Premise:} Tax purists would argue that the value of the homemakers' hard work--and the intrafamily benefits they presumably receive in return for it--should, in fact, be treated as income and taxed, just like the wages paid to outside service providers such as baby sitters and housekeepers.\\ \emph{Hypothesis:} To tax purists, the value of the homemakers' hard work should not be taxed.\end{tabular} & \begin{tabular}[c]{@{}p{1\linewidth}@{}}\emph{Premise:} Tax purists would argue that the value of the Khomemakers' hard owrk--and the intrafamily benefits they presumably receive in return for it--shZuld, in fact, be treated as income and txed, just like the wages paid to outside service providers such as baby sitters and housekeepers.\\ \emph{Hypothesis:} o Sax uprists, the Hvalue of the homemiakers' hard worO should not be taxeOd.\end{tabular} & GPT-2-Large & Neutral & Contradiction & Contradiction \\ \midrule
\begin{tabular}[c]{@{}p{1\linewidth}@{}}\emph{Premise:} Also, the tobacco executives who told Congress they didn't consider nicotine addictive might now be prosecuted for fraud and perjury.\\ \emph{Hypothesis:} Some tobacco executives told Congress nicotine is not addictive.\end{tabular} & \begin{tabular}[c]{@{}p{1\linewidth}@{}}\emph{Premise:} Also, the tobacco executives who told Congress they didn't consider nicotine addictive might now be prosecuted for fraud and perjury.\\ \emph{Hypothesis:} SoEme tobacco executives told Congress nicotine is not addictiAve.\end{tabular} & OPT-125M & Contradiction & Entailment & Entailment \\ \midrule
\begin{tabular}[c]{@{}p{1\linewidth}@{}}\emph{Premise:} 4 million, or about 8 percent of total expenditures for the two programs).\\ \emph{Hypothesis:} The figure of 4 million is likely to rise in the coming years.\end{tabular} & \begin{tabular}[c]{@{}p{1\linewidth}@{}}\emph{Premise:} 4 mlilion, or about 8 percent of total expenditures for the two programs).\\ \emph{Hypothesis:} The figure of 4 million is likelo to rsie in the coming yeavs.\end{tabular} & GPT-2-Medium & Contradiction & Neutral & Neutral \\ \midrule
\begin{tabular}[c]{@{}p{1\linewidth}@{}}\emph{Premise:} Although claims data provide the most accurate information about health care use, ensuring adequate follow-up for purposes of obtaining information from patient self-report is important because many people do not report alcohol-related events to insurance compa-nies.\\ \emph{Hypothesis:} Patients naturally always report to insurance companies when health problems may be a direct result of alcohol.\end{tabular} & \begin{tabular}[c]{@{}p{1\linewidth}@{}}\emph{Premise:} Although claims data provide the most accurate information about health care use, ensuring adequate follow-up for purposes of obtaining information from patient self-report is important because many people do not report alohol-related events to insurance compa-nies.\\ \emph{Hypothesis:} Patients naturally always report to insurance companies when health problems may be a direct result of alcohol.\end{tabular} & RoBERTa-Base & Neutral & Contradiction & Contradiction \\ \midrule
\begin{tabular}[c]{@{}p{1\linewidth}@{}}\emph{Premise:} Today the strait is busy with commercial shipping, ferries, and fishing boats, and its wooded shores are lined with pretty fishing villages, old Ottoman mansions, and the villas of Istanbul's wealthier citizens.\\ \emph{Hypothesis:} Today, the strait is empty after a huge sand storm killed everyone there.\end{tabular} & \begin{tabular}[c]{@{}p{1\linewidth}@{}}\emph{Premise:} Today the strati is busy with commercial shipping, ferries, and fishing boats, and its wooded shores are lined with pretty fishing villages, old Ottoman mansions, and the villas of Istanbul's wealthier citizens.\\ \emph{Hypothesis:} Today, the strait is mpty after a huge sand storm kliled everyone there.\end{tabular} & OPT-1.3B & Neutral & Contradiction & Contradiction \\ \midrule
\begin{tabular}[c]{@{}p{1\linewidth}@{}}\emph{Premise:} I have a situation.\\ \emph{Hypothesis:} Everything is fine and I have nothing on my mind.\end{tabular} & \begin{tabular}[c]{@{}p{1\linewidth}@{}}\emph{Premise:} I have a situation.\\ \emph{Hypothesis:} EvCrything is fien and I have notying on my minkd.\end{tabular} & T5-Large & Neutral & Contradiction & Contradiction \\ \midrule
\begin{tabular}[c]{@{}p{1\linewidth}@{}}\emph{Premise:} She had thrown away her cloak and tied her hair back into a topknot to keep it out of the way.\\ \emph{Hypothesis:} She shaved her head.\end{tabular} & \begin{tabular}[c]{@{}p{1\linewidth}@{}}\emph{Premise:} She had thrown away her cloak and zied her hair back into a topknot to keep it out of the way.\\ \emph{Hypothesis:} She shavfd her head.\end{tabular} & T5-Base & Entailment & Contradiction & Contradiction \\
\bottomrule
\end{tabular}
\caption{Examples of MNLI attacked by DeepWordBug.}
\label{tab:mnli-deepwordbug}
\end{table*}

\begin{table*}[]
\tiny
\begin{tabular}{p{.3\linewidth}p{.3\linewidth}llll}
\toprule
\textbf{Original} & \textbf{Perturbed} & \textbf{Model} & \textbf{Prediction} & \textbf{GPT-4} & \textbf{Label}  \\
\midrule
\begin{tabular}[c]{@{}p{1\linewidth}@{}}\emph{Premise:} This is an excerpt from the voice-over credo read in the opening credits for the new UPN series Star Pitiful Helpless Giant , starring former Secretary of State George Shultz.\\ \emph{Hypothesis:} Star Pitiful Helpless Giant is a show on WGN.\end{tabular} & \begin{tabular}[c]{@{}p{1\linewidth}@{}}\emph{Premise:} This is an excerpt from the voice-over credo read in the initiative acknowledgment for the unexampled UPN series Star Pitiful Helpless Giant , starring former Secretary of State George Shultz.\\ \emph{Hypothesis:} Star Pitiful Helpless Giant is a show on WGN.\end{tabular} & OPT-13B & Neutral & Neutral & Contradiction \\ \midrule
\begin{tabular}[c]{@{}p{1\linewidth}@{}}\emph{Premise:} The game of billiards is also hot.\\ \emph{Hypothesis:} People hate playing billiards.\end{tabular} & \begin{tabular}[c]{@{}p{1\linewidth}@{}}\emph{Premise:} The game of billiards is too blistering.\\ \emph{Hypothesis:} People hate playing billiards.\end{tabular} & T5-3B & Neutral & Neutral & Contradiction \\ \midrule
\begin{tabular}[c]{@{}p{1\linewidth}@{}}\emph{Premise:} Several of the organizations had professional and administrative staffs that provided analytical capabilities and facilitated their members' participation in the organization's activities.\\ \emph{Hypothesis:} Organizations didn't care about members' participation.\end{tabular} & \begin{tabular}[c]{@{}p{1\linewidth}@{}}\emph{Premise:} Several of the organizations had professional and administrative staffs that provided analytical capabilities and alleviate their members' participation in the organization's activities.\\ \emph{Hypothesis:} Organizations didn't like about members' participation.\end{tabular} & OPT-6.7B & Neutral & Contradiction & Contradiction \\ \midrule
\begin{tabular}[c]{@{}p{1\linewidth}@{}}\emph{Premise:} Tax purists would argue that the value of the homemakers' hard work--and the intrafamily benefits they presumably receive in return for it--should, in fact, be treated as income and taxed, just like the wages paid to outside service providers such as baby sitters and housekeepers.\\ \emph{Hypothesis:} To tax purists, the value of the homemakers' hard work should not be taxed.\end{tabular} & \begin{tabular}[c]{@{}p{1\linewidth}@{}}\emph{Premise:} Tax purists would argue that the value of the homemakers' hard work--and the intrafamily benefits they presumably receive in return for it--should, in fact, be treated as income and task, just like the wages paid to outside service providers such as baby sitters and housekeepers.\\ \emph{Hypothesis:} To tax purists, the value of the homemakers' strong workplace should not be taxed.\end{tabular} & GPT-2-Large & Neutral & Contradiction & Contradiction \\ \midrule
\begin{tabular}[c]{@{}p{1\linewidth}@{}}\emph{Premise:} Well, we've just got to get down to it, that's all.\\ \emph{Hypothesis:} We should take a break from this.\end{tabular} & \begin{tabular}[c]{@{}p{1\linewidth}@{}}\emph{Premise:} Well, we've just got to dumbfound down to it, that's all.\\ \emph{Hypothesis:} We should take a break from this.\end{tabular} & OPT-125M & Neutral & Neutral & Contradiction \\ \midrule
\begin{tabular}[c]{@{}p{1\linewidth}@{}}\emph{Premise:} exactly and when i'm sitting here on the sofa cross-stitching and all of a sudden somebody a man's got their hand on my door knob it's like uh like oh no and so i don't i don't like that and i guess the only way to prevent it would be just to pass a city ordinance to prevent that or\\ \emph{Hypothesis:} I don't ever sit on my sofa and do cross-stitch.\end{tabular} & \begin{tabular}[c]{@{}p{1\linewidth}@{}}\emph{Premise:} exactly and when i'm sitting here on the sofa cross-stitching and all of a sudden somebody a man's got their hand on my door knob it's like uh like oh no and so i don't i don't like that and i guess the only way to prevent it would be just to pass a city ordinance to prevent that or\\ \emph{Hypothesis:} I don't always sit on my sofa and do cross-stitch.\end{tabular} & GPT-2-Medium & Neutral & Neutral & Contradiction \\ \midrule
\begin{tabular}[c]{@{}p{1\linewidth}@{}}\emph{Premise:} Such multicolored reef dwellers as the parrotfish and French angelfish, along with weirdly shaped coral, crawfish, or turtles hiding in crevices, can be yours for the viewing in these clear waters where visibility of 30 m (100 ft) is common.\\ \emph{Hypothesis:} Since the reef has been bleached and rendered lifeless by global warming, it's no longer possible to see sea life in the cloudy water.\end{tabular} & \begin{tabular}[c]{@{}p{1\linewidth}@{}}\emph{Premise:} Such multicolored reef dwellers as the parrotfish and French angelfish, along with weirdly shaped coral, crawfish, or turtles hiding in crevices, can be yours for the viewing in these unclouded waters where visibility of 30 m (100 ft) is uncouth.\\ \emph{Hypothesis:} Since the Rand has been bleached and rendered lifeless by global warming, it's no long potential to see sea sprightliness in the cloudy H2O.\end{tabular} & RoBERTa-Base & Neutral & Contradiction & Contradiction \\ \midrule
\begin{tabular}[c]{@{}p{1\linewidth}@{}}\emph{Premise:} Homes or businesses not located on one of these roads must place a mail receptacle along the route traveled.\\ \emph{Hypothesis:} The other roads are far too rural to provide mail service to.\end{tabular} & \begin{tabular}[c]{@{}p{1\linewidth}@{}}\emph{Premise:} Homes or businesses not located on unity of these roads moldiness place a mail receptacle along the route traveled.\\ \emph{Hypothesis:} The other roads are ALIR too rural to allow mail service to.\end{tabular} & OPT-1.3B & Contradiction & Contradiction & Neutral \\ \midrule
\begin{tabular}[c]{@{}p{1\linewidth}@{}}\emph{Premise:} and that you're very much right but the jury may or may not see it that way so you get a little anticipate you know anxious there and go well you know\\ \emph{Hypothesis:} Even if you're correct, I think the jury would pick up on that.\end{tabular} & \begin{tabular}[c]{@{}p{1\linewidth}@{}}\emph{Premise:} and that you're very much right but the jury may or may not escort it that way so you get a little anticipate you know anxious there and go well you know\\ \emph{Hypothesis:} Even if you're correct, I think the jury would pick up on that.\end{tabular} & T5-Large & Neutral & Neutral & Contradiction \\ \midrule
\begin{tabular}[c]{@{}p{1\linewidth}@{}}\emph{Premise:} The technical how-tos for these three strategies will be summarized later in this paper.\\ \emph{Hypothesis:} There are seven strategies discussed in the paper.\end{tabular} & \begin{tabular}[c]{@{}p{1\linewidth}@{}}\emph{Premise:} The technical how-tos for these leash strategies will be summarized later in this paper.\\ \emph{Hypothesis:} There are seven strategies discussed in the paper.\end{tabular} & T5-Base & Neutral & Neutral & Contradiction \\
\bottomrule
\end{tabular}
\caption{Examples of MNLI attacked by PWWS.}
\label{tab:mnli-pwws}
\end{table*}

\begin{table*}[]
\tiny
\begin{tabular}{p{.3\linewidth}p{.3\linewidth}llll}
\toprule
\textbf{Original} & \textbf{Perturbed} & \textbf{Model} & \textbf{Prediction} & \textbf{GPT-4} & \textbf{Label}  \\
\midrule
\begin{tabular}[c]{@{}p{1\linewidth}@{}}\emph{Premise:} It is the official solution, Liq.\\ \emph{Hypothesis:} This is officially the solution.\end{tabular} & \begin{tabular}[c]{@{}p{1\linewidth}@{}}\emph{Premise:} Him is the functionary solution, Liq.\\ \emph{Hypothesis:} This is officially the solution.\end{tabular} & OPT-350M & Neutral & Neutral & Entailment \\ \midrule
\begin{tabular}[c]{@{}p{1\linewidth}@{}}\emph{Premise:} Calcutta seems to be the only other production center having any pretensions to artistic creativity at all, but ironically you're actually more likely to see the works of Satyajit Ray or Mrinal Sen shown in Europe or North America than in India itself.\\ \emph{Hypothesis:} Most of Mrinal Sen's work can be found in European collections.\end{tabular} & \begin{tabular}[c]{@{}p{1\linewidth}@{}}\emph{Premise:} Calcutta seems to be the only other production center having any pretensions to artistic creativity at all, but ironically you're actually more likely to see the works of Satyajit Ray or Mrinal Sen shown in Europe or North America than in India itself.\\ \emph{Hypothesis:} Msot of Mrinal Sen's work can be found in European collections.\end{tabular} & RoBERTa-Large & Entailment & Neutral & Neutral \\ \midrule
\begin{tabular}[c]{@{}p{1\linewidth}@{}}\emph{Premise:} It is at the moment of maximum audience susceptibility that we hear, for the first time, that the woman was fired not because of her gender but because of her sexual preference.\\ \emph{Hypothesis:} The woman was fired not for her sexual preference but purely on the basis of gender.\end{tabular} & \begin{tabular}[c]{@{}p{1\linewidth}@{}}\emph{Premise:} It is at the moment of maximum audience susceptibility that we hear, for the first time, that the woman was fired not because of her gneder but because of her sex preference.\\ \emph{Hypothesis:} The woman was triggered not for her sexual preference but purely on the basis of sexual.\end{tabular} & T5-11B & Entailment & Contradiction & Contradiction \\ \midrule
\begin{tabular}[c]{@{}p{1\linewidth}@{}}\emph{Premise:} It is the official solution, Liq.\\ \emph{Hypothesis:} This is officially the solution.\end{tabular} & \begin{tabular}[c]{@{}p{1\linewidth}@{}}\emph{Premise:} It is the official solution, Liq.\\ \emph{Hypothesis:} This is officially the solution.\end{tabular} & GPT-2-Large & Contradiction & Entailment & Entailment \\ \midrule
\begin{tabular}[c]{@{}p{1\linewidth}@{}}\emph{Premise:} Like Arabs and Jews, Diamond warns, Koreans and Japanese are joined by blood yet locked in traditional enmity.\\ \emph{Hypothesis:} Koreans and Japanese have no tension between them.\end{tabular} & \begin{tabular}[c]{@{}p{1\linewidth}@{}}\emph{Premise:} Like Arabs and Jews, Diamond warns, Norwegians and Japanese are joined by blood still lock ed in traditional enmity.\\ \emph{Hypothesis:} Koreans and Japanese have no ension between them.\end{tabular} & T5-3B & Neutral & Neutral & Contradiction \\ \midrule
\begin{tabular}[c]{@{}p{1\linewidth}@{}}\emph{Premise:} Placido Domingo's appearance on the package, compellingly photographed in costume as the ancient King of Crete, (Anthony Tommasini, the New York Times ) is the main selling point for this new recording of one of Mozart's more obscure operas--a fact that does not make critics happy.\\ \emph{Hypothesis:} The attracting feature of the new Mozart recording is Placido Domingo's appearance.\end{tabular} & \begin{tabular}[c]{@{}p{1\linewidth}@{}}\emph{Premise:} Nunez Domingo's appearane on the package, compellingly photographed in costume as the ancient King of Crete, (Anthony Tommasini, the New York Times ) is the main selling point for this new recording of one of Mozart's more obscure operas--a fact that does not make critics happy.\\ \emph{Hypothesis:} The attracting feature of the new Mozart recording is Placido Domingo's appearance.\end{tabular} & OPT-1.3B & Contradiction & Entailment & Entailment \\ \midrule
\begin{tabular}[c]{@{}p{1\linewidth}@{}}\emph{Premise:} Those Creole men and women you'll see dancing it properly have been moving their hips and knees that way since childhood.\\ \emph{Hypothesis:} Creole dances are learned from childhood.\end{tabular} & \begin{tabular}[c]{@{}p{1\linewidth}@{}}\emph{Premise:} Those Creole men and women you'll see dancing it properly have been moving their hips and knees that way since childhood.\\ \emph{Hypothesis:} Creole dances are learned from childhood.\end{tabular} & OPT-1.3B & Neutral & Entailment & Entailment \\ \midrule
\begin{tabular}[c]{@{}p{1\linewidth}@{}}\emph{Premise:} Exhibitions are often held in the splendid entrance hall.\\ \emph{Hypothesis:} The exhibitions in the entrance hall are usually the most exciting.\end{tabular} & \begin{tabular}[c]{@{}p{1\linewidth}@{}}\emph{Premise:} Exhibitions are often hled in the splendid entrance hall.\\ \emph{Hypothesis:} The exhibitions in the entrance hall are usually the most exciting.\end{tabular} & GPT-2-XL & Entailment & Neutral & Neutral \\ \midrule
\begin{tabular}[c]{@{}p{1\linewidth}@{}}\emph{Premise:} The Edinburgh International Festival (held annually since 1947) is acknowledged as one of the world's most important arts festivals.\\ \emph{Hypothesis:} The festival showcases exhibitions from all seven continents.\end{tabular} & \begin{tabular}[c]{@{}p{1\linewidth}@{}}\emph{Premise:} The Edinburgh International Festival (held annually since 1947) is acknowledged as one of the world's most important arts festivals.\\ \emph{Hypothesis:} Nova festival showcases exhibitions from all seven continents.\end{tabular} & GPT-2 & Contradiction & Neutral & Neutral \\ \midrule
\begin{tabular}[c]{@{}p{1\linewidth}@{}}\emph{Premise:} Calcutta seems to be the only other production center having any pretensions to artistic creativity at all, but ironically you're actually more likely to see the works of Satyajit Ray or Mrinal Sen shown in Europe or North America than in India itself.\\ \emph{Hypothesis:} Most of Mrinal Sen's work can be found in European collections.\end{tabular} & \begin{tabular}[c]{@{}p{1\linewidth}@{}}\emph{Premise:} Calcutta seems to be the only other production center having any pretensions to artistry creativity at all, but ironically you're actually more likely to see the works of Satyajit Ray or Mrinal Sen shown in Europeans or North America than in India itself.\\ \emph{Hypothesis:} Most of Mrinal Sen's work can be found in European collections.\end{tabular} & GPT-2-Large & Entailment & Neutral & Neutral \\
\bottomrule
\end{tabular}
\caption{Examples of MNLI attacked by TextBugger.}
\label{tab:mnli-textbugger}
\end{table*}

\begin{table*}[t]
\tiny
\begin{tabular}{p{.3\linewidth}p{.3\linewidth}llll}
\toprule
\textbf{Original} & \textbf{Perturbed} & \textbf{Model} & \textbf{Prediction} & \textbf{GPT-4} & \textbf{Label}  \\
\midrule
\begin{tabular}[c]{@{}p{1\linewidth}@{}}\emph{Premise:} That is exactly what our head coupon issuer Alan Greenspan did in 1987--and what I believe he would do again.\\ \emph{Hypothesis:} This is what Greenspan did in 1987 and what I think he will do again, much to the detriment of the economy.\end{tabular} & \begin{tabular}[c]{@{}p{1\linewidth}@{}}\emph{Premise:} That is exactly what our head coupon issuer Alan Greenspan did in 1987--and what I believe he would do again.\\ \emph{Hypothesis:} Hong is what Greenspan didnt in 1987 and what I inkling he yearn do again, much to the afflict of the economically.\end{tabular} & GPT-2 & Contradiction & Contradiction & Neutral \\ \midrule
\begin{tabular}[c]{@{}p{1\linewidth}@{}}\emph{Premise:} GAO recommends that the Secretary of Defense revise policy and guidance\\ \emph{Hypothesis:} GAO recommends that the Secretary of Defense keep policy and guidance the same\end{tabular} & \begin{tabular}[c]{@{}p{1\linewidth}@{}}\emph{Premise:} GAO recommends that the Secretary of Defense revise policy and guidance\\ \emph{Hypothesis:} BRED insinuated that the Department of Defends conservation polices and hints the same\end{tabular} & DeBERTa-v3-Base & Neutral & Neutral & Contradiction \\ \midrule
\begin{tabular}[c]{@{}p{1\linewidth}@{}}\emph{Premise:} But when he was persuaded by divers means to help us, he gave up after one week, declaring it beyond his powers.\\ \emph{Hypothesis:} He decided it was too difficult because he was distracted by other topics.\end{tabular} & \begin{tabular}[c]{@{}p{1\linewidth}@{}}\emph{Premise:} But when he was persuaded by divers means to help us, he gave up after one week, declaring it beyond his powers.\\ \emph{Hypothesis:} He decided it was too rigid for he was distracted by other issuing.\end{tabular} & T5-Small & Contradiction & Neutral & Neutral \\ \midrule
\begin{tabular}[c]{@{}p{1\linewidth}@{}}\emph{Premise:} Today the strait is busy with commercial shipping, ferries, and fishing boats, and its wooded shores are lined with pretty fishing villages, old Ottoman mansions, and the villas of Istanbul's wealthier citizens.\\ \emph{Hypothesis:} Today, the strait is empty after a huge sand storm killed everyone there.\end{tabular} & \begin{tabular}[c]{@{}p{1\linewidth}@{}}\emph{Premise:} Today the strait is busy with commercial shipping, ferries, and fishing boats, and its wooded shores are lined with pretty fishing villages, old Ottoman mansions, and the villas of Istanbul's wealthier citizens.\\ \emph{Hypothesis:} Sundays, the strait is empty after a huge sand storm killed everyone there.\end{tabular} & OPT-350M & Neutral & Contradiction & Contradiction \\ \midrule
\begin{tabular}[c]{@{}p{1\linewidth}@{}}\emph{Premise:} get something from from the Guess Who or\\ \emph{Hypothesis:} Get something from the Guess Who,\end{tabular} & \begin{tabular}[c]{@{}p{1\linewidth}@{}}\emph{Premise:} get something from from the Guess Who or\\ \emph{Hypothesis:} Get something from the Bet Who,\end{tabular} & T5-Small & Contradiction & Contradiction & Entailment \\ \midrule
\begin{tabular}[c]{@{}p{1\linewidth}@{}}\emph{Premise:} However, co-requesters cannot approve additional co-requesters or restrict the timing of the release of the product after it is issued.\\ \emph{Hypothesis:} They cannot restrict timing of the release of the product.\end{tabular} & \begin{tabular}[c]{@{}p{1\linewidth}@{}}\emph{Premise:} However, co-requesters cannot approve additional co-requesters or restrict the timing of the release of the product after it is issued.\\ \emph{Hypothesis:} They cannot coerce timing of the discards of the product.\end{tabular} & DeBERTa-v3-Large & Contradiction & Neutral & Entailment \\ \midrule
\begin{tabular}[c]{@{}p{1\linewidth}@{}}\emph{Premise:} Tax purists would argue that the value of the homemakers' hard work--and the intrafamily benefits they presumably receive in return for it--should, in fact, be treated as income and taxed, just like the wages paid to outside service providers such as baby sitters and housekeepers.\\ \emph{Hypothesis:} To tax purists, the value of the homemakers' hard work should not be taxed.\end{tabular} & \begin{tabular}[c]{@{}p{1\linewidth}@{}}\emph{Premise:} Tax purists would argue that the value of the homemakers' hard work--and the intrafamily benefits they presumably receive in return for it--should, in fact, be treated as income and taxed, just like the wages paid to outside service providers such as baby sitters and housekeepers.\\ \emph{Hypothesis:} To tax purists, the appraised of the homemakers' hard work should not are imposition.\end{tabular} & GPT-2 & Neutral & Contradiction & Contradiction \\ \midrule
\begin{tabular}[c]{@{}p{1\linewidth}@{}}\emph{Premise:} Cop Bud White (Crowe) and Ed Exley (Pearce) almost mix it up (59 seconds) :\\ \emph{Hypothesis:} Bud White and Ed Exley almost mix it up.\end{tabular} & \begin{tabular}[c]{@{}p{1\linewidth}@{}}\emph{Premise:} Cop Bud White (Crowe) and Ed Exley (Pearce) almost mix it up (59 seconds) :\\ \emph{Hypothesis:} Bud White and Ed Exley almost melting it up.\end{tabular} & OPT-2.7B & Neutral & Contradiction & Entailment \\ \midrule
\begin{tabular}[c]{@{}p{1\linewidth}@{}}\emph{Premise:} According to this plan, areas that were predominantly Arab the Gaza Strip, the central part of the country, the northwest corner, and the West Bank were to remain under Arab control as Palestine, while the southern Negev Des?Yrt and the northern coastal strip would form the new State of Israel.\\ \emph{Hypothesis:} We want to give Palestine and Israel a two-state solution that benefits both of them.\end{tabular} & \begin{tabular}[c]{@{}p{1\linewidth}@{}}\emph{Premise:} According to this plan, areas that were predominantly Arab the Gaza Strip, the central part of the country, the northwest corner, and the West Bank were to remain under Arab control as Palestine, while the southern Negev Des?Yrt and the northern coastal strip would form the new State of Israel.\\ \emph{Hypothesis:} We going to give Palestine and Israel a two-state solution that prerogatives both of them.\end{tabular} & GPT-2-Medium & Entailment & Neutral & Neutral \\ \midrule
\begin{tabular}[c]{@{}p{1\linewidth}@{}}\emph{Premise:} Search out the House of Dionysos and the House of the Trident with their simple floor patterns, and the House of Dolphins and the House of Masks for more elaborate examples, including Dionysos riding a panther, on the floor of the House of Masks.\\ \emph{Hypothesis:} The House of Dolphins and the House of Masks are more elaborate than the House of Dionysos and the House of the Trident.\end{tabular} & \begin{tabular}[c]{@{}p{1\linewidth}@{}}\emph{Premise:} Search out the House of Dionysos and the House of the Trident with their simple floor patterns, and the House of Dolphins and the House of Masks for more elaborate examples, including Dionysos riding a panther, on the floor of the House of Masks.\\ \emph{Hypothesis:} The House of Dolphins and the House of Masks are more devising than the House of Dionysos and the House of the Trident.\end{tabular} & OPT-6.7B & Neutral & Neutral & Entailment \\
\bottomrule
\end{tabular}
\caption{Examples of MNLI attacked by TextFooler.}
\label{tab:mnli-textfooler}
\end{table*}

\begin{table}
\begin{tabular}{llrl}
\toprule
& \textbf{Model}  & \textbf{Size}  & \textbf{PT} \\
\midrule
\multirow{4.5}{*}{\rotatebox{90}{\small \emph{Encoder-Only}}}  & \href{https://huggingface.co/roberta-base}{RoBERTa-Base} \citep{zhuang-etal-2021-robustly}   & 124M  & \multirow{2}{*}{MLM}                   \\
& \href{https://huggingface.co/roberta-large}{RoBERTa-Large} \citep{zhuang-etal-2021-robustly}    & 355M    &                    \\
\arrayrulecolor{black!20}\cmidrule{2-4}
& \href{https://huggingface.co/microsoft/deberta-v3-base}{DeBERTa-v3-Base} \citep{DBLP:conf/iclr/HeLGC21}  & 184M    & \multirow{2}{*}{MLM}                    \\
& \href{https://huggingface.co/microsoft/deberta-v3-large}{DeBERTa-v3-Large} \citep{DBLP:conf/iclr/HeLGC21} & 435M   &                    \\
\arrayrulecolor{black}\midrule
\multirow{10}{*}{\rotatebox{90}{\small \emph{Decoder-Only}}} & \href{https://huggingface.co/facebook/opt-125m}{OPT-125M}   \citep{DBLP:journals/corr/abs-2205-01068}      & 125M    & \multirow{6}{*}{LM}                     \\
& \href{https://huggingface.co/facebook/opt-350m}{OPT-350M}        \citep{DBLP:journals/corr/abs-2205-01068} & 331M    &                     \\
& \href{https://huggingface.co/facebook/opt-1.3b}{OPT-1.3B}        \citep{DBLP:journals/corr/abs-2205-01068}  & 1.3B  &                     \\
& \href{https://huggingface.co/facebook/opt-2.7b}{OPT-2.7B}        \citep{DBLP:journals/corr/abs-2205-01068}  & 2.7B  &                     \\
& \href{https://huggingface.co/facebook/opt-6.7b}{OPT-6.7B}        \citep{DBLP:journals/corr/abs-2205-01068}  & 6.7B &                     \\
& \href{https://huggingface.co/facebook/opt-13b}{OPT-13B}         \citep{DBLP:journals/corr/abs-2205-01068} & 12.8B &                     \\
\arrayrulecolor{black!20}\cmidrule{2-4}
& \href{https://huggingface.co/gpt2}{GPT-2}  \citep{radford2019language}  & 124M   & \multirow{4}{*}{LM}                     \\
& \href{https://huggingface.co/gpt2}{GPT-2-Medium}  \citep{radford2019language}   & 354M    &                     \\
& \href{https://huggingface.co/gpt2-large}{GPT-2-Large}  \citep{radford2019language}   & 774M   &                     \\
& \href{https://huggingface.co/gpt2-xl}{GPT-2-XL}    \citep{radford2019language}     &   1.6B       &                     \\
\arrayrulecolor{black}\midrule
\multirow{5}{*}{\rotatebox{90}{\small\emph{Encoder-Decoder}}} & \href{https://huggingface.co/t5-small}{T5-Small}    \citep{2020t5}     & 60M     & \multirow{5}{*}{\rotatebox{90}{\makecell[l]{text-to-text\\MLM + MTL}}}       \\
& \href{https://huggingface.co/t5-base}{T5-Base}       \citep{2020t5}   & 222M    &        \\
& \href{https://huggingface.co/t5-large}{T5-Large}     \citep{2020t5}     & 737M    &      \\
& \href{https://huggingface.co/t5-3b}{T5-XL (3B)}      \citep{2020t5}    & 2.8B  &       \\
& \href{https://huggingface.co/t5-11b}{T5-XXL (11B)}   \citep{2020t5}     & 11.3B &    \\
\bottomrule
\end{tabular}%
\caption{Citation and source information for the 19 models used for finetuning. The name of each model contains a link to the implementation used.}
\label{tab:models_source_info}
\end{table}

\end{document}